\documentclass[%
 reprint,
 amsmath,amssymb,
 aps,
]{revtex4-2}

\usepackage{graphicx}
\usepackage{dcolumn}
\usepackage{bm}

\begin{document}

\preprint{APS/123-QED}
 
\title{Connecting NTK and NNGP:\\Unified Theoretical Framework for\\ Wide Neural Network Learning Dynamics}

\author{Yehonatan Avidan}
\affiliation{Racah Institute of Physics, The Hebrew University of Jerusalem, Jerusalem 91904, Israel}
\affiliation{Edmond and Lily Safra Center for Brain Sciences, Hebrew University, Jerusalem 91904, Israel}

\author{Qianyi Li}
\affiliation{The Biophysics Program, Harvard University, Cambridge, Massachusetts 02138, USA}
\affiliation{Center for Brain Science, Harvard University, Cambridge, Massachusetts 02138, USA}

\author{Haim Sompolinsky}
\affiliation{Racah Institute of Physics, The Hebrew University of Jerusalem, Jerusalem 91904, Israel}
\affiliation{Edmond and Lily Safra Center for Brain Sciences, Hebrew University, Jerusalem 91904, Israel}
\affiliation{Center for Brain Science, Harvard University, Cambridge, Massachusetts 02138, USA}


\begin{abstract}
Artificial neural networks have revolutionized machine learning in recent years, but a complete theoretical framework for their learning process is still lacking. Substantial theoretical advances have been achieved for wide networks, within two disparate theoretical frameworks: the Neural Tangent Kernel (NTK), which assumes linearized gradient descent dynamics, and the Bayesian Neural Network Gaussian Process (NNGP) framework. Here we unify these two theories using gradient descent learning dynamics with an additional small noise in an ensemble of wide deep networks. We construct an exact analytical theory for the network input-output function and introduce a new time-dependent Neural Dynamical Kernel (NDK) from which both NTK and NNGP kernels are derived. We identify two learning phases characterized by different time scales: an initial gradient-driven learning phase, dominated by deterministic minimization of the loss, in which the time scale is mainly governed by the variance of the weight initialization. It is followed by a slow diffusive learning stage, during which the network parameters sample the solution space, with a time constant that is determined by the noise level and the variance of the Bayesian prior. The two variance parameters can strongly affect the performance in the two regimes, particularly in sigmoidal neurons.  In contrast to the exponential convergence of the mean predictor in the initial phase, the convergence to the final equilibrium is more complex and may exhibit nonmonotonic behavior. By characterizing the diffusive learning phase, our work sheds light on the phenomenon of representational drift in the brain, explaining how neural activity can exhibit continuous changes in internal representations without degrading performance, either by ongoing weak gradient signals that synchronize the drifts of different synapses or by architectural biases that generate invariant code, i.e., task-relevant information that is robust against the drift process. This work closes the gap between the NTK and NNGP theories, providing a comprehensive framework for understanding the learning process of deep wide neural networks and for analyzing learning dynamics in biological neural circuits.

\end{abstract}

\maketitle


\section{\label{sec:level1}Introduction}

Despite the empirical success of artificial neural networks,
theoretical understanding of their underlying learning process is
still limited. One promising theoretical approach focuses on deep
wide networks, in which the number of parameters in each layer goes
to infinity whereas the number of training examples remains finite
\cite{hazan2015steps,jacot2018neural,lee2017deep,lee2019wide,matthews2018gaussian,neal1994priors,novak2018bayesian,novak2019neural,sohl2020infinite,williams1996computing,yang2019wide}.
In this regime, the neural network (NN) is highly over-parameterized,
and there is a degenerate space of solutions achieving zero training
error. Investigating the properties of the solution space offers an
opportunity for understanding learning in over-parameterized NNs \cite{chizat2020implicit,jin2020implicit,min2021explicit}.
The two well-studied theoretical frameworks in the infinite width
limit focus on two different scenarios for exploring the solution
space during learning. One considers randomly initialized NNs trained
with gradient descent dynamics, and the learned NN parameters are
largely dependent on their value at initialization. In this case,
the infinitely wide NN's input-output relation is captured by the
Neural Tangent Kernel (NTK) \cite{jacot2018neural,lee2019wide}.
The other scenario considers Bayesian neural networks with
an i.i.d. Gaussian prior over their parameters, and a learning-induced
posterior distribution. In this case, the statistics of the NN's input-output
relation in the infinite width limit is given by the Neural Network
Gaussian Process (NNGP) kernel \cite{lee2017deep,cho2009kernel}.
These two scenarios make different assumptions regarding the learning
process and regularization. Furthermore, the generalization performance
of the two kernels on benchmark datasets differs \cite{lee2020finite}. It is therefore
important to generate a unified dynamical process with a single set of priors
and regularizations that captures both
cases. From a neuroscience perspective, a better understanding of the exploratory
process leading to Bayesian equilibrium may shed light on the empirical
and hotly debated phenomenon of representational drift \cite{deitch2021representational,marks2021stimulus,rokni2007motor,schoonover2021representational,rule2019causes,qin2023coordinated,masset2022drifting}.
To this end, we derive a new analytical theory of the learning dynamics.

1. We derive analytical equations for the time evolution of the
input-output relation (i.e. the predictor) of a network learning with
Langevin gradient descent dynamics \cite{coffey2012langevin,welling2011bayesian}.
We show that the equations for the mean and variance of the predictor
are in the form of integral equations, and present their numerical solutions for benchmark datasets.

2. A new time-dependent kernel, the Neural Dynamical Kernel (NDK),
naturally emerges from our theory.
This kernel can be understood as a time-dependent generalization of
the known NTK.

3. Our theory reveals two important learning phases characterized
by different time scales: gradient-driven, and diffusive learning.
In the initial gradient-driven learning phase, the dynamics are primarily
governed by deterministic gradient descent and described by
the NTK theory. This phase is followed by the slow exploration stage,
during which the network parameters sample the solution space, ultimately
approaching the equilibrium posterior distribution corresponding to
NNGP (Another perspective on the two phases was offered in \cite{shwartz2017opening,ratzon2024representational}).

4. We show that the generalization error
may exhibit diverse behaviors during the diffusive learning phase
depending on the network activation function, initialization, and regularization strength. Our theory provides insights into the
roles of these hyper-parameters in the trajectory of the dynamics.

5. Through analysis of the temporal correlation between network weights
during diffusive learning, we show that despite the random diffusion
of hidden layer weights, the training error remains low due to learning signal causing continuous realignment of the readout and the hidden layer weights. 
Conversely, ceasing this signal decreases the network performance due to decorrelation of the representations, ultimately leading to degraded generalization. 
We derive conditions under which the performance upon completely decorrelated readout and hidden weights remains well above chance. 
This provides insight into potential mechanisms for maintaining cognitive computation in the presence of representational drift, which can be tested in biological neural circuits.

\section{Model}

In this section, we describe the learning dynamics of fully connected Deep Neural Networks (DNNs) using Langevin dynamics.
We first define the model and our notations. 

\subsection{Notations and Setup for the Dynamical Theory \protect\label{subsec:notations}}

We consider a fully connected DNN with an input ${\bf x}\in\mathbb{R}^{N_{0}}$,  $L$
hidden layers. and a single output $f(\Theta,{\bf x})$ (i.e. the predictor), where $\Theta$ denotes all weight parameters. The input-output function is given by: 
\begin{equation}
f(\Theta,{\bf x})=\frac{1}{\sqrt{N_{L}}}{\bf a}\cdot{\bf x}^{L},\ \ \ {\bf a}\in\mathbb{R}^{N_{L}}
\end{equation}
\begin{equation}
{\bf x}^{l}({\bf x})=\phi\left({\bf z}^{l}({\bf x})\right),\quad {\bf z}^{l}\in\mathbb{R}^{N_{l}},\quad l=1,...L
\end{equation}

where the preactivations ${\bf z}^{l}({\bf x})$ are defined as
\begin{equation}
{\bf z}^{l}({\bf x})=\frac{1}{\sqrt{N_{l-1}}}{\bf W}^{l}\cdot{\bf x}^{l-1}\left({\bf x}\right)
\end{equation}
$N_{l}$ denotes the number of nodes in hidden layer $l$, and $N_{0}$
is the input dimension. ${\bf a}\in\mathbb{R}^{N_{L}}$ denotes the linear readout
weights and ${\bf W}^{l}\in\mathbb{R}^{N_{l}\times N_{l-1}}$ denotes
the hidden layer weights between layers $l-1$ and $l$. $\phi\left({\bf z}\right)$
is an element-wise nonlinear function of the preactivation vector.
The set of all hidden layer weights is denoted as ${\bf W}\equiv\left\{ {\bf W}^{1},\cdots,{\bf W}^{L}\right\}$ and all the network parameters are denoted collectively as $\Theta\equiv\left\{ {\bf W},{\bf a}\right\} $. ${\bf x}^l$ stands for the activations of the neurons in the $l$-th layer, and ${\bf x}\in\mathbb{R}^{N_{0}}$ represents the input vector
to the first layer of the network (${\bf x}^{l=0}\equiv{\bf x}$).
The training data is a set of $P$ labeled examples $\mathcal{D}:\left\{ {\bf x}^{\mu},y^{\mu}\right\} _{\mu=1,\cdots,P}$
where ${\bf x}^{\mu}\in\mathbb{R}^{N_{0}}$ is a training data point, and
$y^{\mu}\in\mathbb{R}$ is the target label of ${\bf x}^{\mu}$.
It is convenient to define a vector that contains all the label values $Y\in\mathbb{R}^{P}$ and a vector of the predictor values
on all the training points $f_{\text{train}}\left(t\right)\in\mathbb{R}^{P},$
such that $f_{\text{train}}^{\mu}=f(\Theta,{\bf x}^{\mu})$.

We assume an architecture with a single output unit for the ease of notation. 
It is straightforward to generalize the model and our theory to multiple outputs,
(see SI Sec. \ref{sec:multiple outputs}).

We consider the following supervised learning cost function: 
\begin{equation}
E\left(\Theta_{t}|\mathcal{D}\right)=\frac{1}{2}\sum_{\mu=1}^{P}\left(f_{\text{train}}^{\mu}(t)-y^{\mu}\right)^{2}+\frac{T}{2\sigma^{2}}\left|\Theta_{t}\right|^{2}\label{eq:cost}
\end{equation}
The first term is the loss function, specifically square error empirical loss (SE loss), and the
second term is a regularization term that favors weights with small
$L_{2}$ norm (weight decay term). $L_2$ regularization term has been
shown to improve the generalization performance \cite{krogh1991simple, galanti2022characterizing}. We introduce the parameter $T\sigma^{-2}$ as controlling the relative strength of the regularization and the  SE loss. The reason for using both temperature $T$ and $\sigma$ is that the temperature $T$ separately controls the level of noise in the stochastic dynamics (as will be defined below), and $\sigma^2$ is equivalent to the variance of the Gaussian prior in a Bayesian framework. 

We consider gradient descent learning dynamics with an additive noise given by
continuous-time Langevin equation. The weights of the system
start from an i.i.d. Gaussian initial condition with zero mean and
variance $\sigma_{0}^{2}$. The weights evolve under gradient descent
with respect to the cost function above with noise
$\xi$:

\begin{equation}\label{eq:Langevin}
\frac{d}{dt}\Theta_{t}=-\nabla_{\Theta}E\left(\Theta_{t}\right)+\xi\left(t\right)
\end{equation}
where $\xi\left(t\right)$ has a white noise statistics $\left\langle \xi\left(t\right)\right\rangle =0,\left\langle \xi\left(t\right)\xi^{\top}\left(t^{\prime}\right)\right\rangle =2IT\delta\left(t-t^{\prime}\right)$
. The temperature $T$ controls the level of noise in the
system.  We note that during the learning dynamics defined above, the dataset $\mathcal{D}$ and inputs $\bf{x}$ are constant
in time. Hence, the time dependence of the predictor $f$ is through the dynamics of the weight parameters, i.e. $f(t,{\bf x})\equiv f(\Theta_{t},{\bf x})$. Likewise, ${\bf x}_t^{l}({\bf x})={\bf x}^{l}\left( {\bf{W}}_t,{\bf x}\right)$ and ${\bf z}_t^{l}({\bf x})={\bf z}^{l}({\bf{W}}_t,{\bf x})$. 

Given a distribution of initial weights, the Langevin dynamics defines a time-dependent posterior distribution on weight space, $P_t\left(\Theta\right)$, which converges at long times to an equilibrium Gibbs distribution,  $P_{eq}(\Theta)\propto\exp\left(-\frac{1}{T}E(\Theta)\right)$. This distribution
is equivalent to the posterior of the Bayesian formulation of
learning \cite{li2021statistical}. 

\textbf{The Dynamics of the Prior:} In the absence of training signal
the Langevin dynamics are a random walk with a quadratic potential
(an Orenstein-Ulenbeck process \cite{uhlenbeck1930theory}). The induced statistics of $\Theta$
is that of temporally correlated i.i.d Gaussian variables with zero
mean
\begin{equation}
\left\langle \Theta_{t}\right\rangle _{0}=0,\quad \left\langle \Theta_{t}\Theta_{t^{\prime}}^{\top}\right\rangle _{0}=m(t,t^{\prime})I\label{eq:prior statistics}
\end{equation}
\vspace{-2mm}
\begin{equation}
m(t,t^{\prime})=\sigma^{2}e^{-T\sigma^{-2}\left|t-t^{\prime}\right|}+\left(\sigma_{0}^{2}-\sigma^{2}\right)e^{-T\sigma^{-2}(t+t^{\prime})}\label{eq:mtt'}
\end{equation}
where $\left\langle \right\rangle _{0}$ denotes henceforth averaging
over the dynamics induced by the regularization and the noise. As expected,
$m(0,0)=\sigma_{0}^{2}$. At long times, the second term of Eq.\ref{eq:mtt'} representing
the transient of the dynamics vanishes and the dominant term is
$\sigma^{2}e^{-T\sigma^{-2}\left|t-t^{\prime}\right|}$, with no dependence on $\sigma_{0}^{2}$. 

\subsection{Two Phases of Learning in the Limit of Small Noise\label{subsec:two learning phases}}

We focus on the dynamics of overparameterized DNNs in the limit of small noise - $T\rightarrow0$. Overparameterization
creates a degeneracy, where many sets of parameters $\Theta$ achieve
a zero loss function, defining the solution space to the optimization
problem. While these states are equivalent in performance
on the training data, they differ in performance on unseen test examples.
At low $T$ the Gibbs distribution facilitates generalization by sampling the solution space
with $L_2$ bias. 

\begin{figure*}
\centering{}\includegraphics[width=1.00\textwidth]{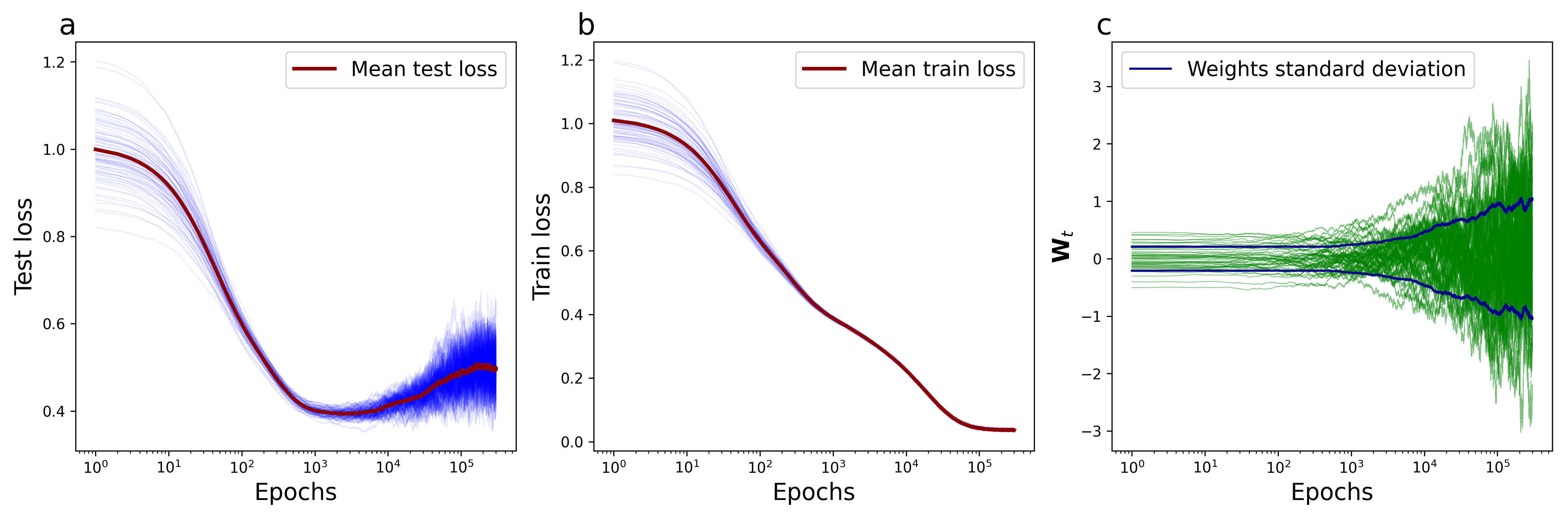}
\caption{Two Phases of Learning Dynamics: Simulation results of  a deep network with a single hidden layer and error function activation, trained by Langevin dynamics (Eq.\ref{eq:Langevin}) on binary classification using two classes from  CIFAR-10 dataset \cite{krizhevsky2014cifar}. 
(a) Test loss: The mean squared error (MSE) loss on test data reveals two distinct phases: an initial fast, approximately deterministic stage culminating in convergence to a low error and a subsequent slow, stochastic exploration phase characterized by large fluctuations. At long times, the network converges to an equilibrium state where the statistics of the weights and performance stabilize over time. (b) Training loss: The loss on the training data shows rapid relaxation to a state with low training error, with fluctuations on the order of $\mathcal{O}(T)$, indicating the restricted diffusive dynamics in the subspace of low training error.
(c) Weight dynamics: The weights exhibit a constrained random walk process, with their standard deviation gradually increasing from an initial value  $\sigma_{0}$ to the equilibrium value $\sigma$ (in this example $\sigma >\sigma_{0}$). This stochastic process remains confined to the solutions subspace as evident by the low training error (b). 
Parameter and details are in Sec.~\ref{sec:Details-of-the}.}
\label{fig: empirical figure}
\end{figure*}

In the small noise limit, a separation of time scales emerges due to the scales of the different components of the dynamics.
During the initial gradient-driven phase, the loss is $\mathcal{O}\left(1\right)$,
while the regularization term and the noise are $\mathcal{O}\left(T\right)$.
Thus, when the temperature is low, the loss dominates the dynamics,
making them approximately deterministic. After a time of $\mathcal{O}\left(1\right)$,
the network reaches a low training error solution where the loss function is
$\mathcal{O}\left(T\right)$. At this stage, the residual loss signal
is of the same order as the noise and regularization terms, leading
to richer dynamics as the solution space is explored. These dynamics
involve all three components and are no longer deterministic. As we show below, even in the infinite-width limit, these dynamics
are not simple exponential relaxation. We emphasize that even though
during the diffusive dynamics the fluctuations in training error remain $\mathcal{O}\left(T\right)$ (see Fig.\ref{fig: empirical figure} (b)).
The weights themselves undergo
a constrained random walk in the solution subspace (see Fig.\ref{fig: empirical figure} (c)), with fluctuations of the scale of $\sigma$.
The test performance also exhibits large fluctuations of the same order (Fig.\ref{fig: empirical figure} (a)).
At the end of the diffusive learning phase, the network reaches an
equilibrium state where the overall statistics of the weight
no longer change, converging to a Gibbs distribution. 

\section{Moment Generating Function of the Predictor in the Large Width
Limit}

The performance of the network is determined by its input-output function, hence we are mainly interested in the predictor statistics induced by the Langevin dynamics at all times. SI Sec.\ref{sec:Replica-calculation-of} presents a derivation
of a path-integral formulation of the above Langevin dynamics using
a Markov proximal learning framework.

Evaluating statistical quantities
using these integrals is generally intractable. However, in the infinite
width limit, where the hidden
layer widths are taken to infinity, i.e. $N_{1},\dots,N_{L}\rightarrow\infty$ , while the number of training examples
$P$ remains finite, the moments of the predictor can be derived from a moment 
generating function (MGF) formulation

\begin{equation}
    \mathcal{M}\left[\ell\right]\equiv\left\langle \exp\left(\sum_{{\bf x}}\int dt\ell(t,{\bf x})f(t,{\bf x})\right)\right\rangle _{\Theta}
\end{equation}
\vspace{-3mm}
\begin{equation}
    \left\langle f^{n}\left(t,{\bf x}\right)\right\rangle _{\Theta}=\left.\frac{\partial^{n}\mathcal{M}\left[\ell\right]}{\partial\ell^{n}(t,{\bf x})}\right|_{\ell(t,{\bf x})=0}
\end{equation}

The square brackets $[\cdot]$ represent a functional of the argument, $\sum_{\bf x}$ is summation over all the available data points ${\bf x}$ and the angular brackets denote the average of the statistics of
all the possible trajectories of the weight parameters, marginalizing over the noise and the random initial
condition.  $\ell(t,{\bf x})$ is the source term of the MGF, and taking derivatives w.r.t. the source yields the statistics of the predictor at time $t$ on specific input ${\bf x}$. In the infinite width limit, $\mathcal{M}\left[\ell\right]$
takes the form of a path integral over two time-dependent vectors,
$f_{\text{train}}(t)\in\mathbb{R}^{P}$ and $u(t)\in\mathbb{R}^{P}$.
$f_{\text{train}}(t)$ was introduced before, and is the vector gathering the predictor
values on the training examples, while $u(t)$ is an auxiliary field which mediates the interactions between $f_{\text{train}}(t)$ at different times.
\onecolumngrid
\begin{align}
 \mathcal{M}\left[\ell\right]=\intop Df_{\text{train}}\intop Du\exp\left(-S\left[f_{\text{train}},u\right]-Q\left[\ell,f_{\text{train}},u\right]\right)
 \label{eq:MGF}
\end{align}
\begin{align}
 S\left[f_{\text{train}},u\right]= & \frac{1}{2}\intop_{0}^{\infty}dt\intop_{0}^{\infty}dt^{\prime}m\left(t,t^{\prime}\right)u^{\top}\left(t\right)K^{L}\left(t,t^{\prime}\right)u\left(t^{\prime}\right)\label{eq:S}\\
 & +i\intop_{0}^{\infty}dt\left(\intop_{0}^{t}dt^{\prime}\left[K_{d}^{L}\left(t,t^{\prime}\right)\left(Y-f_{\text{train}}\left(t^{\prime}\right)\right)\right]-f_{\text{train}}\left(t\right)\right)^{\top}u\left(t\right)\nonumber 
\end{align}
\begin{align}
 Q\left[\ell,f_{\text{train}},u\right]= & -\sum_{{\bf x}}\intop_{0}^{\infty}dt\intop_{0}^{t}dt^{\prime}\left(k_{d}^{L}\left(t,t^{\prime},{\bf x}\right)\right)^{\top}\left(Y-f_{\text{train}}\left(t'\right)\right)\ell\left(t,{\bf x}\right)\label{eq:Q}\\
 & +i\sum_{{\bf x}}\intop_{0}^{\infty}dt\intop_{0}^{\infty}dt^{\prime}m\left(t,t^{\prime}\right)\left(k^{L}\left(t,t^{\prime},{\bf x}\right)\right)^{\top}u\left(t^{\prime}\right)\ell\left(t,{\bf x}\right)\nonumber \\
 & -\frac{1}{2}\sum_{{\bf x},{\bf x}^{\prime}}\intop_{0}^{\infty}dt\intop_{0}^{\infty}dt^{\prime}m\left(t,t^{\prime}\right)\mathcal{K}^{L}\left(t,t^{\prime},{\bf x},{\bf x}^{\prime}\right)\ell\left(t,{\bf x}\right)\ell\left(t^{\prime},{\bf x}^{\prime}\right)\nonumber 
\end{align}
\twocolumngrid 
where $Df_\text{train},Du$ stands for a summation over all possible trajectories of the time-dependent vectors $f_\text{train}(t),u(t)$. Thus, the MGF defines a Gaussian measure on  $f_{\text{train}}(t)$ and $u(t)$. $S\left[f_{\text{train}},u\right]$ is a functional
represents the source-independent part and is related to the dynamics
of the predictor on the training data, while $Q\left[\ell,f_{\text{train}},u\right]$ is a functional 
contains the source-dependent part and determines the dynamics of the
predictor on a test point. The scalar coefficient $m(t,t^{\prime})$
is the time-dependent auto-correlations of the weights w.r.t.
the Gaussian prior, Eqs. \ref{eq:prior statistics}-\ref{eq:mtt'}. The
remaining coefficients of the MGF are various two-time kernel functions defined
in the next section.

\section{The Neural Dynamical Kernel \protect\label{subsec:The-neural-dynamical}}

We introduce the definitions of the kernels appearing in Eq.\ref{eq:MGF}, and the relations between these kernels and the known NTK \cite{jacot2018neural} and NNGP \cite{lee2017deep} kernels. Henceforth, for all kernel types, we denote by $\mathcal{K}({\bf x},{\bf x}')$ the kernel function of two inputs, and by $K$ the kernel matrix obtained by applying the kernel function over all pairs of data points, such that $K_{\mu\nu}=\mathcal{K}\left({\bf x}_{\mu},{\bf x}_{\nu}\right)$, where $({\bf x}_{\mu},{\bf x}_{\nu})$ are a pair of training input
vectors. 
The quantity $K^{L}(t,t')$ appearing in Eq.\ref{eq:MGF} is a $P\times P$
matrix $K_{\mu\nu}^{L}(t,t^{\prime})=\mathcal{K}^{L}\left(t,t^{\prime},{\bf x}_{\mu},{\bf x}_{\nu}\right)$
 and $\mathcal{K}^{l}\left(t,t^{\prime},{\bf x},{\bf x}^{\prime}\right)$
is defined for any two inputs ${\bf x},{\bf x}^{\prime}$
and any layer $l$ as
\begin{equation}
\mathcal{K}^{l}\left(t,t^{\prime},{\bf x},{\bf x}^{\prime}\right)=\frac{1}{N_{l}}\left\langle {\bf x}_{t}^{l}\left({\bf x}\right)\cdot{\bf x}_{t^{\prime}}^{l}\left({\bf x}^{\prime}\right)\right\rangle _{0}
\end{equation}
The integer $N_{l}$ is the width of the $l$-th layer and the average
is w.r.t. to the prior statistics (Eq.\ref{eq:prior statistics}).
At equal times $t=t'$, the kernel function is equivalent to the usual NNGP kernel function, as the average is over Gaussian parameters with time-dependent variance $\mathcal{N}\sim(0, m(t,t))$ (see Eq.\ref{eq:mtt'}) which transitions between $\sigma_{0}^{2}$
at initialization to $\sigma^{2}$ at long times. 

The quantity $K^{L}_{d}(t,t')$ appearing in Eq.\ref{eq:MGF}, is
a $P\times P$ matrix, $K_{d, \mu\nu}^{L}\left(t,t^{\prime}\right)=\mathcal{K}^{L}_{d}\left(t,t^{\prime},{\bf x}_{\mu},{\bf x}_{\nu}\right)$. This matrix is defined via a novel Neural Dynamical Kernel
(NDK) function of any two input vectors, as follows
\begin{align}
 & \mathcal{K}^{L}_{d}\left(t,t^{\prime},{\bf x},{\bf x}^{\prime}\right)=\label{eq:interp}\\
 & e^{-T\sigma^{-2}\left|t-t^{\prime}\right|}\left\langle \nabla_{\Theta}f(t,{\bf x})\cdot\nabla_{\Theta}f(t^{\prime},{\bf x}^{\prime})\right\rangle _{0}\nonumber 
\end{align}
where $\nabla_{\Theta}f(t,{\bf x})\equiv\nabla_{\Theta}f(\Theta,{\bf x})|_{\Theta_{t}}$ where
$\Theta_{t}$ obeys the stochastic statistics given in Eqs.\ref{eq:prior statistics},\ref{eq:mtt'}. 
At equal times: $t=t'$, this kernel has a simple relation to the NTK \cite{jacot2018neural}. Specifically,

\begin{equation}
\mathcal{K}_{d}^{L}\left(t=t^{\prime},{\bf x},{\bf x}^{\prime}\right)=\mathcal{K}_{NTK}^{L}\left({\bf x},{\bf x}^{\prime}\right)_{\mathcal{N}\sim(0,m(t,t))}\label{eq:equal time}
\end{equation}

 In particular, at initialization -  $\mathcal{K}^{L}_{d}\left(t=0,t^{\prime}=0,{\bf x},{\bf x}^{\prime}\right)=\mathcal{K}_{NTK}^{L}\left({\bf x},{\bf x}^{\prime}\right)$, where the average is only on the weights random initial condition, like in the usual NTK. Furthermore, the NNGP kernel can also be evaluated from
the NDK by an integral over long times (see SI Sec.\ref{subsec:SI The-neural-dynamical}
for detailed proof)
\begin{equation}
\lim_{t\rightarrow\infty}\left(\frac{T}{\sigma^{2}}\intop_{0}^{t}\mathcal{K}^{L}_{d}\left(t,t^{\prime},{\bf x},{\bf x}^{\prime}\right)dt^{\prime}\right)=\mathcal{K}_{GP}^{L}\left({\bf x},{\bf x}^{\prime}\right)\label{eq:kd and nngp}
\end{equation}
where $\mathcal{K}_{GP}^{l}\left({\bf x},{\bf x}^{\prime}\right)=\frac{1}{N_{l}}\left\langle {\bf x}^{l}\left({\bf x}\right)\cdot{\bf x}^{l}\left({\bf x}^{\prime}\right)\right\rangle _{\mathcal{N}\sim(0,\sigma^{2})}$. This identity is important for reaching the Bayesian
equilibrium at long times (see Sec.\ref{subsec:Long-time-equilibrium}) and is related to the well-known relation between correlations and response functions, the Fluctuation Dissipation Theorem (FDT) in statistical mechanics \cite{kubo1966fluctuation}. 
The NDK can be obtained recursively in terms of the time-dependent kernel $\mathcal{K}^{l}\left(t,t^{\prime},{\bf x},{\bf x}^{\prime}\right)$
and the derivative kernel $\dot{\mathcal{K}}^{l}\left(t,t^{\prime},{\bf x},{\bf x}^{\prime}\right)$,
similar to the NTK (see SI Sec.\ref{subsec:SI The-neural-dynamical} for detailed proof).

\begin{align}
& \mathcal{K}^{L}_{d}\left(t,t^{\prime},{\bf x},{\bf x}^{\prime}\right)=\label{eq:recursive kd} \\ &  m\left(t,t^{\prime}\right)\dot{\mathcal{K}}^{L}\left(t,t^{\prime},{\bf x},{\bf x}^{\prime}\right)\mathcal{K}_{d}^{L-1}\left(t,t^{\prime},{\bf x},{\bf x}^{\prime}\right)\nonumber\\
 & +e^{-T\sigma^{-2}\left|t-t^{\prime}\right|}\mathcal{K}^{L}\left(t,t^{\prime},{\bf x},{\bf x}^{\prime}\right)\nonumber 
\end{align}
\begin{equation}
\mathcal{K}_{d}^{l=0}\left(t,t^{\prime},{\bf x},{\bf x}^{\prime}\right)=e^{-T\sigma^{-2}\left|t-t^{\prime}\right|}\left(\frac{1}{N_{0}}{\bf x}\cdot{\bf x}^{\prime}\right)
\end{equation}
 The derivative kernel, $\dot{\mathcal{K}}^{l}\left(t,t^{\prime},{\bf x},{\bf x}^{\prime}\right)$
appearing in Eq.\ref{eq:recursive kd} is the dot product of the derivative of
the activation function

\begin{equation}
\dot{\mathcal{K}}^{l}\left(t,t^{\prime},{\bf x},{\bf x}^{\prime}\right)=\frac{1}{N_{l}}\left\langle \phi^{\prime}\left({\bf z}_{t}^{l}({\bf x})\right)\cdot\phi^{\prime}\left({\bf z}_{t^{\prime}}^{l}({\bf x}^{\prime})\right)\right\rangle _{0}
\end{equation}
where $\phi^{\prime}$ stands for elementwise derivative of the activation
of the layer $l$ w.r.t to their preactivations ($\frac{\partial\phi({\bf z})}{\partial{\bf z}}$),
induced by a given input at time $t$.  The time-dependent kernels $\mathcal{K}^{L}$ and $\dot{\mathcal{K}}^{L}$
obey recursion relations similar to their static counterparts (see
SI Sec.\ref{subsec:SI The-neural-dynamical}). These recursion relations - 
as well as those of Eq.\ref{eq:recursive kd} - have a closed-form expression for some
nonlinearities such as ReLU and error function (inspired by the static
expressions for these kernels \cite{cho2009kernel,williams1996computing}), and explicit solutions for linear activation (see SI Sec.\ref{subsec:SI The-neural-dynamical}). Finally, in Eqs.\ref{eq:S},\ref{eq:Q}, $k_{\mu}^{L}(t,t^{\prime},{\bf x})\in\mathbb{R}^{P\times1}$
and $k^{L}_{d}\left(t,t^{\prime},{\bf x}\right)\in\mathbb{R}^{P\times1}$
are vectors of the kernels of a test point with the training data,
such that $k_{\mu}^{L}\left(t,t^{\prime},{\bf x}\right)=\mathcal{K}^{L}\left(t,t^{\prime},{\bf x},{\bf x}_{\mu}\right),$
and similarly $k_{d,\mu}^{L}\left(t,t^{\prime},{\bf x}\right)=\mathcal{K}^{L}_{d}\left(t,t^{\prime},{\bf x},{\bf x}_{\mu}\right)$. For the convenience of the reader we summarize  all the relevant kernel functions in the following table:

\begin{figure}[hbt!]
    \centering
    \includegraphics[width=1.05\linewidth]{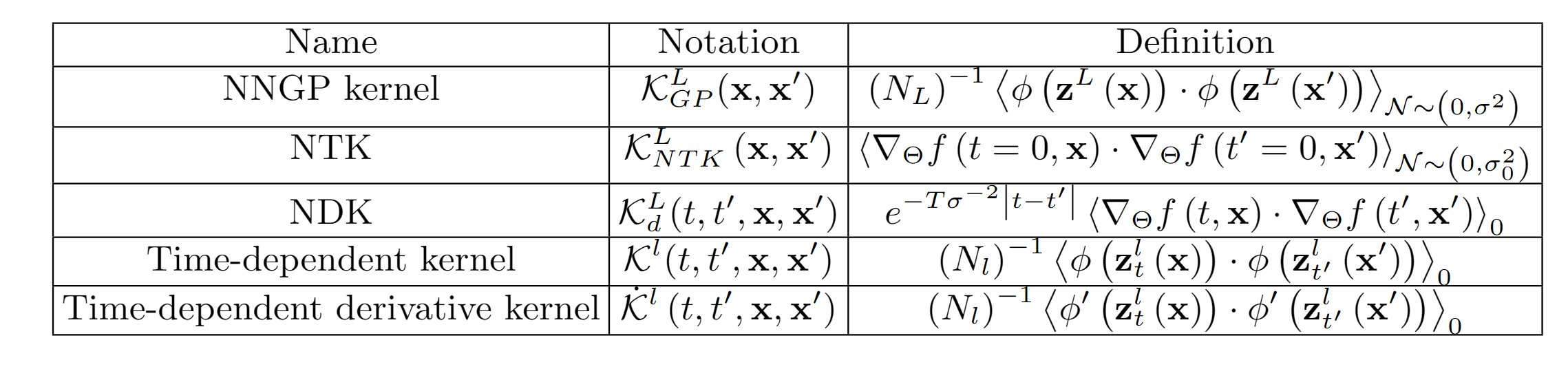}
    \label{fig:table}
\end{figure}
\begin{figure*}[hbt!]
\centering{}\includegraphics[width=0.9\textwidth]{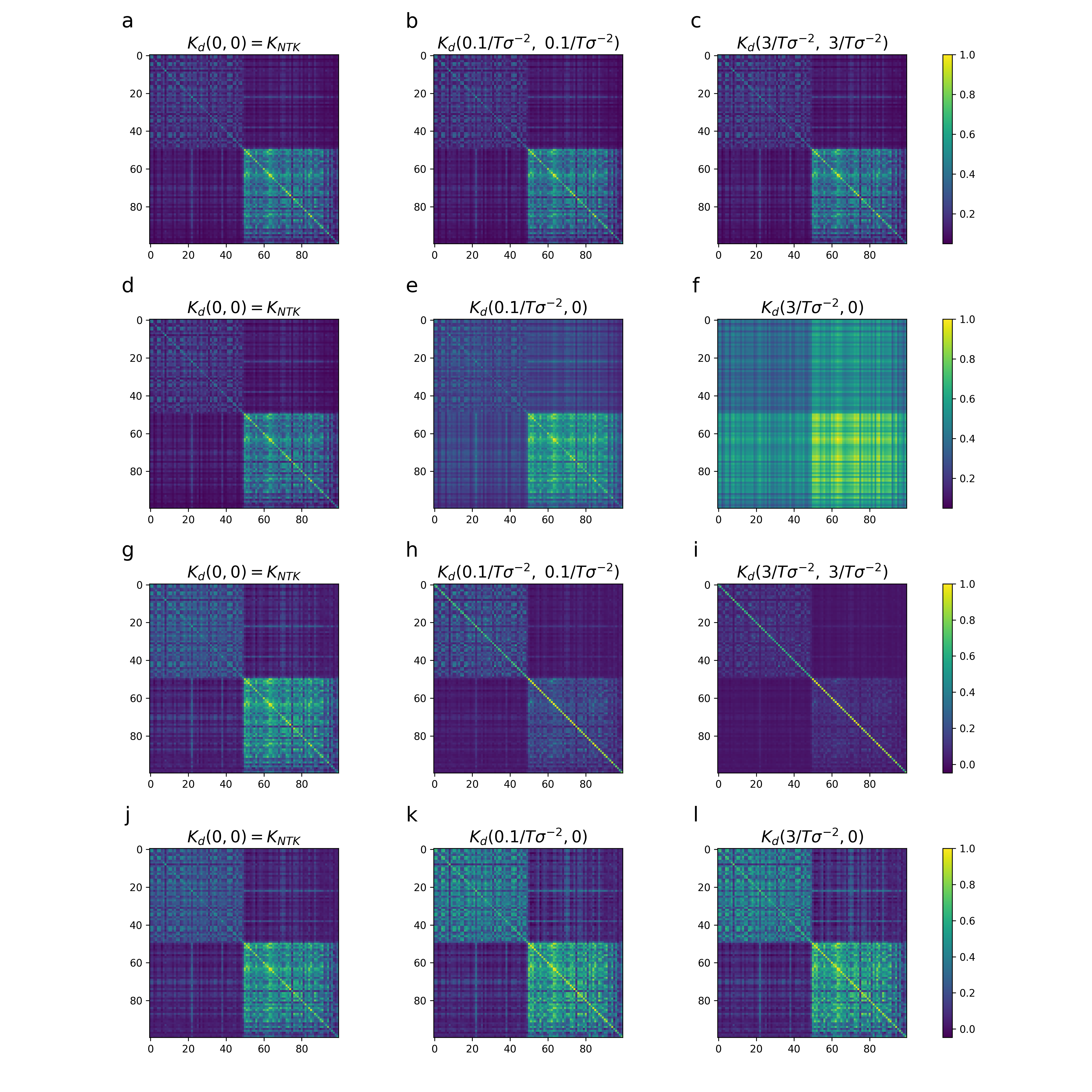}
\caption{The Neural Dynamical Kernel (NDK): The figure presents the NDK for various nonlinearities, parameters, and times, using examples from the MNIST dataset (0,1 digits). To focus on the kernel's structure independently of scale, the kernels are normalized by their maximum value. We present the NDK for equatl times ( $t=t'$, Eq.\ref{eq:equal time}) and for time difference from initialization (evaluating  Eq.\ref{eq:recursive kd} at $t'=0$)
(a-c) ReLU kernel (equal times) with parameters $\sigma_0=0.2$ and $\sigma=1$. Since ReLU is a homogeneous function, changes in the variance of the distribution do not alter the kernel's structure, which is preserved for all times.
(d-f) ReLU kernel (time difference): The ReLU kernel depends on the angles between pairs of input vectors. As the time difference increases, the representations decouple, leaving only information about the amplitude of each example, $\left\Vert {\bf x}_{\mu}\right\Vert$, which is reflected in the rows and columns of the kernel. This uncorrelated kernel is critical for understanding representational drift (see Sec.\ref{sec:representational drift})
(g-i) Error Function kernel (equal times) with parameters $\sigma_0=0.2$ and $\sigma=10$. For small $\sigma_0$, the kernel resembles a linear kernel, closely reflecting the structure of the input. A large  $\sigma$ causes a step-function-like behavior of the kernel, with a strong peak along the diagonal.
(j-l) Error Function kernel (time difference) with parameters $\sigma_0=0.2$ and $\sigma=10$. The effect of time difference is similar to that of small variance, resulting in a kernel that resembles a linear kernel. 
\label{fig: kernels}}
\end{figure*}

\textbf{ReLU Dynamical Kernel:} Due to its homogeneity property, i.e. $\text{ReLU}(\lambda x)=\lambda \cdot\text{ReLU}(x)$, altering the variance of the weight distribution changes the global scale of the kernel but does not change its structural properties. This implies that the equal-time kernel in ReLU preserves its structure, encoding the underlying data correlations (see Fig.~\ref{fig: kernels} (a-c)). 

We analyze the effect of time differences on the ReLU NDK. The ReLU kernel depends on the angle between two inputs, defined as $\theta(\bf{x},\bf{x^\prime})=\frac{\bf{x}\cdot\bf{x^\prime}}{\left\Vert {\bf x}\right\Vert \left\Vert {\bf x}^{\prime}\right\Vert }$ . When the time difference is large, the inputs become uncorrelated due to random fluctuations in the weights. As a result, all angles between different inputs approach $\theta_{\mu\nu}\rightarrow(1-\delta_{\mu\nu})\frac{\pi}{2}$, leading to a loss of structure in the kernel (see Fig.~\ref{fig: kernels} (d-f)). In this scenario, the only information retained is the amplitudes of the inputs, $\left\Vert {\bf x}\right\Vert$ and $\left\Vert{\bf x}^{\prime}\right\Vert$. This behavior is critical for understanding representational drift, as we discuss in Sec.\ref{sec:representational drift}.

\textbf{Sigmoidal Dynamical Kernel:} Sigmoidal functions are defined as monotonically increasing functions with linear behavior around zero and saturation at both $\pm\infty$. Examples of such functions include hyperbolic tangent and error function. These functions are sensitive to changes in the variance of the weights. When the variance is small, the kernel closely resembles a linear kernel. On the other hand, when the variance is large, the function acts like a step function, leading to strong non-linearity. In the NDK, this nonlinearity is most prominent along the diagonal due to the influence of the derivative kernel, as shown in Fig.~\ref{fig: kernels} (g-i). 

In sigmoidal functions, significant time differences result in a kernel that is nearly linear. Unlike the ReLU kernel, the structure of the kernel is preserved, with the primary effect being a change in its scale (see Fig.\ref{fig: kernels}(j-l)).

\textbf{The Mean Predictor:}
The above explicit expression for the MGF allows for the evaluation
of the statistics of the predictor by differentiating the MGF w.r.t. to the source $\ell(t,{\bf x})$. The equations describing the second moment for a general nonlinearity are complex, and given in SI Sec.\ref{sec: variance}. Here we bring the equations for the mean predictor for train and test.

The mean predictor on the training
inputs obeys the following integral equation 
\begin{equation}
\left\langle f_{\text{train}}\left(t\right)\right\rangle =\intop_{0}^{t}dt^{\prime}K^{L}_{d}\left(t,t^{\prime}\right)\left(Y-\left\langle f_{\text{train}}\left(t^{\prime}\right)\right\rangle \right)\label{eq:meanftrain}
\end{equation}
where the average on $\langle f_{\text{train}}(t)\rangle$ is over all possible trajectories of the parameters, encompassing both the randomness of the noise and the initial condition. The mean predictor on any test point ${\bf x}$ is given by an integral over the training predictor with the NDK of
the test

\begin{equation}
\left\langle f\left(t,{\bf x}\right)\right\rangle =\intop_{0}^{t}dt^{\prime}k^{L}_{d}\left(t,t^{\prime},{\bf x}\right)^{\top}\left(Y-\left\langle f_{\text{train}}\left(t^{\prime}\right)\right\rangle \right)\label{eq:meanf}
\end{equation}

\section{Dynamics at low $T$}

As discussed in Sec.\ref{subsec:two learning phases}, in the limit of $T \rightarrow 0$, the learning dynamics can be divided into two distinct phases: a gradient-driven phase, occurring on a timescale of $\mathcal{O}(1)$ and dominated by deterministic minimization of the SE loss, and a diffusive phase, during which the weights explore the solution subspace, and characterized by time scale of $t\sim\mathcal{O}(1/T)$.
\subsection{Gradient-Driven Phase Corresponding to NTK Dynamics\protect\label{subsec:Gradient-driven-phase-correspond}}

The time dependence of the NDK (Eq.$\text{\ref{eq:recursive kd}}$) comes from
exponents with time scale of $\sigma^{2}/T$ (Eqs.$\text{\ref{eq:mtt'}, \ref{eq:recursive kd}}$),
and thus, in the limit of $T\rightarrow0$ and $t\sim\mathcal{O}\left(1\right)$,
we can substitute $\mathcal{K}^{L}_{d}\left(t,t^{\prime},{\bf x},{\bf x}^{\prime}\right)=\mathcal{K}^{L}_{d}\left(t=0,t^{\prime}=0,{\bf x}\text{,}{\bf x}^{\prime}\right)=\mathcal{K}_{NTK}^{L}\left({\bf x}\text{,}{\bf x}^{\prime}\right)$
in Eq. \ref{eq:meanftrain}.

\begin{figure*}[hbt!]
\centering{}\includegraphics[width=0.9\textwidth]{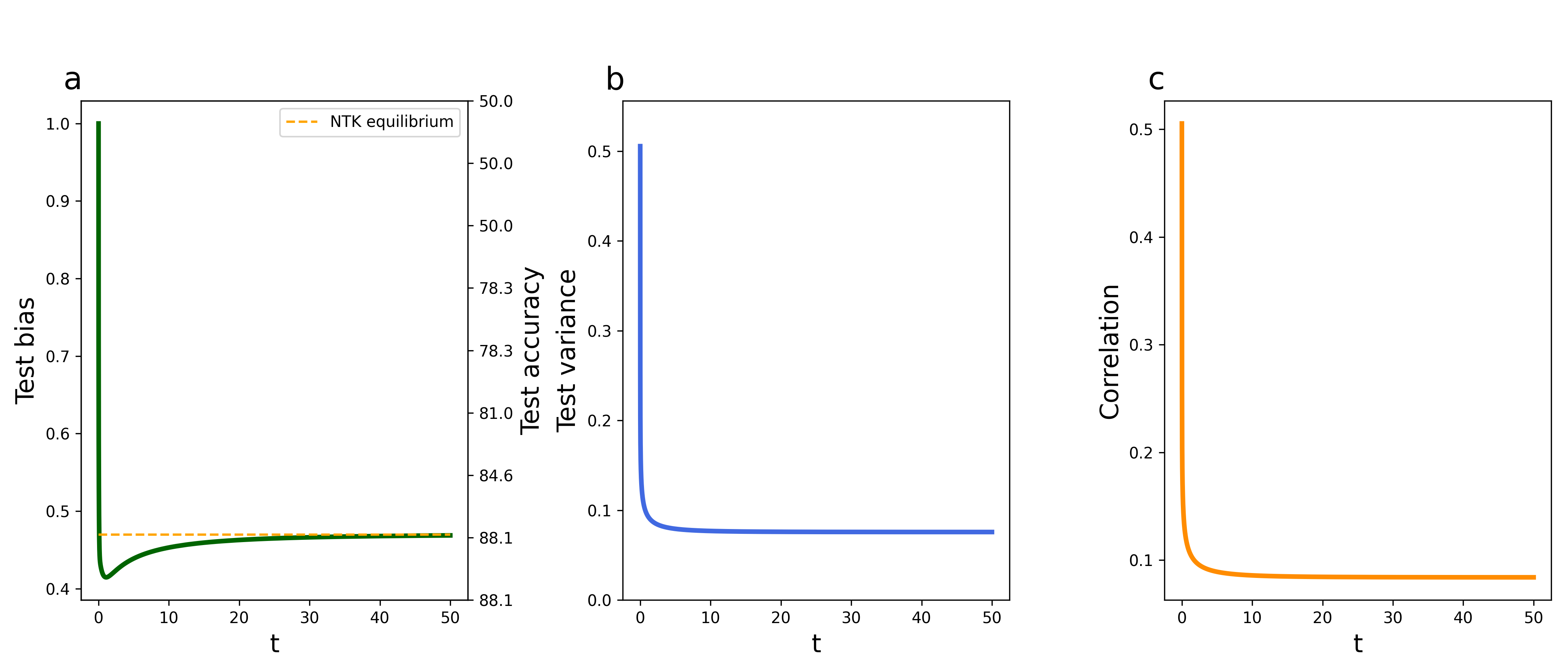}
\caption{Gradient Driven Phase: NTK theory for a ReLU deep network with one hidden layer. The network is trained on binary classification in CIFAR-10. (a) The dynamics of the test bias, defined as $\left(\left\langle f\left({\bf x},t\right)\right\rangle -Y\right)^{2}$, averaged over the test dataset, show convergence to the NTK equilibrium. (b) The variance of the predictor, $\left\langle \delta f\left(t,{\bf x}\right)\delta f\left(t,{\bf x}\right)\right\rangle$, averaged over the test dataset, decreases with learning to an equilibrium value (Eq.\ref{eq:ntk eq var}). (c) The correlation with the initial condition, $\left\langle \delta f\left(t,{\bf x}\right)\delta f\left({\bf x},0\right)\right\rangle$, do not vanish in the NTK equilibrium at long times but rather go to an equilibrium value (Eq.\ref{eq:ntk eq corr}).  
This implies that long-term generalization depends on the random initialization of weights in deterministic gradient descent process.}
\label{fig:ntk}
\end{figure*}

\textbf{Mean predictor:} Differentiating Eq.\ref{eq:meanftrain} reduces the integral equation into the following linear ODE

\begin{align}
& \frac{d}{dt}\left\langle f_{\text{train}}(t)\right\rangle =K_{NTK}^{L}\left(Y-\left\langle f_{\text{train}}(t)\right\rangle \right)\label{eq:meanftrainntk} \\ &  \,\left\langle f_{\text{train}}(t=0)\right\rangle =0\nonumber
\end{align}
which yields $\lim_{T\rightarrow0}\left\langle f_{\text{train}}\left(t\right)\right\rangle =\left(I-\exp\left(-K_{NTK}^{L}t\right)\right)Y$,
and the well-known mean predictor in the NTK theory \cite{jacot2018neural} (see Fig.\ref{fig:ntk}(a)): 
\begin{align}
& \lim_{T\rightarrow0}\left\langle f\left(t,{\bf x}\right)\right\rangle = \label{eq:ntkdynamics} \\ & k_{NTK}^{L}\left({\bf x}\right)^{\top}\left(K_{NTK}^{L}\right)^{-1}\left(I-\exp\left(-K_{NTK}^{L}t\right)\right)Y\nonumber
\end{align}
We define $K_{NTK}^{L}\in\mathbb{R}^{P\times P}$  and $k_{NTK}^{L}\left({\bf x}\right)\in\mathbb{R}^{P}$
as the NTK applied on the
train and test data respectively, similar to Sec.$\text{\ref{subsec:The-neural-dynamical}}$.

In this regime both the Langevin noise and the $L_{2}$ regularizer can be neglected, resulting in dynamics that can be approximated by gradient descent, as assumed in the NTK theory.
In particular, the ``NTK equilibrium'' defined by first taking $T\to 0$ and then $t\to\infty$, yields
the well-known static NTK result
\begin{equation}
\lim_{t\rightarrow\infty}\lim_{T\rightarrow0}\langle f\left(t,{\bf x})\right\rangle=k_{NTK}^{L}\left({\bf x}\right)^{\top}\left(K_{NTK}^{L}\right)^{-1}Y
\end{equation}
The NTK equilibrium signifies the transition between the gradient-driven
phase and the diffusive learning phase, after which the effect of the Langevin noise and the
$L_{2}$ regularizer cannot be neglected.

\textbf{Predictor covariance: } We present the results of the covariance of the predictor in the NTK theory, achieved by taking the limit $T\rightarrow0$ of the expressions of the second moment in our theory (see SI Sec.\ref{sec: variance} for details). For the predictor of training points
\begin{align}
 \lim_{T\rightarrow0}&\space\sigma_{0}^{-2}\left\langle \delta f_{\text{train}}\left(t\right)\delta f_{\text{train}}^{\top}\left(t^{\prime}\right)\right\rangle = \\ & \exp\left(-K_{NTK}^{L}t\right)K_{GP_{0}}^{L}\exp\left(-K_{NTK}^{L}t^{\prime}\right)\nonumber
\end{align}
We denote $K_{GP_{0}}$ as $K_{GP}^{L}\left(\sigma=\sigma_{0}\right)$,
where the statistics of the kernel are over the Gaussian initialization
with $\sigma_{0}$ standard deviation, and not the Gaussian prior. The variance
on the training data vanishes at long times, as all the outputs
converge to their target labels. The covariance of the test
inputs (see Fig.\ref{fig:ntk}(b)) is
\onecolumngrid
\begin{align}
& \lim_{T\rightarrow0}\space\sigma_{0}^{-2}\left\langle \delta f\left(t,{\bf x}\right)\delta f\left(t^{\prime},{\bf x}^{\prime}\right)\right\rangle = \mathcal{K}_{GP_{0}}^{L}\left({\bf x},{\bf x^{\prime}}\right)-k_{GP_{0}}^{L}\left({\bf x}\right)(K_{GP_{0}}^{L})^{-1}k_{GP_{0}}^{L}\left({\bf x^{\prime}}\right)\\ & 
+\left[\left(I-\exp\left(-K_{NTK}^{L}t\right)\right)(K_{NTK}^{L})^{-1}k_{NTK}^{L}\left({\bf x}\right)-(K_{GP_{0}}^{L})^{-1}k_{GP_{0}}^{L}\left({\bf x}\right)\right]^{\top}K_{GP_{0}}^{L}\nonumber
\\ & \times\left[\left(I-\exp\left(-K_{NTK}^{L}t^{\prime}\right)\right)(K_{NTK}^{L})^{-1}k_{NTK}^{L}\left({\bf x}^{\prime}\right)-(K_{GP_{0}}^{L})^{-1}k_{GP_{0}}^{L}\left({\bf x^{\prime}}\right)\right]\nonumber\\
 & \nonumber
 \end{align}
 We note that at initialization, the result $\left\langle \delta f\left(0,{\bf x}\right)\delta f\left(0,{\bf x'}\right)\right\rangle=\sigma_{0}^{2}\mathcal{K}_{GP_{0}}({\bf x},{\bf x}')$ is independent of the limit $T\rightarrow0$ and it is true for any temperature (as can be seen in SI Sec.\ref{sec: variance}).
Taking the limit of $t,t'\rightarrow\infty$ yields
\begin{align}
& \lim_{t,t^{\prime}\rightarrow\infty}\lim_{T\rightarrow0}\space\sigma_{0}^{-2}\left\langle \delta f\left(t,{\bf x}\right)\delta f\left(t^{\prime},{\bf x}^{\prime}\right)\right\rangle = \mathcal{K}_{GP_{0}}^{L}\left({\bf x},{\bf x^{\prime}}\right)-k_{GP_{0}}^{L}\left({\bf x}\right)(K_{GP_{0}}^{L})^{-1}k_{GP_{0}}^{L}\left({\bf x^{\prime}}\right) \label{eq:ntk eq var}\\ &
+\left[(K_{NTK}^{L})^{-1}k_{NTK}^{L}\left({\bf x}\right)-(K_{GP_{0}}^{L})^{-1}k_{GP_{0}}^{L}\left({\bf x}\right)\right]^{\top}K_{GP_{0}}^{L}\left[(K_{NTK}^{L})^{-1}k_{NTK}^{L}\left({\bf x}^{\prime}\right)-(K_{GP_{0}}^{L})^{-1}k_{GP_{0}}^{L}\left({\bf x^{\prime}}\right)\right]\nonumber
 \end{align}
Finally, the correlation between the predictor at time $t$ and $t'=0$ converges to the non-zero values (see Fig.\ref{fig:ntk}(c)).

\begin{align}
\lim_{t\rightarrow\infty}\lim_{T\rightarrow0}\space\sigma_{0}^{-2}\left\langle \delta f\left(t,{\bf x}\right)\delta f\left(0,{\bf x}^{\prime}\right)\right\rangle = \mathcal{K}_{GP_{0}}^{L}\left({\bf x},{\bf x^{\prime}}\right)-k_{NTK}^{L}\left({\bf x}\right)^{\top}(K_{NTK}^{L})^{-1}k_{GP_{0}}^{L}\left({\bf x^{\prime}}\right)
\label{eq:ntk eq corr}\end{align}
The fact that the correlation does not vanish signifies the strong dependence between the generalization and the random initial condition in deterministic gradient descent. 
\twocolumngrid

\begin{figure*}[hbt!]
\centering{}\includegraphics[width=0.8\textwidth]{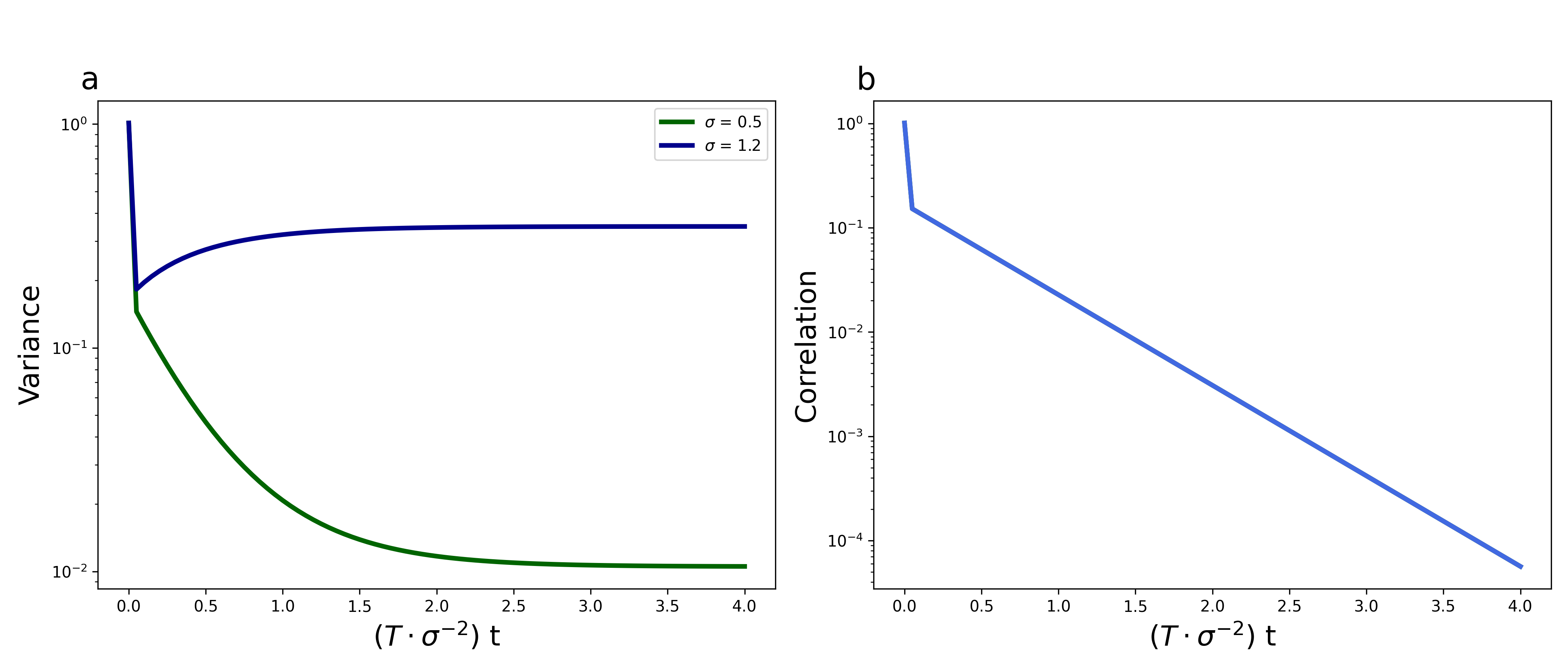}
\caption{Predictor Covariance in Linear Network: The theory of the predictor's covariance in linear network during the diffusive phase (a) The variance of the predictor, $\left\langle \delta f\left(t,{\bf x}\right)\delta f\left(t,{\bf x}\right)\right\rangle$, averaged on the test dataset, is shown for two values of $\sigma$ with $\sigma_0=1$. For $\sigma < \sigma_0$, the variance decreases during the diffusive learning phase due to the additional constraints imposed by L2 regularization. For $\sigma > \sigma_0$, the variance increases as the network explores the solution subspace. (b) The correlation with the initial condition, $\left\langle \delta f\left(t,{\bf x}\right)\delta f\left({\bf x},0\right)\right\rangle$. A rapid decrease during the gradient-driven phase followed by an exponential decay in the diffusive learning phase, reflecting the decorrelation caused by random changes in the weights.
}\label{fig:linear var}

\end{figure*}
\subsection{Diffusive Dynamics\protect\label{subsec:Long-time-equilibrium}}
\hspace{-1mm}
The diffusive regime is characterized by $t\sim\mathcal{O}\left(1/T\right)$.
The SE loss at this stage is $\mathcal{O}(T)$ as the network explores the solution subspace. The mean field equations (Eq.\ref{eq:meanftrain},
\ref{eq:meanf}) in this regime do not admit an analytical solution and
have been solved numerically, with an exception of linear networks, where they are tractable. The numerical solutions are presented in Sec.\ref{sec:Numerical-Evaluations-of}.
In linear networks, the NTK and the NNGP kernels are the same up to a constant, causing the mean predictor to remain fixed at the NTK equilibrium value (to leading order in $T$). 
The predictor covariance for linear networks takes the following simple form, at low temperatures and times of $\mathcal{O}(1/T)$:
\begin{align}
&\left\langle \delta f\left(t,{\bf x}\right)\delta f\left(t^{\prime},{\bf x}^{\prime}\right)\right\rangle =\label{eq:main text linear var} \\ &m^{L+1}\left(t,t^{\prime}\right)\left[\mathcal{K}_{in}\left({\bf x},{\bf x}^{\prime}\right) - k_{in}\left({\bf x}\right)^{\top}\left(K_{in}\right)^{-1}k_{in}\left({\bf x}^{\prime}\right)\right] \nonumber
\end{align}

where $\mathcal{K}_{in}\left({\bf x},{\bf x^{\prime}}\right)=\frac{1}{N_{0}}{\bf x}\cdot{\bf x^{\prime}}$ is the input's covariance matrix. Similar to Sec.\ref{subsec:The-neural-dynamical} we define the $P\times P$ matrix $\left(K_{in}\right)_{\mu\nu}=\mathcal{K}_{in}\left({\bf x}_{\mu},{\bf x}_{\nu}\right)$ , and the input $P$ dimensional test vector as $\left(k_{in}\left({\bf x}\right)\right)_{\mu}=\mathcal{K}_{in}\left({\bf x},{\bf x}_{\mu}\right)$. 
During the gradient-driven phase, the variance decreases as learning progresses, projecting the initial condition onto the subspace of zero training error. In the diffusive learning phase, the behavior of the variance depends on the ratio between $\sigma$ and $\sigma_0$. When $\sigma < \sigma_0$, the variance is further reduced by the constraints of the $L_2$ regularizer. When $\sigma > \sigma_0$, the variance increases as the weights explore the solution subspace with weaker constraints (see Fig.~\ref{fig:linear var} (a)).

The correlation with the random initialization, $\left\langle \delta f\left(t,{\bf x}\right)\delta f\left(0,{\bf x}\right)\right\rangle$, is shown in Fig.\ref{fig:linear var} (b). In the NTK regime, this correlation remains $\mathcal{O}(1)$ even at long times (see Eq.\ref{eq:ntk eq corr}), due to the strong dependence on the initial condition in the linearized dynamics. However, in the diffusive regime, the temporal correlation decays exponentially with a time scale of $\sigma^2/{T}$ due to the stochastic nature of Langevin dynamics.

Another important limit is the equilibrium limit $t\gg1/T$.
In particular, the mean predictor
reaches a constant value, given by looking for a constant solution to Eq.\ref{eq:meanftrain} at  $t\rightarrow\infty$, which together with Eq.\ref{eq:kd and nngp} yields
\begin{equation}
\lim_{t\rightarrow\infty}\left(\left\langle f_{\text{train}}\left(t\right)\right\rangle \right)=K_{GP}^{L}\left(IT\sigma^{-2}+K_{GP}^{L}\right)^{-1}Y
\label{eq:equilibrium train}\end{equation}
Similarly, for the test mean predictor, 
\begin{equation}
\lim_{t\rightarrow\infty}\left\langle f\left(t,{\bf x}\right)\right\rangle =k_{GP}^{L}\left({\bf x}\right)^{\top}\left(IT\sigma^{-2}+K_{GP}^{L}\right)^{-1}Y
\label{eq:equilibrium test}\end{equation}
which agrees with the well-known equilibrium NNGP result \cite{lee2017deep}.
We emphasize that Eqs.\ref{eq:equilibrium train}-\ref{eq:equilibrium test} hold for any temperature.

\begin{figure*}[hbt!]
\centering
\includegraphics[width=0.8\textwidth]{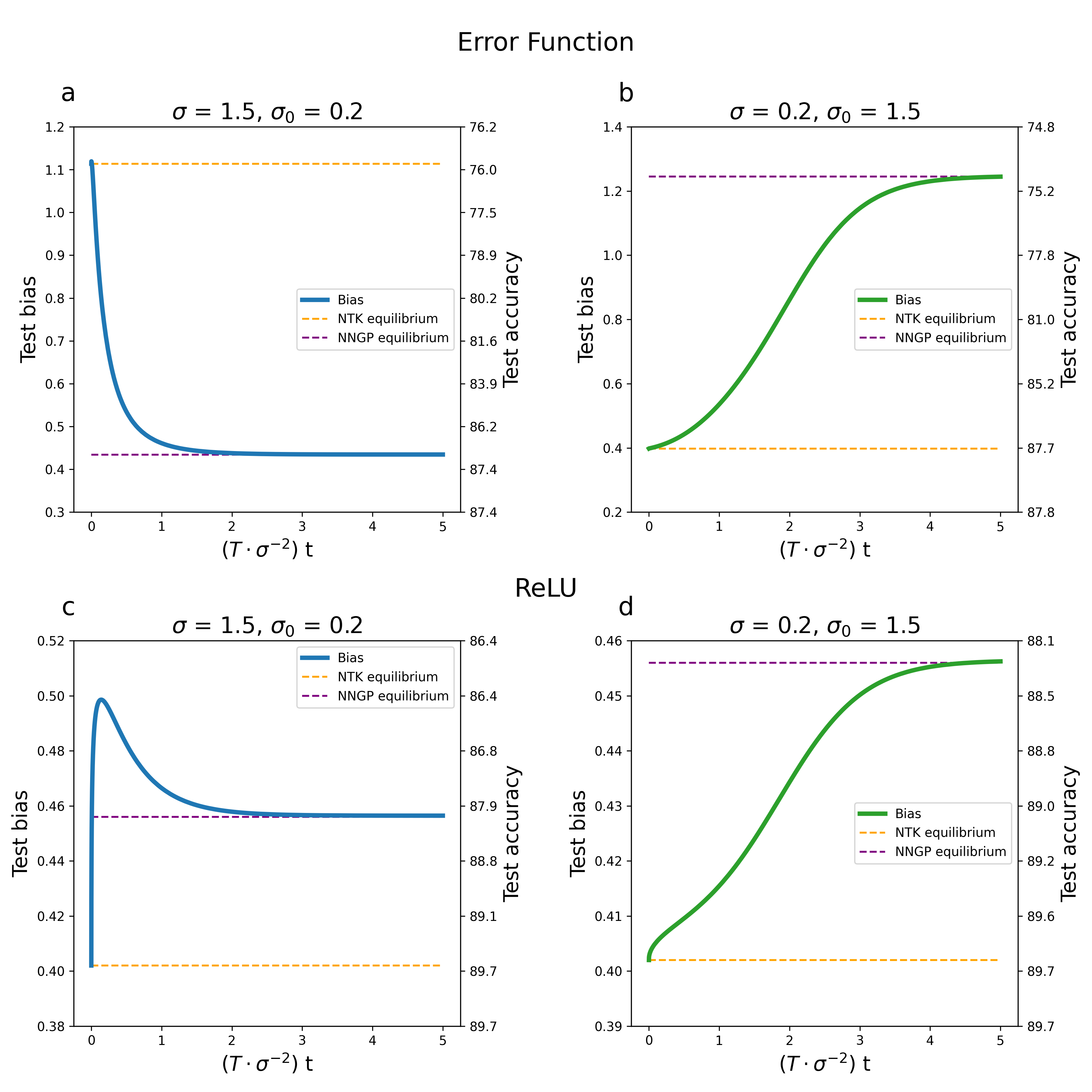}
\caption{
    Diffusive Dynamics: Numerical solutions of the mean field equations (Eqs.\ref{eq:meanftrain}, \ref{eq:meanf}) in the diffusive phase, starting from the NTK equilibrium, which marks the end of the gradient-driven phase. We evaluated the test bias, $\left(\left\langle f\left({\bf x},t\right)\right\rangle -y({\bf x})\right)^{2}$, averaged over the test dataset for various nonlinear functions and parameters in a binary classification task in CIFAR-10.$\quad$ 
    (a-b) Performance comparison with error function activation for different $\sigma, \sigma_0$. Small values cause the sigmoidal function to behave like a linear function, hindering the generalization. Large values lead to strong nonlinearity and improved performance. (a) Small $\sigma_0$ and large $\sigma$. The accuracy after the exploratory diffusive phase is better by 10\% compared to the gradient-driven phase. (b) Large $\sigma_0$ and small $\sigma$. The accuracy after the gradient-driven phase is better by 12\% compared with the equilibrium accuracy (c-d) Comparison of the performance with ReLU activation for different $\sigma, \sigma_0$. The values of the NTK and NNGP equilibria do not depend on the values of $\sigma, \sigma_0$ (d) For $\sigma_0$ and small $\sigma$, the accuracy converges monotonously to the NNGP equilibrium, while in (c) for large $\sigma$ and small $\sigma_0$ the learning is non-monotonic. 
}
\label{fig:accuracy}
\end{figure*}
\section{Numerical Evaluations of the Dynamic Mean Field Equations\protect\label{sec:Numerical-Evaluations-of}}

In this section, we present the numerical solutions of the mean field  Eqs.\ref{eq:meanftrain}-\ref{eq:meanf}, focusing on the mean-predictor in the diffusive phase, where $t \sim \mathcal{O}(1/T)$. 

To analyze the dynamics' dependence on the hyperparameters, we formally take the limit $T \rightarrow 0$ of Eqs.\ref{eq:meanftrain},\ref{eq:meanf}. We present the theoretical predictions for the test bias, defined as $\left(\left\langle f\left({\bf x},t\right)\right\rangle -y({\bf x})\right)^{2}$ and the resultant test accuracy on binary classification task in CIFAR-10 dataset (Fig.\ref{fig:accuracy}). This accuracy can be thought of as taking the majority vote in an ensemble of neural networks trained on the same data. The methodology for solving the equations for low $T$ is detailed in SI Sec.\ref{sec:lowT}.  
As shown in Fig.\ref{fig:accuracy}, the type of nonlinearity plays a significant role in shaping the learning process. 

\hspace{-0.5cm}
In ReLU kernels, varying $\sigma$ and $\sigma_{0}$ serves as a scaling factor without changing their structure. Notably, this indicates that both NTK and NNGP equilibria remain unchanged in ReLU networks regardless of $\sigma$ and $\sigma_{0}$. Consequently, the start and end points of the diffusive learning phase are fixed and affected only by changing the data or the depth of the NN. However, altering the ratio between $\sigma$ and $\sigma_{0}$ can lead to qualitatively different dynamics and non-monotonous behavior, as demonstrated by the comparison of Fig.\ref{fig:accuracy} (c) and (d).

For error functions and other sigmoidal activation functions , the values of $\sigma$ and $\sigma_{0}$ induce a qualitative change in the dynamics, regardless of their ratio. With small $\sigma$ and $\sigma_{0}$, the preactivations are small in magnitude, causing the sigmoidal function to behave almost linearly, which typically results in poor performance. In contrast, larger values of $\sigma$ and $\sigma_{0}$ increase the nonlinearity, often resulting in improved performance. In the binary classification task presented in Fig.~\ref{fig:accuracy}(a-b), the test accuracy has improved by up to 12\% by transitioning from small $\sigma_0$ to large $\sigma$ and vice versa.
\vspace{-10mm}

The difference in average test accuracy between NTK and NNGP is sometimes small in practice \cite{lee2020finite}. We show that when there is a gap between $\sigma$ and $\sigma_0$, the discrepancy can be significant, leading to interesting diffusion learning dynamics. For a more detailed analysis of the equilibria of NTK and NNGP and their dependence on depth and dataset size, see SI Sec.\ref{Sec:NTK and NNGP equilibria}.

\section{Neural Representations and Representational drift}\label{sec:representational drift}

We now explore the implications of diffusive learning dynamics
on the phenomenon of representational drift. Representational drift refers to the
observations that neuronal activity patterns accumulate random changes
over time without noticeable consequences for the relevant animal behavior \cite{rule2019causes,rule2020stable,rule2022self},.
These observations raise fundamental questions about the causal relation
between neuronal representations and the underlying computation.
Here we build on our analysis of the learning dynamics to study the nature of the representations in wide networks and the implications of their drift. 

\subsection{Neuronal Representations\label{subsec:neuronal reps}}

\begin{figure*}[hbt!]
\centering
\includegraphics[width=0.8\textwidth]{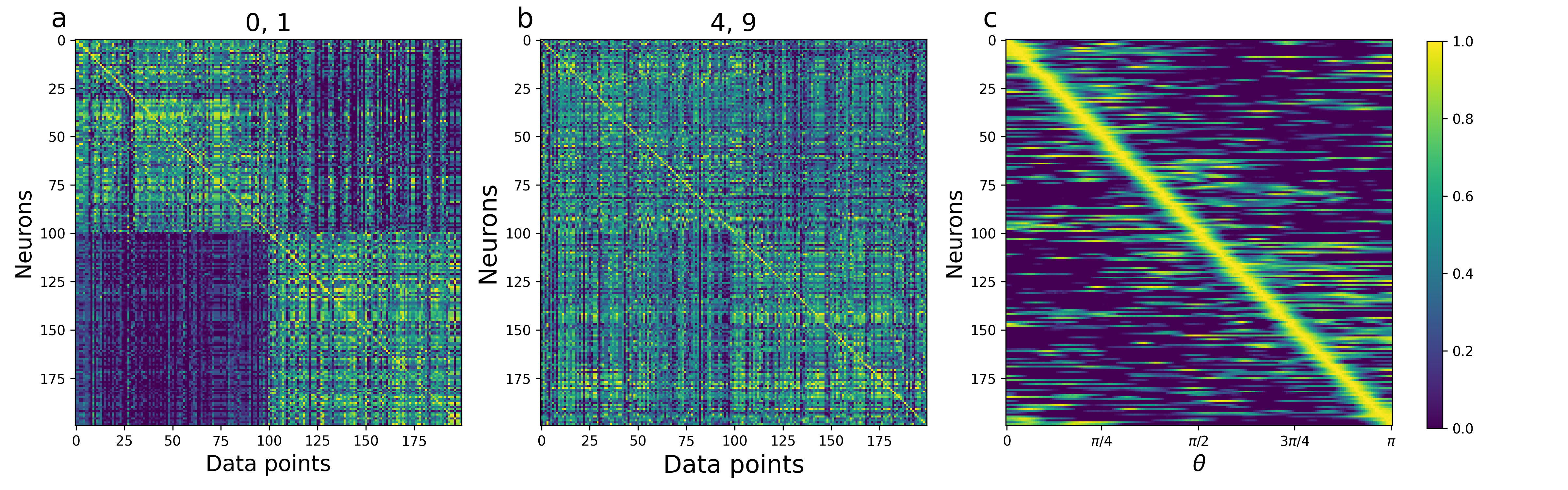}
\caption{Random Representations: The representations of a network with random Gaussian weights are presented for three cases: MNIST binary classificatio with the digits 0,1 (a), the digits 4,9 (b), and (c) for a data governed by a single scalar $\theta$, consists of a sum of harmonics with decaying amplitude (see SI Sec.\ref{sec:Details-of-the} for details). For each data point the neuron with maximum activation was chosen, and the activations presented are normalized by their maximum value. (a) There is no selectivity of a single example, but rather a selectivity for class due to differences in class amplitude (see Sec.\ref{sec:invariant code}).
(b) No clear pattern emerges, and the two categories are indistinguishable.
(c) A clear tuning curve pattern emerges because the data is governed by low-dimensional structure.}
\label{fig:random_representations}
\end{figure*}

\begin{figure*}[hbt!]
\centering{}\includegraphics[width=0.9\textwidth]{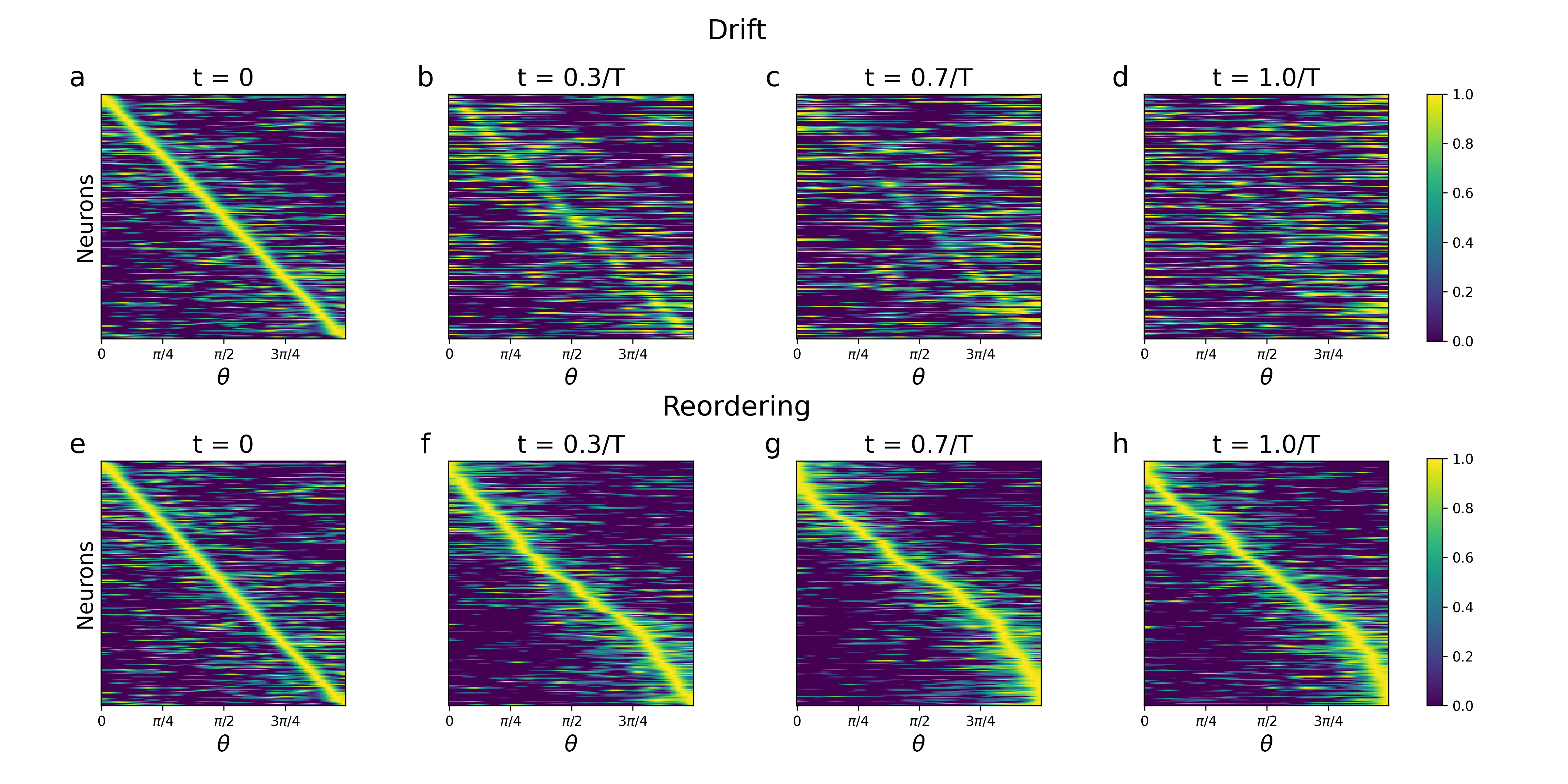}\caption{\label{fig:tuning curve}Tuning Curves and Drift in Random Neural Networks: A deep ReLU network with one hidden layer and a Gaussian initial conditions, with weights drifting according to the prior dynamics (Eq.\ref{eq:prior statistics}), without learning constraints. The input data consists of a sum of harmonics with decaying amplitude (see Sec.\ref{sec:Details-of-the} for details).
(a,e) The tuning curve was constructed by selecting the neuron with maximum activation for each angle $\theta$, and normalized its activation. (a-d) The evolution of neuronal activations as they drift over time. The initial tuning curve gradually fades to a random pattern.
(e-h) By reordering the same neurons during drift, a new tuning function emerges, due to the low dimensional structure of the data. This result resembles findings from experimental neuroscience \cite{rule2019causes}.}
\end{figure*}

{\bf{Random weights}}: Representation in our model refers to the patterns of activity of the neurons at the top hidden layer. In the wide networks studied here (with $N^{-1/2}$ normalization), the hidden layer weights are only slightly modified by learning, resulting in weak feature learning. However, this does not mean that the representation itself is random. As is clear from the non-random structure of the kernels (Fig.\ref{fig: kernels} above), the statistical structure of the inputs is reflected in the properties of the hidden neurons even after filtering by largely random weights. The kernel sums over the properties of all neurons in the layer; hence, even small statistical stimulus selectivity of individual neurons may result in a distinct structure of the kernel. It is, thus, important to examine the tuning properties of individual neurons. 
In fact, as shown in Fig.\ref{fig:random_representations}, in high-dimensional inputs such as MNIST, the selectivity of individual representation neurons to class identity varies across input digits. For instance, it is pronounced for digits 0 and 1 but less so for digits 4 and 9. It is determined by the differences in amplitude for the classes, as will be discussed below (Sec. \ref{sec:invariant code}). In contrast, when inputs are governed by low-dimensional statistical structure (Fig.\ref{fig:tuning curve} (c)), the feature layer exhibits significant single neuron tuning curves similar to those observed in the cortex or hippocampus, even though hidden weights are completely random.  

{\bf{Effect of learning}}: Even in wide networks, hidden weights are not completely random and are affected by learning. This is clear from the derivation of the NDK where both changes in readout weights and hidden layer weights contribute to the kernel structure. Although these changes are small (of order $N^{-1/2}$), they have a task-dependent low-rank structure, and hence they have $\mathcal{O}(1)$ contribution to the predictor. 
Although the exact form of the learned-induced changes in the representation is complicated, elsewhere \cite{li2021statistical,li2022globally, li2024representations} we found that, at equilibrium, the changes in the representations $\phi\left({\bf z}^{l}\left({\bf x}\right)\right)$ can be approximated by
$\phi\left({\bf z}^{l}\left({\bf x}\right)\right)\approx\phi_0\left({\bf x}\right)+\frac{1}{\sqrt{N}}{\bf v}Y^{\top} $
where $\phi_0\left({\bf x}\right)$ are the activations drawn from the prior distribution, ${\bf v}$ is a Gaussian $N_l$ dimensional vector with zero mean and variance $\mathcal{O}(\sigma^{2L})$ and $Y$ is the labels vector, as defined in Sec. \ref{subsec:notations}. 

\hspace{-1cm}
\subsection{Representational Drift}
Consider again the snapshot representation map shown in Fig.\ref{fig:random_representations} above. In Langevin dynamics, these tuning functions will gradually change over time due to the additive noise in the dynamics. Tracking a tuning curve during this process ultimately results in a random pattern (Fig.\ref{fig:tuning curve} (a-d)). However, in any snapshot in time, the population statistics remain the same. Thus, by reordering the neurons, essentially the same tuning map reemegres (Fig.\ref{fig:tuning curve} (e-h)), resembling observations in experimental data. 
It is interesting to note that the complete reordering of the representation also applies to the diffusion dynamics of the learned component of the representation. Specifically, the representational dynamics has approximately the form of

\begin{equation}
\phi({\bf z}^l_t) \approx \phi\left({\bf{W}}_0(t),{\bf x}\right) + N^{-1/2}{\bf v}(t)Y^\top   \label{eq:representation}
\end{equation}
where ${\bf W}_0(t)$ represents the time-dependent hidden layer weights, and the second term represents the low-rank 'feature learning' component. Both ${\bf{W}}_0(t)$ and ${\bf v}(t)\in \mathbb{R}^{N}$ obey
the prior time-dependent statistics (Eq.\ref{eq:prior statistics}) and thus possess temporal correlations that decay exponentially to zero with a time scale of $\sigma^2/T$. 
Accordingly, even the dynamics of the learned features do not break the permutation symmetry and completely reorder the representation over time. 

\begin{figure}[hbt!]
\centering{}\includegraphics[width=0.5\textwidth]{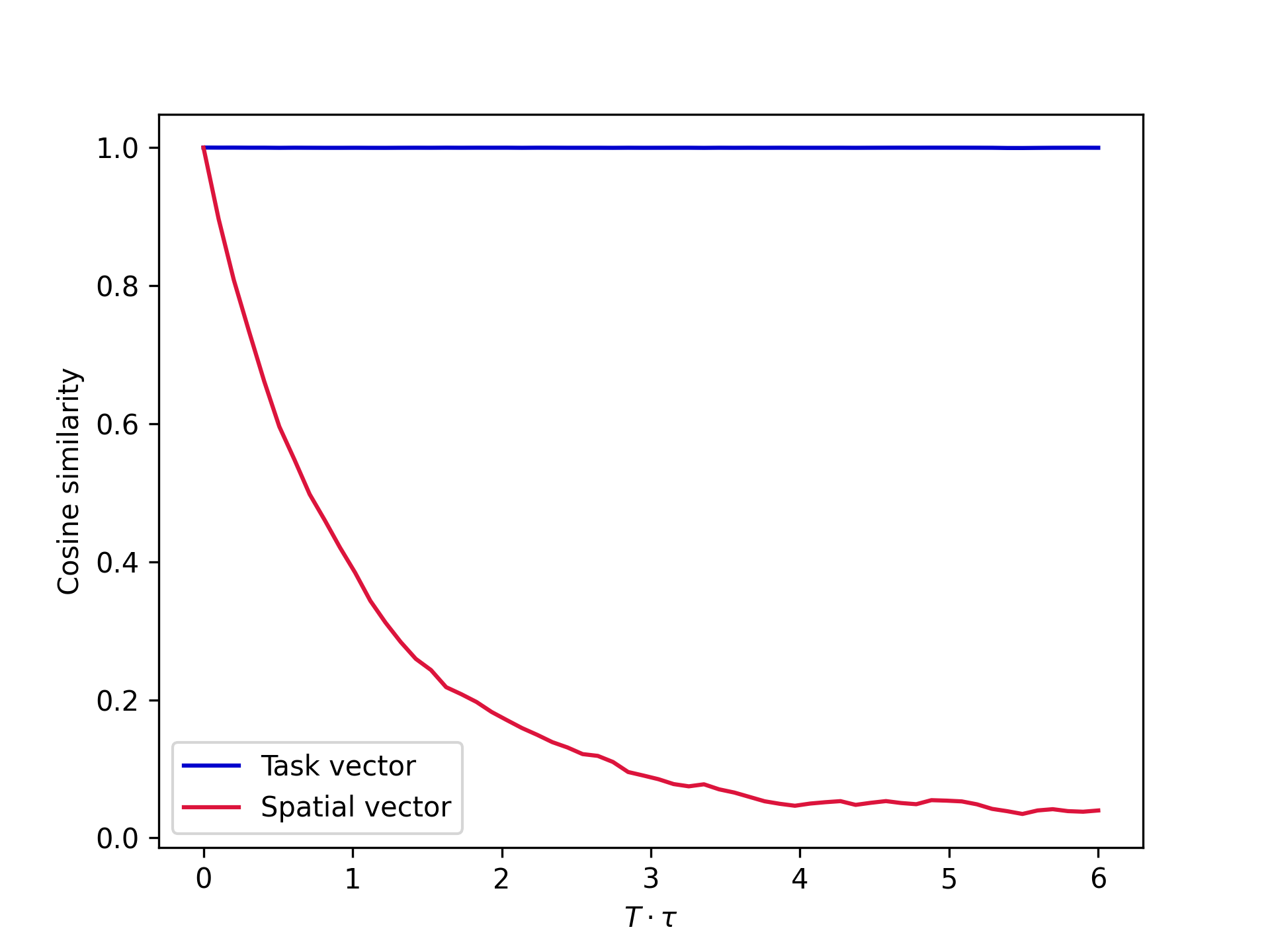}\caption{Low Rank Structure and Drift: A ReLU network with a single hidden layer and a single output, trained on CIFAR-10 binary classification with
Langevin dynamics (Eq.\ref{eq:Langevin}). We analyze the SVD of the representation $\phi\left({\bf z}_{t}^{l}({\bf x})\right)$. We track the cosine similarity of the top right and left singular unit vectors, the P-dimensional task vector $h(t)$ and the N-dimensional spatial vector ${\bf g}(t)$, respectively. The similarity of the $P$-dimensional vector remains stable during the drift process, while the similarity of the $N$-dimensional vector decays exponentially over time with a timescale of $1/T$, which is consistent with the low rank term in Eq.\ref{eq:representation}.}\label{fig:rep drift}
\end{figure}

We have tested the reordering hypothesis by simulating a ReLU network with a single hidden layer and a single output, trained on CIFAR-10 binary classification with Langevin dynamics (Eq.\ref{eq:Langevin}). We track the hidden layer representations on the training data $\phi\left({\bf z}_{t}\left({\bf x}^{\mu}\right)\right)$ during training. 
In order to characterize the drift phenomenon and reveal the low-rank structure, we compute the top right and left singular vectors of $\phi\left({\bf z}_{t}\left({\bf x}^{\mu}\right)\right)$, denoted by $h(t)\in\mathbb{R}^P$ and ${\bf g}(t)\in\mathbb{R}^N$, respectively, and track their cosine similarity after the dynamics reaches equilibrium in the diffusive learning stage. This time-dependent cosine similarity is defined as 
$\rho_h(\tau) \equiv \lim_{t\rightarrow \infty}h(t+\tau)^\top h(t)$, and $\rho_{\bf g}(\tau) \equiv \lim_{t\rightarrow\infty}  {\bf g}(t+\tau)^\top {\bf g}(t)$.
As shown in Fig.\ref{fig:rep drift}, we find that $\rho_h(\tau)$ is stable during drift, while ${\bf g}(t)$ gradually decorrelates over time, representing the drift in the $N$-dimensional feature space. This pattern is consistent with the low-rank learned-induced correction predicted by Eq.\ref{eq:representation}.
\subsection{Stability of Computation}
How does the system retain its functionality in the presence of constant reordering of the tuning of individual neurons? 
In our
model, the stability of the input-output function of the network during the equilibrium diffusion
phase is due to the continuous realignment of the readout weights ${\bf a}({t})$
and the hidden layers weights ${\bf{W}}({t})$
as they drift simultaneously, staying within the space of solutions as was suggested previously \cite{rule2020stable,rule2022self,masset2022drifting}.
The above alignment scenario requires an ongoing learning signal acting
on the weights, in the form of a weak ($\mathcal{O}(T)$) gradient. In the absence of such a signal, 
the changes in the predictor due to the drift relative to an initial time $t_0$ (in which the system has already achieved small error)  is given by
\begin{align}
& \left\langle f\left({\bf x},t,t_{0}\right)\right\rangle = \\ &e^{-T\sigma^{-2}\left(t-t_{0}\right)}k^{L}\left(t,t_{0},{\bf x}\right)^{\top}\left(IT\sigma^{-2}+K_{GP}^{L}\right)^{-1}Y\nonumber
\end{align}
where the exponential prefactor is due to decorrelation of the readout weights from their learned values at time $t_0$, resulting in an overall decay of the predictor to zero and chance level performance. The effect of decorrelation in the hidden layer weights is represented in the time-dependence of the kernel (see below). 

We next consider an alternative scenario where the readout weights are frozen at their learned values at  $t_{0}$ while the weights of the hidden layers ${\bf W}({t})$ continue
to drift without an external learning signal. We denote
the output of the network in this scenario as $f_{\text{drift}}\left(t,t_{0},{\bf x}\right)$. Our theory predicts that  the mean of $f_{\text{drift}}\left({\bf x},t,t_{0}\right)$ (see
SI Sec.\ref{subsec:details-rep-drift} for details)
is given by
\begin{align}
& \left\langle f_{\text{drift}}\left(t,t_{0},{\bf x}\right)\right\rangle = \label{eq:drift predictor}\\ & k^{L}\left(t,t_{0},{\bf x}\right)^{\top}\left(IT\sigma^{-2}+K_{GP}^{L}\right)^{-1}Y\nonumber
\end{align}
The kernel $k^{L}({\bf x},t,t_{0})$ represents the overlap between
the top layer activations at time $t_{0}$, induced by  the training inputs and that
of a test point at time $t$. When $t-t_{0}$ is large, the two representations
completely decorrelate and the predictor is determined by the 'mean' kernel function.
\begin{equation}
\mathcal{K}_{mean}^{L}\left({\bf x},{\bf x}^{\prime}\right)=\frac{1}{N_{L}}\left\langle \phi\left({\bf z}^{L}({\bf x}\right)\right\rangle _{{0}}\cdot\left\langle \phi\left({\bf z}^{L}({\bf x'}\right)\right\rangle _{{0}}\label{eq:K mean}
\end{equation}
which is a modified version of the NNGP kernel where the Gaussian averages
are performed separately for each data point.
Thus, the long time limit of the mean predictor in this scenario is
\begin{align}
& \lim_{t-t_{0}\rightarrow\infty}\left\langle f_{\text{drift}}\left(t,t_{0},{\bf x}\right)\right\rangle = \\ & k_{mean}^{L}\left({\bf x}\right)^{\top}\left(IT\sigma^{-2}+K_{GP}^{L}\right)^{-1}Y\nonumber
\end{align}
where $k_{mean}^{L}\left({\bf x}\right)$ is defined as applying the
mean kernel function to the test data. In this scenario, the predictor does not necessarily decay to chance level, as discussed in the next section. 

\begin{figure*}[hbt!]
\centering{}\includegraphics[width=0.95\textwidth]{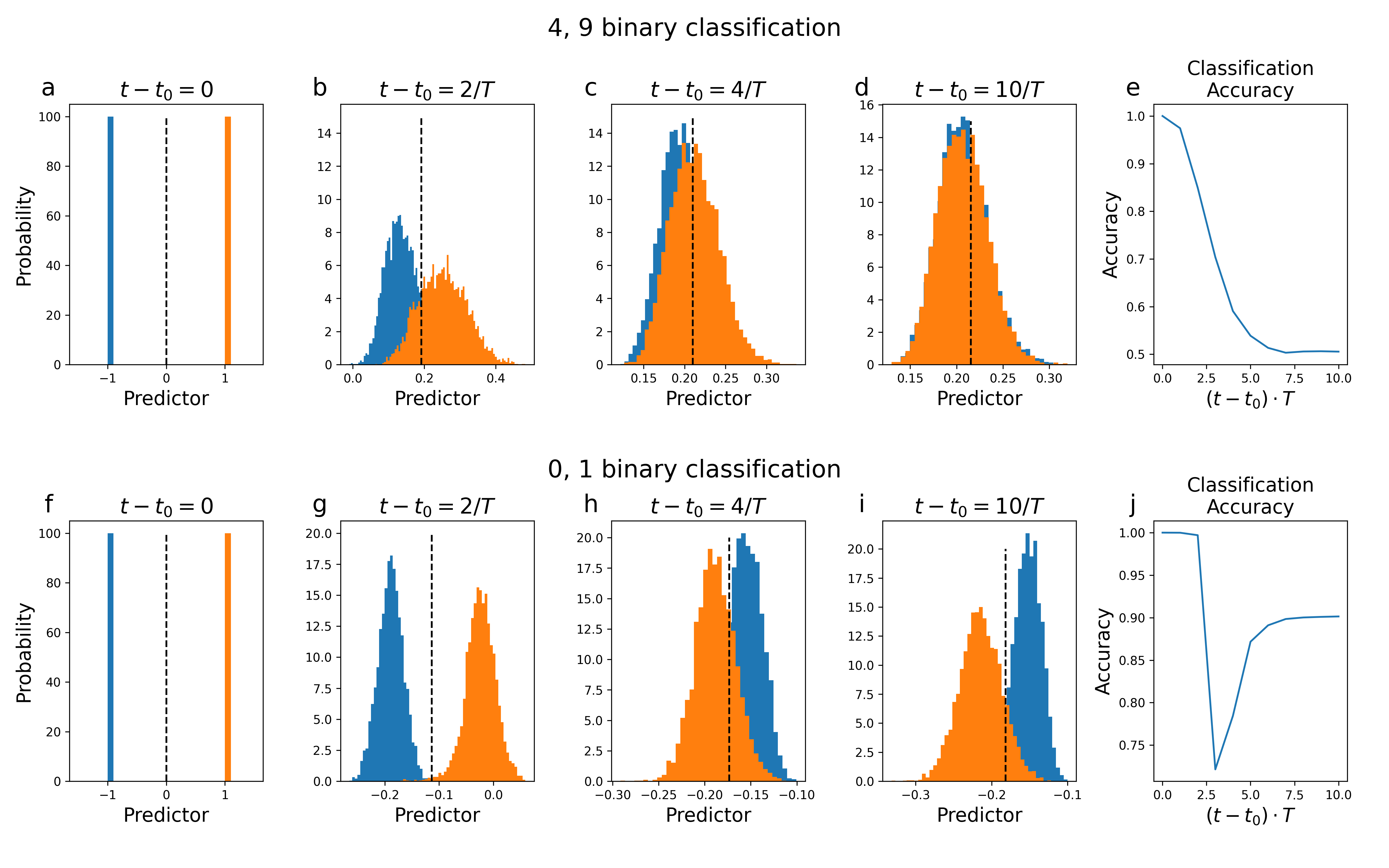}\caption{\label{fig:equilibrium drift}Representational Drift with Fixed Readout Weights After Learning in ReLU network:
(a-d, f-i) The histogram of the predictor during drift, $f_{\text{drift}}\left({\bf x}, t,t_{0}\right)$, on the training data. (a,f) Initially, there is perfect class separation at $\pm1$. Performance gradually deteriorates as the readout weights ${\bf a}(t_{0})$ and the hidden layer weights ${\bf W}({t})$ lose alignment due to drift. In the classification task involving digits 0 and 1, the histogram is still separable after complete decorrelation, due to differences in the class norms (see Sec.\ref{sec:invariant code}). In the classification task of the digits 4 and 9, performance declines to chance level.
(e, j) The classification accuracy using an optimal threshold is plotted as a function of the time difference from the freezing time $t_{0}$.}
\end{figure*}
\begin{figure*}[hbt!]
\centering{}\includegraphics[width=1.0\textwidth]{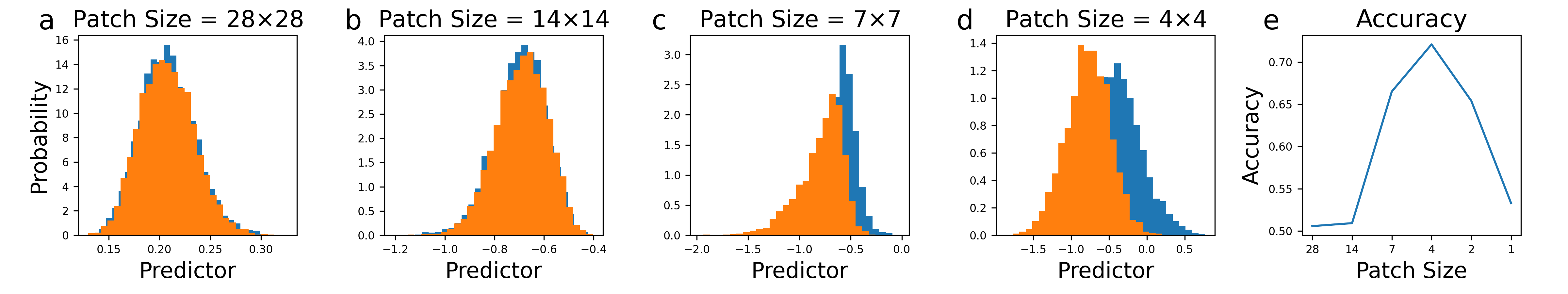}\caption{\label{fig:drift bias}Architectural Bias: We demonstrate how limiting the receptive field (via convolution) in ReLU networks can enhance performance after drift. Each neuron in the network receives input from a fixed patch from the total image, and we vary patch size (the total image is $28\times28$). (a-d) The histograms are more separable when the receptive field is limited, with an optimal patch size of $4\times4$. (f) The accuracy measured by an optimal threshold reaches 72\% compared to chance-level with an unlimited receptive field. The limited receptive field emphasizes class differences, as the norms of small patches exhibit greater variability between classes than the norms of the entire image. Importantly, if the patch size is too small, structural information about the image is lost, leading to poor performance.}
\end{figure*}
\subsection {Invariant Code \label{sec:invariant code}}
It has been hypothesized  that the observed stability of the network function against the drift process is due to the fact that the readout utilizes a representational feature which is invariant to permutation of the neurons \cite{deitch2021representational,rule2019causes,marks2021stimulus,rubin2019revealing,druckmann2012neuronal,kaufman2014cortical}. A simple example would be the case where the statistics of the population firing rates (e.g., mean firing rate) is modulated by the stimulus. Does our network exhibit a representational code which is invariant to the random sampling of the hidden weights over long times? The answer depends on both the type of non-linearity of the hidden layer neurons and the nature of the task. Above, we have shown that any information about the input which is invariant to the weight sampling should be contained in the mean-kernel, see Eq.\ref{eq:K mean}.  For odd nonlinearities (e.g., linear and error function activations), $\mathcal{K}_{mean}^{L}\left({\bf x},{\bf x}^{\prime}\right)$ is identically zero. However, this is not true for other nonlinearities (e.g., ReLU) where the mean activity remains non-zero. Since the distribution of the preactivations is symmetric around zero, it is invariant to the sign of the input activations, hence it cannot encode any task information embedded in the first moment of the inputs. However, the mean post activation does  depend on the norm of the input vectors, $\left\Vert {\bf x}\right\Vert$ and $\left\Vert {\bf x}^{\prime}\right\Vert$, as mentioned in Sec.\ref{subsec:The-neural-dynamical} and can be seen in Fig.\ref{fig: kernels} (f) above. As a result, if the distribution of the norms carries information relevant to the given task, the predictor can retain significant information despite the drift. On the other hand, if the norms of the inputs are not modulated in a task-dependent manner, the decorrelated representations yield chance-level performance.

We present examples illustrating both scenarios in Fig.\ref{fig:equilibrium drift}. Specifically, we consider two MNIST binary classification tasks after reaching the long-time equilibrium. For each task, we show the evolution of the histograms of the predictor on the training examples at time $t$, after freezing the readout weights at an earlier time $t_{0}$. To evaluate the amount of discriminability retained in the predictor histograms, we evaluated the classification accuracy given by the optimal separating threshold (see SI Sec.\ref{sec:Details-of-the} for details). 
In the case of  digits 4 and 9, the two histograms eventually overlap, resulting in long-time chance-level accuracy and a complete loss of the learned information. In contrast, for the digits 0 and 1 (Fig.~\ref{fig:equilibrium drift}(f-j)), the histograms of the two classes remain partially separated, leading to a long-time accuracy of 90\%, which reflects the residual information contained in the input norms. Interestingly, during the transition from the initial state to the long-time state, the distributions temporarily cross, causing a brief period of chance performance.

Additional architectural biases can be leveraged to enhance the stability of function. As an example, we consider feedforward networks with local receptive fields (as in convolutional networks). With such constraints, even if the overall amplitude of an example is not informative of the class, the local amplitude, i.e., the $L_2$ norm of the inputs within a receptive field of each representation neuron, may carry information about the task, which will be preserved in the 'mean-kernel'. In Fig.~\ref{fig:drift bias}, we demonstrate that reducing the receptive field allows the overlapping distributions of digits 4 and 9 to become separable. The highest accuracy is achieved with a receptive field of $4 \times 4$ pixels, increasing performance from chance level to 72\%. When the receptive field becomes too small, the spatial structure of the digits is lost, and the performance is poor again. We stress that the limited receptive field breaks the permutation symmetry between neurons, and thus the drift process does not act as a complete reordering like was observed with unlimited receptive field in Fig.\ref{fig:tuning curve}. In biological circuits with limited receptive fields, this symmetry breaking may play a role in retaining computation in the presence of drift.

\section{Discussion}
Our work provides a theoretical framework for the complete trajectory of gradient descent learning dynamics in wide deep neural networks in the presence of small noise, unifying the NTK theory and the NNGP theory as two limits of the same underlying process. The dynamics is captured by the time-dependent Neural Dynamical Kernel (NDK),  a dynamical generalization of the NTK. Although noise is externally introduced in our setup, stochasticity in practical machine learning often arises from the random sampling of examples in mini-batches (such as in SGD). We speculate that the insights from the current theory may be relevant to this types of noise as well \cite{wu2020noisy,noh2017regularizing,mignacco2022effective,dalalyan2017further}.

The theory provides new insights into the learning dynamics during the diffusive learning phase, where the learning process explores the solution space. We focus on characterizing commonly used activation functions and elucidating their interactions with other hyperparameters. In particular, we highlight the impact of $\sigma_{0}^{2}$ and $\sigma^{2}$, which correspond to the variance of the weights at initialization and that of the Bayesian prior regularization, respectively. These parameters play a pivotal role in shaping the trajectories of the predictor and the time scales of the different stages of the learning dynamics. We demonstrate that for sigmoidal activation functions, the values of these variances may substantially affect the learning trajectory and performance. Small weight variance leads to linear-like behavior and poor generalization, whereas large values induce strong nonlinearity and typically result in improved performance. 
As a result, large differences in the two variances may cause drastic changes in behavior across time as the system moves from being dominated by the initial weights to the diffusive stage dominated by the prior regularization. These insights can be utilized in practical machine-learning applications.

Our Bayesian framework provides a model of representational drift
where the weights undergo random drifts which are compensated by continuous realignment of the hidden and readout weights, keeping the system in the solution space, as was
previously suggested \cite{rule2020stable,rule2022self}. In our Langevin learning dynamics,
this realignment is due to the presence of an ongoing learning signal, which is nonzero even at the equilibrium stage.
The source of the putative realignment signals in brain circuits is
unclear. An alternative hypothesis is that the computation in neuronal
circuits is based on representational features that are invariant to the 
drift process \cite{deitch2021representational,rule2019causes,marks2021stimulus,rubin2019revealing,druckmann2012neuronal,kaufman2014cortical}.
We show that in our framework, this scenario requires (1) that the readout weights are frozen after learning and (2) that task-relevant information impacts the structure of the representation kernel even after its decorrelation (the 'mean-kernel', Eq. \ref{eq:K mean}). We provide examples of such features and illustrate how invariant codes can be enhanced by appropriate architectural biases that further constrain the drift. 

So far, we have focused on learning in infinitely wide networks in
the lazy regime, where the time dependence of the NDK arises from
random drift in the solution space. Learning dynamics in finite width networks or in infinite width networks with non-lazy architecture are likely more complex \cite{shan2021theory,vyas2022limitations,canatar2022kernel, bordelon2022self,flesch2022orthogonal}. Previous works have also extended the equilibrium theory of Bayesian learning to lazy and non-lazy networks in which data size is proportional to the network
width \cite{li2022globally,li2021statistical,van2024coding,woodworth2020kernel,azulay2021implicit, pacelli2023statistical, cui2023bayes, fischer2024critical, rubin2024unified}. It will be important to generalize the current dynamical theory to capture these architectures and regimes.

\textbf{Acknowledgments:} We thank the anonymous reviewers for their helpful comments. This research is supported by the Gatsby Charitable Foundation, the Swartz Foundation,  ONR grant No.N0014-23-1-2051, and the Kempner Institute for the Study of Natural and Artificial Intelligence.

\newpage
\bibliography{main}

\begin{thebibliography}{73}%
\makeatletter
\providecommand \@ifxundefined [1]{%
 \@ifx{#1\undefined}
}%
\providecommand \@ifnum [1]{%
 \ifnum #1\expandafter \@firstoftwo
 \else \expandafter \@secondoftwo
 \fi
}%
\providecommand \@ifx [1]{%
 \ifx #1\expandafter \@firstoftwo
 \else \expandafter \@secondoftwo
 \fi
}%
\providecommand \natexlab [1]{#1}%
\providecommand \enquote  [1]{``#1''}%
\providecommand \bibnamefont  [1]{#1}%
\providecommand \bibfnamefont [1]{#1}%
\providecommand \citenamefont [1]{#1}%
\providecommand \href@noop [0]{\@secondoftwo}%
\providecommand \href [0]{\begingroup \@sanitize@url \@href}%
\providecommand \@href[1]{\@@startlink{#1}\@@href}%
\providecommand \@@href[1]{\endgroup#1\@@endlink}%
\providecommand \@sanitize@url [0]{\catcode `\\12\catcode `\$12\catcode `\&12\catcode `\#12\catcode `\^12\catcode `\_12\catcode `\%12\relax}%
\providecommand \@@startlink[1]{}%
\providecommand \@@endlink[0]{}%
\providecommand \url  [0]{\begingroup\@sanitize@url \@url }%
\providecommand \@url [1]{\endgroup\@href {#1}{\urlprefix }}%
\providecommand \urlprefix  [0]{URL }%
\providecommand \Eprint [0]{\href }%
\providecommand \doibase [0]{https://doi.org/}%
\providecommand \selectlanguage [0]{\@gobble}%
\providecommand \bibinfo  [0]{\@secondoftwo}%
\providecommand \bibfield  [0]{\@secondoftwo}%
\providecommand \translation [1]{[#1]}%
\providecommand \BibitemOpen [0]{}%
\providecommand \bibitemStop [0]{}%
\providecommand \bibitemNoStop [0]{.\EOS\space}%
\providecommand \EOS [0]{\spacefactor3000\relax}%
\providecommand \BibitemShut  [1]{\csname bibitem#1\endcsname}%
\let\auto@bib@innerbib\@empty
\bibitem [{\citenamefont {Hazan}\ and\ \citenamefont {Jaakkola}(2015)}]{hazan2015steps}%
  \BibitemOpen
  \bibfield  {author} {\bibinfo {author} {\bibfnamefont {T.}~\bibnamefont {Hazan}}\ and\ \bibinfo {author} {\bibfnamefont {T.}~\bibnamefont {Jaakkola}},\ }\bibfield  {title} {\bibinfo {title} {Steps toward deep kernel methods from infinite neural networks},\ }\href@noop {} {\bibfield  {journal} {\bibinfo  {journal} {arXiv preprint arXiv:1508.05133}\ } (\bibinfo {year} {2015})}\BibitemShut {NoStop}%
\bibitem [{\citenamefont {Jacot}\ \emph {et~al.}(2018)\citenamefont {Jacot}, \citenamefont {Gabriel},\ and\ \citenamefont {Hongler}}]{jacot2018neural}%
  \BibitemOpen
  \bibfield  {author} {\bibinfo {author} {\bibfnamefont {A.}~\bibnamefont {Jacot}}, \bibinfo {author} {\bibfnamefont {F.}~\bibnamefont {Gabriel}},\ and\ \bibinfo {author} {\bibfnamefont {C.}~\bibnamefont {Hongler}},\ }\bibfield  {title} {\bibinfo {title} {Neural tangent kernel: Convergence and generalization in neural networks},\ }\href@noop {} {\bibfield  {journal} {\bibinfo  {journal} {Advances in neural information processing systems}\ }\textbf {\bibinfo {volume} {31}} (\bibinfo {year} {2018})}\BibitemShut {NoStop}%
\bibitem [{\citenamefont {Lee}\ \emph {et~al.}(2018)\citenamefont {Lee}, \citenamefont {Sohl-dickstein}, \citenamefont {Pennington}, \citenamefont {Novak}, \citenamefont {Schoenholz},\ and\ \citenamefont {Bahri}}]{lee2017deep}%
  \BibitemOpen
  \bibfield  {author} {\bibinfo {author} {\bibfnamefont {J.}~\bibnamefont {Lee}}, \bibinfo {author} {\bibfnamefont {J.}~\bibnamefont {Sohl-dickstein}}, \bibinfo {author} {\bibfnamefont {J.}~\bibnamefont {Pennington}}, \bibinfo {author} {\bibfnamefont {R.}~\bibnamefont {Novak}}, \bibinfo {author} {\bibfnamefont {S.}~\bibnamefont {Schoenholz}},\ and\ \bibinfo {author} {\bibfnamefont {Y.}~\bibnamefont {Bahri}},\ }\bibfield  {title} {\bibinfo {title} {Deep neural networks as gaussian processes},\ }in\ \href@noop {} {\emph {\bibinfo {booktitle} {International Conference on Learning Representations}}}\ (\bibinfo {year} {2018})\BibitemShut {NoStop}%
\bibitem [{\citenamefont {Lee}\ \emph {et~al.}(2019)\citenamefont {Lee}, \citenamefont {Xiao}, \citenamefont {Schoenholz}, \citenamefont {Bahri}, \citenamefont {Novak}, \citenamefont {Sohl-Dickstein},\ and\ \citenamefont {Pennington}}]{lee2019wide}%
  \BibitemOpen
  \bibfield  {author} {\bibinfo {author} {\bibfnamefont {J.}~\bibnamefont {Lee}}, \bibinfo {author} {\bibfnamefont {L.}~\bibnamefont {Xiao}}, \bibinfo {author} {\bibfnamefont {S.}~\bibnamefont {Schoenholz}}, \bibinfo {author} {\bibfnamefont {Y.}~\bibnamefont {Bahri}}, \bibinfo {author} {\bibfnamefont {R.}~\bibnamefont {Novak}}, \bibinfo {author} {\bibfnamefont {J.}~\bibnamefont {Sohl-Dickstein}},\ and\ \bibinfo {author} {\bibfnamefont {J.}~\bibnamefont {Pennington}},\ }\bibfield  {title} {\bibinfo {title} {Wide neural networks of any depth evolve as linear models under gradient descent},\ }\href@noop {} {\bibfield  {journal} {\bibinfo  {journal} {Advances in neural information processing systems}\ }\textbf {\bibinfo {volume} {32}} (\bibinfo {year} {2019})}\BibitemShut {NoStop}%
\bibitem [{\citenamefont {Matthews}\ \emph {et~al.}(2018)\citenamefont {Matthews}, \citenamefont {Rowland}, \citenamefont {Hron}, \citenamefont {Turner},\ and\ \citenamefont {Ghahramani}}]{matthews2018gaussian}%
  \BibitemOpen
  \bibfield  {author} {\bibinfo {author} {\bibfnamefont {A.~G. d.~G.}\ \bibnamefont {Matthews}}, \bibinfo {author} {\bibfnamefont {M.}~\bibnamefont {Rowland}}, \bibinfo {author} {\bibfnamefont {J.}~\bibnamefont {Hron}}, \bibinfo {author} {\bibfnamefont {R.~E.}\ \bibnamefont {Turner}},\ and\ \bibinfo {author} {\bibfnamefont {Z.}~\bibnamefont {Ghahramani}},\ }\bibfield  {title} {\bibinfo {title} {Gaussian process behaviour in wide deep neural networks},\ }\href@noop {} {\bibfield  {journal} {\bibinfo  {journal} {arXiv preprint arXiv:1804.11271}\ } (\bibinfo {year} {2018})}\BibitemShut {NoStop}%
\bibitem [{\citenamefont {Neal}(1994)}]{neal1994priors}%
  \BibitemOpen
  \bibfield  {author} {\bibinfo {author} {\bibfnamefont {R.~M.}\ \bibnamefont {Neal}},\ }\bibfield  {title} {\bibinfo {title} {Priors for infinite networks (tech. rep. no. crg-tr-94-1)},\ }\href@noop {} {\bibfield  {journal} {\bibinfo  {journal} {University of Toronto}\ }\textbf {\bibinfo {volume} {415}} (\bibinfo {year} {1994})}\BibitemShut {NoStop}%
\bibitem [{\citenamefont {Novak}\ \emph {et~al.}(2018)\citenamefont {Novak}, \citenamefont {Xiao}, \citenamefont {Lee}, \citenamefont {Bahri}, \citenamefont {Yang}, \citenamefont {Hron}, \citenamefont {Abolafia}, \citenamefont {Pennington},\ and\ \citenamefont {Sohl-Dickstein}}]{novak2018bayesian}%
  \BibitemOpen
  \bibfield  {author} {\bibinfo {author} {\bibfnamefont {R.}~\bibnamefont {Novak}}, \bibinfo {author} {\bibfnamefont {L.}~\bibnamefont {Xiao}}, \bibinfo {author} {\bibfnamefont {J.}~\bibnamefont {Lee}}, \bibinfo {author} {\bibfnamefont {Y.}~\bibnamefont {Bahri}}, \bibinfo {author} {\bibfnamefont {G.}~\bibnamefont {Yang}}, \bibinfo {author} {\bibfnamefont {J.}~\bibnamefont {Hron}}, \bibinfo {author} {\bibfnamefont {D.~A.}\ \bibnamefont {Abolafia}}, \bibinfo {author} {\bibfnamefont {J.}~\bibnamefont {Pennington}},\ and\ \bibinfo {author} {\bibfnamefont {J.}~\bibnamefont {Sohl-Dickstein}},\ }\bibfield  {title} {\bibinfo {title} {Bayesian deep convolutional networks with many channels are gaussian processes},\ }\href@noop {} {\bibfield  {journal} {\bibinfo  {journal} {arXiv preprint arXiv:1810.05148}\ } (\bibinfo {year} {2018})}\BibitemShut {NoStop}%
\bibitem [{\citenamefont {Novak}\ \emph {et~al.}(2019)\citenamefont {Novak}, \citenamefont {Xiao}, \citenamefont {Hron}, \citenamefont {Lee}, \citenamefont {Alemi}, \citenamefont {Sohl-Dickstein},\ and\ \citenamefont {Schoenholz}}]{novak2019neural}%
  \BibitemOpen
  \bibfield  {author} {\bibinfo {author} {\bibfnamefont {R.}~\bibnamefont {Novak}}, \bibinfo {author} {\bibfnamefont {L.}~\bibnamefont {Xiao}}, \bibinfo {author} {\bibfnamefont {J.}~\bibnamefont {Hron}}, \bibinfo {author} {\bibfnamefont {J.}~\bibnamefont {Lee}}, \bibinfo {author} {\bibfnamefont {A.~A.}\ \bibnamefont {Alemi}}, \bibinfo {author} {\bibfnamefont {J.}~\bibnamefont {Sohl-Dickstein}},\ and\ \bibinfo {author} {\bibfnamefont {S.~S.}\ \bibnamefont {Schoenholz}},\ }\bibfield  {title} {\bibinfo {title} {Neural tangents: Fast and easy infinite neural networks in python},\ }\href@noop {} {\bibfield  {journal} {\bibinfo  {journal} {arXiv preprint arXiv:1912.02803}\ } (\bibinfo {year} {2019})}\BibitemShut {NoStop}%
\bibitem [{\citenamefont {Sohl-Dickstein}\ \emph {et~al.}(2020)\citenamefont {Sohl-Dickstein}, \citenamefont {Novak}, \citenamefont {Schoenholz},\ and\ \citenamefont {Lee}}]{sohl2020infinite}%
  \BibitemOpen
  \bibfield  {author} {\bibinfo {author} {\bibfnamefont {J.}~\bibnamefont {Sohl-Dickstein}}, \bibinfo {author} {\bibfnamefont {R.}~\bibnamefont {Novak}}, \bibinfo {author} {\bibfnamefont {S.~S.}\ \bibnamefont {Schoenholz}},\ and\ \bibinfo {author} {\bibfnamefont {J.}~\bibnamefont {Lee}},\ }\bibfield  {title} {\bibinfo {title} {On the infinite width limit of neural networks with a standard parameterization},\ }\href@noop {} {\bibfield  {journal} {\bibinfo  {journal} {arXiv preprint arXiv:2001.07301}\ } (\bibinfo {year} {2020})}\BibitemShut {NoStop}%
\bibitem [{\citenamefont {Williams}(1996)}]{williams1996computing}%
  \BibitemOpen
  \bibfield  {author} {\bibinfo {author} {\bibfnamefont {C.}~\bibnamefont {Williams}},\ }\bibfield  {title} {\bibinfo {title} {Computing with infinite networks},\ }\href@noop {} {\bibfield  {journal} {\bibinfo  {journal} {Advances in neural information processing systems}\ }\textbf {\bibinfo {volume} {9}} (\bibinfo {year} {1996})}\BibitemShut {NoStop}%
\bibitem [{\citenamefont {Yang}(2019)}]{yang2019wide}%
  \BibitemOpen
  \bibfield  {author} {\bibinfo {author} {\bibfnamefont {G.}~\bibnamefont {Yang}},\ }\bibfield  {title} {\bibinfo {title} {Wide feedforward or recurrent neural networks of any architecture are gaussian processes},\ }\href@noop {} {\bibfield  {journal} {\bibinfo  {journal} {Advances in Neural Information Processing Systems}\ }\textbf {\bibinfo {volume} {32}} (\bibinfo {year} {2019})}\BibitemShut {NoStop}%
\bibitem [{\citenamefont {Chizat}\ and\ \citenamefont {Bach}(2020)}]{chizat2020implicit}%
  \BibitemOpen
  \bibfield  {author} {\bibinfo {author} {\bibfnamefont {L.}~\bibnamefont {Chizat}}\ and\ \bibinfo {author} {\bibfnamefont {F.}~\bibnamefont {Bach}},\ }\bibfield  {title} {\bibinfo {title} {Implicit bias of gradient descent for wide two-layer neural networks trained with the logistic loss},\ }in\ \href@noop {} {\emph {\bibinfo {booktitle} {Conference on Learning Theory}}}\ (\bibinfo {organization} {PMLR},\ \bibinfo {year} {2020})\ pp.\ \bibinfo {pages} {1305--1338}\BibitemShut {NoStop}%
\bibitem [{\citenamefont {Jin}\ and\ \citenamefont {Mont{\'u}far}(2020)}]{jin2020implicit}%
  \BibitemOpen
  \bibfield  {author} {\bibinfo {author} {\bibfnamefont {H.}~\bibnamefont {Jin}}\ and\ \bibinfo {author} {\bibfnamefont {G.}~\bibnamefont {Mont{\'u}far}},\ }\bibfield  {title} {\bibinfo {title} {Implicit bias of gradient descent for mean squared error regression with wide neural networks},\ }\href@noop {} {\bibfield  {journal} {\bibinfo  {journal} {arXiv preprint arXiv:2006.07356}\ } (\bibinfo {year} {2020})}\BibitemShut {NoStop}%
\bibitem [{\citenamefont {Min}\ \emph {et~al.}(2021)\citenamefont {Min}, \citenamefont {Tarmoun}, \citenamefont {Vidal},\ and\ \citenamefont {Mallada}}]{min2021explicit}%
  \BibitemOpen
  \bibfield  {author} {\bibinfo {author} {\bibfnamefont {H.}~\bibnamefont {Min}}, \bibinfo {author} {\bibfnamefont {S.}~\bibnamefont {Tarmoun}}, \bibinfo {author} {\bibfnamefont {R.}~\bibnamefont {Vidal}},\ and\ \bibinfo {author} {\bibfnamefont {E.}~\bibnamefont {Mallada}},\ }\bibfield  {title} {\bibinfo {title} {On the explicit role of initialization on the convergence and implicit bias of overparametrized linear networks},\ }in\ \href@noop {} {\emph {\bibinfo {booktitle} {International Conference on Machine Learning}}}\ (\bibinfo {organization} {PMLR},\ \bibinfo {year} {2021})\ pp.\ \bibinfo {pages} {7760--7768}\BibitemShut {NoStop}%
\bibitem [{\citenamefont {Cho}\ and\ \citenamefont {Saul}(2009)}]{cho2009kernel}%
  \BibitemOpen
  \bibfield  {author} {\bibinfo {author} {\bibfnamefont {Y.}~\bibnamefont {Cho}}\ and\ \bibinfo {author} {\bibfnamefont {L.}~\bibnamefont {Saul}},\ }\bibfield  {title} {\bibinfo {title} {Kernel methods for deep learning},\ }\href@noop {} {\bibfield  {journal} {\bibinfo  {journal} {Advances in neural information processing systems}\ }\textbf {\bibinfo {volume} {22}} (\bibinfo {year} {2009})}\BibitemShut {NoStop}%
\bibitem [{\citenamefont {Lee}\ \emph {et~al.}(2020)\citenamefont {Lee}, \citenamefont {Schoenholz}, \citenamefont {Pennington}, \citenamefont {Adlam}, \citenamefont {Xiao}, \citenamefont {Novak},\ and\ \citenamefont {Sohl-Dickstein}}]{lee2020finite}%
  \BibitemOpen
  \bibfield  {author} {\bibinfo {author} {\bibfnamefont {J.}~\bibnamefont {Lee}}, \bibinfo {author} {\bibfnamefont {S.}~\bibnamefont {Schoenholz}}, \bibinfo {author} {\bibfnamefont {J.}~\bibnamefont {Pennington}}, \bibinfo {author} {\bibfnamefont {B.}~\bibnamefont {Adlam}}, \bibinfo {author} {\bibfnamefont {L.}~\bibnamefont {Xiao}}, \bibinfo {author} {\bibfnamefont {R.}~\bibnamefont {Novak}},\ and\ \bibinfo {author} {\bibfnamefont {J.}~\bibnamefont {Sohl-Dickstein}},\ }\bibfield  {title} {\bibinfo {title} {Finite versus infinite neural networks: an empirical study},\ }\href@noop {} {\bibfield  {journal} {\bibinfo  {journal} {Advances in Neural Information Processing Systems}\ }\textbf {\bibinfo {volume} {33}},\ \bibinfo {pages} {15156} (\bibinfo {year} {2020})}\BibitemShut {NoStop}%
\bibitem [{\citenamefont {Deitch}\ \emph {et~al.}(2021)\citenamefont {Deitch}, \citenamefont {Rubin},\ and\ \citenamefont {Ziv}}]{deitch2021representational}%
  \BibitemOpen
  \bibfield  {author} {\bibinfo {author} {\bibfnamefont {D.}~\bibnamefont {Deitch}}, \bibinfo {author} {\bibfnamefont {A.}~\bibnamefont {Rubin}},\ and\ \bibinfo {author} {\bibfnamefont {Y.}~\bibnamefont {Ziv}},\ }\bibfield  {title} {\bibinfo {title} {Representational drift in the mouse visual cortex},\ }\href@noop {} {\bibfield  {journal} {\bibinfo  {journal} {Current biology}\ }\textbf {\bibinfo {volume} {31}},\ \bibinfo {pages} {4327} (\bibinfo {year} {2021})}\BibitemShut {NoStop}%
\bibitem [{\citenamefont {Marks}\ and\ \citenamefont {Goard}(2021)}]{marks2021stimulus}%
  \BibitemOpen
  \bibfield  {author} {\bibinfo {author} {\bibfnamefont {T.~D.}\ \bibnamefont {Marks}}\ and\ \bibinfo {author} {\bibfnamefont {M.~J.}\ \bibnamefont {Goard}},\ }\bibfield  {title} {\bibinfo {title} {Stimulus-dependent representational drift in primary visual cortex},\ }\href@noop {} {\bibfield  {journal} {\bibinfo  {journal} {Nature communications}\ }\textbf {\bibinfo {volume} {12}},\ \bibinfo {pages} {5169} (\bibinfo {year} {2021})}\BibitemShut {NoStop}%
\bibitem [{\citenamefont {Rokni}\ \emph {et~al.}(2007)\citenamefont {Rokni}, \citenamefont {Richardson}, \citenamefont {Bizzi},\ and\ \citenamefont {Seung}}]{rokni2007motor}%
  \BibitemOpen
  \bibfield  {author} {\bibinfo {author} {\bibfnamefont {U.}~\bibnamefont {Rokni}}, \bibinfo {author} {\bibfnamefont {A.~G.}\ \bibnamefont {Richardson}}, \bibinfo {author} {\bibfnamefont {E.}~\bibnamefont {Bizzi}},\ and\ \bibinfo {author} {\bibfnamefont {H.~S.}\ \bibnamefont {Seung}},\ }\bibfield  {title} {\bibinfo {title} {Motor learning with unstable neural representations},\ }\href@noop {} {\bibfield  {journal} {\bibinfo  {journal} {Neuron}\ }\textbf {\bibinfo {volume} {54}},\ \bibinfo {pages} {653} (\bibinfo {year} {2007})}\BibitemShut {NoStop}%
\bibitem [{\citenamefont {Schoonover}\ \emph {et~al.}(2021)\citenamefont {Schoonover}, \citenamefont {Ohashi}, \citenamefont {Axel},\ and\ \citenamefont {Fink}}]{schoonover2021representational}%
  \BibitemOpen
  \bibfield  {author} {\bibinfo {author} {\bibfnamefont {C.~E.}\ \bibnamefont {Schoonover}}, \bibinfo {author} {\bibfnamefont {S.~N.}\ \bibnamefont {Ohashi}}, \bibinfo {author} {\bibfnamefont {R.}~\bibnamefont {Axel}},\ and\ \bibinfo {author} {\bibfnamefont {A.~J.}\ \bibnamefont {Fink}},\ }\bibfield  {title} {\bibinfo {title} {Representational drift in primary olfactory cortex},\ }\href@noop {} {\bibfield  {journal} {\bibinfo  {journal} {Nature}\ }\textbf {\bibinfo {volume} {594}},\ \bibinfo {pages} {541} (\bibinfo {year} {2021})}\BibitemShut {NoStop}%
\bibitem [{\citenamefont {Rule}\ \emph {et~al.}(2019)\citenamefont {Rule}, \citenamefont {O'Leary},\ and\ \citenamefont {Harvey}}]{rule2019causes}%
  \BibitemOpen
  \bibfield  {author} {\bibinfo {author} {\bibfnamefont {M.~E.}\ \bibnamefont {Rule}}, \bibinfo {author} {\bibfnamefont {T.}~\bibnamefont {O'Leary}},\ and\ \bibinfo {author} {\bibfnamefont {C.~D.}\ \bibnamefont {Harvey}},\ }\bibfield  {title} {\bibinfo {title} {Causes and consequences of representational drift},\ }\href@noop {} {\bibfield  {journal} {\bibinfo  {journal} {Current opinion in neurobiology}\ }\textbf {\bibinfo {volume} {58}},\ \bibinfo {pages} {141} (\bibinfo {year} {2019})}\BibitemShut {NoStop}%
\bibitem [{\citenamefont {Qin}\ \emph {et~al.}(2023)\citenamefont {Qin}, \citenamefont {Farashahi}, \citenamefont {Lipshutz}, \citenamefont {Sengupta}, \citenamefont {Chklovskii},\ and\ \citenamefont {Pehlevan}}]{qin2023coordinated}%
  \BibitemOpen
  \bibfield  {author} {\bibinfo {author} {\bibfnamefont {S.}~\bibnamefont {Qin}}, \bibinfo {author} {\bibfnamefont {S.}~\bibnamefont {Farashahi}}, \bibinfo {author} {\bibfnamefont {D.}~\bibnamefont {Lipshutz}}, \bibinfo {author} {\bibfnamefont {A.~M.}\ \bibnamefont {Sengupta}}, \bibinfo {author} {\bibfnamefont {D.~B.}\ \bibnamefont {Chklovskii}},\ and\ \bibinfo {author} {\bibfnamefont {C.}~\bibnamefont {Pehlevan}},\ }\bibfield  {title} {\bibinfo {title} {Coordinated drift of receptive fields in hebbian/anti-hebbian network models during noisy representation learning},\ }\href@noop {} {\bibfield  {journal} {\bibinfo  {journal} {Nature Neuroscience}\ ,\ \bibinfo {pages} {1}} (\bibinfo {year} {2023})}\BibitemShut {NoStop}%
\bibitem [{\citenamefont {Masset}\ \emph {et~al.}(2022)\citenamefont {Masset}, \citenamefont {Qin},\ and\ \citenamefont {Zavatone-Veth}}]{masset2022drifting}%
  \BibitemOpen
  \bibfield  {author} {\bibinfo {author} {\bibfnamefont {P.}~\bibnamefont {Masset}}, \bibinfo {author} {\bibfnamefont {S.}~\bibnamefont {Qin}},\ and\ \bibinfo {author} {\bibfnamefont {J.~A.}\ \bibnamefont {Zavatone-Veth}},\ }\bibfield  {title} {\bibinfo {title} {Drifting neuronal representations: Bug or feature?},\ }\href@noop {} {\bibfield  {journal} {\bibinfo  {journal} {Biological Cybernetics}\ }\textbf {\bibinfo {volume} {116}},\ \bibinfo {pages} {253} (\bibinfo {year} {2022})}\BibitemShut {NoStop}%
\bibitem [{\citenamefont {Coffey}\ and\ \citenamefont {Kalmykov}(2012)}]{coffey2012langevin}%
  \BibitemOpen
  \bibfield  {author} {\bibinfo {author} {\bibfnamefont {W.}~\bibnamefont {Coffey}}\ and\ \bibinfo {author} {\bibfnamefont {Y.~P.}\ \bibnamefont {Kalmykov}},\ }\href@noop {} {\emph {\bibinfo {title} {The Langevin equation: with applications to stochastic problems in physics, chemistry and electrical engineering}}},\ Vol.~\bibinfo {volume} {27}\ (\bibinfo  {publisher} {World Scientific},\ \bibinfo {year} {2012})\BibitemShut {NoStop}%
\bibitem [{\citenamefont {Welling}\ and\ \citenamefont {Teh}(2011)}]{welling2011bayesian}%
  \BibitemOpen
  \bibfield  {author} {\bibinfo {author} {\bibfnamefont {M.}~\bibnamefont {Welling}}\ and\ \bibinfo {author} {\bibfnamefont {Y.~W.}\ \bibnamefont {Teh}},\ }\bibfield  {title} {\bibinfo {title} {Bayesian learning via stochastic gradient langevin dynamics},\ }in\ \href@noop {} {\emph {\bibinfo {booktitle} {Proceedings of the 28th international conference on machine learning (ICML-11)}}}\ (\bibinfo {year} {2011})\ pp.\ \bibinfo {pages} {681--688}\BibitemShut {NoStop}%
\bibitem [{\citenamefont {Shwartz-Ziv}\ and\ \citenamefont {Tishby}(2017)}]{shwartz2017opening}%
  \BibitemOpen
  \bibfield  {author} {\bibinfo {author} {\bibfnamefont {R.}~\bibnamefont {Shwartz-Ziv}}\ and\ \bibinfo {author} {\bibfnamefont {N.}~\bibnamefont {Tishby}},\ }\bibfield  {title} {\bibinfo {title} {Opening the black box of deep neural networks via information},\ }\href@noop {} {\bibfield  {journal} {\bibinfo  {journal} {arXiv preprint arXiv:1703.00810}\ } (\bibinfo {year} {2017})}\BibitemShut {NoStop}%
\bibitem [{\citenamefont {Ratzon}\ \emph {et~al.}(2024)\citenamefont {Ratzon}, \citenamefont {Derdikman},\ and\ \citenamefont {Barak}}]{ratzon2024representational}%
  \BibitemOpen
  \bibfield  {author} {\bibinfo {author} {\bibfnamefont {A.}~\bibnamefont {Ratzon}}, \bibinfo {author} {\bibfnamefont {D.}~\bibnamefont {Derdikman}},\ and\ \bibinfo {author} {\bibfnamefont {O.}~\bibnamefont {Barak}},\ }\bibfield  {title} {\bibinfo {title} {Representational drift as a result of implicit regularization},\ }\href@noop {} {\bibfield  {journal} {\bibinfo  {journal} {Elife}\ }\textbf {\bibinfo {volume} {12}},\ \bibinfo {pages} {RP90069} (\bibinfo {year} {2024})}\BibitemShut {NoStop}%
\bibitem [{\citenamefont {Krogh}\ and\ \citenamefont {Hertz}(1991)}]{krogh1991simple}%
  \BibitemOpen
  \bibfield  {author} {\bibinfo {author} {\bibfnamefont {A.}~\bibnamefont {Krogh}}\ and\ \bibinfo {author} {\bibfnamefont {J.}~\bibnamefont {Hertz}},\ }\bibfield  {title} {\bibinfo {title} {A simple weight decay can improve generalization},\ }\href@noop {} {\bibfield  {journal} {\bibinfo  {journal} {Advances in neural information processing systems}\ }\textbf {\bibinfo {volume} {4}} (\bibinfo {year} {1991})}\BibitemShut {NoStop}%
\bibitem [{\citenamefont {Galanti}\ \emph {et~al.}(2022)\citenamefont {Galanti}, \citenamefont {Siegel}, \citenamefont {Gupte},\ and\ \citenamefont {Poggio}}]{galanti2022characterizing}%
  \BibitemOpen
  \bibfield  {author} {\bibinfo {author} {\bibfnamefont {T.}~\bibnamefont {Galanti}}, \bibinfo {author} {\bibfnamefont {Z.~S.}\ \bibnamefont {Siegel}}, \bibinfo {author} {\bibfnamefont {A.}~\bibnamefont {Gupte}},\ and\ \bibinfo {author} {\bibfnamefont {T.}~\bibnamefont {Poggio}},\ }\bibfield  {title} {\bibinfo {title} {Characterizing the implicit bias of regularized sgd in rank minimization},\ }\href@noop {} {\bibfield  {journal} {\bibinfo  {journal} {CoRR, abs/2206.05794 v6}\ } (\bibinfo {year} {2022})}\BibitemShut {NoStop}%
\bibitem [{\citenamefont {Li}\ and\ \citenamefont {Sompolinsky}(2021)}]{li2021statistical}%
  \BibitemOpen
  \bibfield  {author} {\bibinfo {author} {\bibfnamefont {Q.}~\bibnamefont {Li}}\ and\ \bibinfo {author} {\bibfnamefont {H.}~\bibnamefont {Sompolinsky}},\ }\bibfield  {title} {\bibinfo {title} {Statistical mechanics of deep linear neural networks: The backpropagating kernel renormalization},\ }\href@noop {} {\bibfield  {journal} {\bibinfo  {journal} {Physical Review X}\ }\textbf {\bibinfo {volume} {11}},\ \bibinfo {pages} {031059} (\bibinfo {year} {2021})}\BibitemShut {NoStop}%
\bibitem [{\citenamefont {Uhlenbeck}\ and\ \citenamefont {Ornstein}(1930)}]{uhlenbeck1930theory}%
  \BibitemOpen
  \bibfield  {author} {\bibinfo {author} {\bibfnamefont {G.~E.}\ \bibnamefont {Uhlenbeck}}\ and\ \bibinfo {author} {\bibfnamefont {L.~S.}\ \bibnamefont {Ornstein}},\ }\bibfield  {title} {\bibinfo {title} {On the theory of the brownian motion},\ }\href@noop {} {\bibfield  {journal} {\bibinfo  {journal} {Physical review}\ }\textbf {\bibinfo {volume} {36}},\ \bibinfo {pages} {823} (\bibinfo {year} {1930})}\BibitemShut {NoStop}%
\bibitem [{\citenamefont {Krizhevsky}\ \emph {et~al.}(2014)\citenamefont {Krizhevsky}, \citenamefont {Nair},\ and\ \citenamefont {Hinton}}]{krizhevsky2014cifar}%
  \BibitemOpen
  \bibfield  {author} {\bibinfo {author} {\bibfnamefont {A.}~\bibnamefont {Krizhevsky}}, \bibinfo {author} {\bibfnamefont {V.}~\bibnamefont {Nair}},\ and\ \bibinfo {author} {\bibfnamefont {G.}~\bibnamefont {Hinton}},\ }\bibfield  {title} {\bibinfo {title} {The cifar-10 dataset},\ }\href@noop {} {\bibfield  {journal} {\bibinfo  {journal} {online: http://www. cs. toronto. edu/kriz/cifar. html}\ }\textbf {\bibinfo {volume} {55}} (\bibinfo {year} {2014})}\BibitemShut {NoStop}%
\bibitem [{\citenamefont {Kubo}(1966)}]{kubo1966fluctuation}%
  \BibitemOpen
  \bibfield  {author} {\bibinfo {author} {\bibfnamefont {R.}~\bibnamefont {Kubo}},\ }\bibfield  {title} {\bibinfo {title} {The fluctuation-dissipation theorem},\ }\href@noop {} {\bibfield  {journal} {\bibinfo  {journal} {Reports on progress in physics}\ }\textbf {\bibinfo {volume} {29}},\ \bibinfo {pages} {255} (\bibinfo {year} {1966})}\BibitemShut {NoStop}%
\bibitem [{\citenamefont {Rule}\ \emph {et~al.}(2020)\citenamefont {Rule}, \citenamefont {Loback}, \citenamefont {Raman}, \citenamefont {Driscoll}, \citenamefont {Harvey},\ and\ \citenamefont {O'Leary}}]{rule2020stable}%
  \BibitemOpen
  \bibfield  {author} {\bibinfo {author} {\bibfnamefont {M.~E.}\ \bibnamefont {Rule}}, \bibinfo {author} {\bibfnamefont {A.~R.}\ \bibnamefont {Loback}}, \bibinfo {author} {\bibfnamefont {D.~V.}\ \bibnamefont {Raman}}, \bibinfo {author} {\bibfnamefont {L.~N.}\ \bibnamefont {Driscoll}}, \bibinfo {author} {\bibfnamefont {C.~D.}\ \bibnamefont {Harvey}},\ and\ \bibinfo {author} {\bibfnamefont {T.}~\bibnamefont {O'Leary}},\ }\bibfield  {title} {\bibinfo {title} {Stable task information from an unstable neural population},\ }\href@noop {} {\bibfield  {journal} {\bibinfo  {journal} {Elife}\ }\textbf {\bibinfo {volume} {9}},\ \bibinfo {pages} {e51121} (\bibinfo {year} {2020})}\BibitemShut {NoStop}%
\bibitem [{\citenamefont {Rule}\ and\ \citenamefont {O'Leary}(2022)}]{rule2022self}%
  \BibitemOpen
  \bibfield  {author} {\bibinfo {author} {\bibfnamefont {M.~E.}\ \bibnamefont {Rule}}\ and\ \bibinfo {author} {\bibfnamefont {T.}~\bibnamefont {O'Leary}},\ }\bibfield  {title} {\bibinfo {title} {Self-healing codes: How stable neural populations can track continually reconfiguring neural representations},\ }\href@noop {} {\bibfield  {journal} {\bibinfo  {journal} {Proceedings of the National Academy of Sciences}\ }\textbf {\bibinfo {volume} {119}},\ \bibinfo {pages} {e2106692119} (\bibinfo {year} {2022})}\BibitemShut {NoStop}%
\bibitem [{\citenamefont {Li}\ and\ \citenamefont {Sompolinsky}(2022)}]{li2022globally}%
  \BibitemOpen
  \bibfield  {author} {\bibinfo {author} {\bibfnamefont {Q.}~\bibnamefont {Li}}\ and\ \bibinfo {author} {\bibfnamefont {H.}~\bibnamefont {Sompolinsky}},\ }\bibfield  {title} {\bibinfo {title} {Globally gated deep linear networks},\ }\href@noop {} {\bibfield  {journal} {\bibinfo  {journal} {arXiv preprint arXiv:2210.17449}\ } (\bibinfo {year} {2022})}\BibitemShut {NoStop}%
\bibitem [{\citenamefont {Li}\ \emph {et~al.}(2024)\citenamefont {Li}, \citenamefont {Sorscher},\ and\ \citenamefont {Sompolinsky}}]{li2024representations}%
  \BibitemOpen
  \bibfield  {author} {\bibinfo {author} {\bibfnamefont {Q.}~\bibnamefont {Li}}, \bibinfo {author} {\bibfnamefont {B.}~\bibnamefont {Sorscher}},\ and\ \bibinfo {author} {\bibfnamefont {H.}~\bibnamefont {Sompolinsky}},\ }\bibfield  {title} {\bibinfo {title} {Representations and generalization in artificial and brain neural networks},\ }\href@noop {} {\bibfield  {journal} {\bibinfo  {journal} {Proceedings of the National Academy of Sciences}\ }\textbf {\bibinfo {volume} {121}},\ \bibinfo {pages} {e2311805121} (\bibinfo {year} {2024})}\BibitemShut {NoStop}%
\bibitem [{\citenamefont {Rubin}\ \emph {et~al.}(2019)\citenamefont {Rubin}, \citenamefont {Sheintuch}, \citenamefont {Brande-Eilat}, \citenamefont {Pinchasof}, \citenamefont {Rechavi}, \citenamefont {Geva},\ and\ \citenamefont {Ziv}}]{rubin2019revealing}%
  \BibitemOpen
  \bibfield  {author} {\bibinfo {author} {\bibfnamefont {A.}~\bibnamefont {Rubin}}, \bibinfo {author} {\bibfnamefont {L.}~\bibnamefont {Sheintuch}}, \bibinfo {author} {\bibfnamefont {N.}~\bibnamefont {Brande-Eilat}}, \bibinfo {author} {\bibfnamefont {O.}~\bibnamefont {Pinchasof}}, \bibinfo {author} {\bibfnamefont {Y.}~\bibnamefont {Rechavi}}, \bibinfo {author} {\bibfnamefont {N.}~\bibnamefont {Geva}},\ and\ \bibinfo {author} {\bibfnamefont {Y.}~\bibnamefont {Ziv}},\ }\bibfield  {title} {\bibinfo {title} {Revealing neural correlates of behavior without behavioral measurements},\ }\href@noop {} {\bibfield  {journal} {\bibinfo  {journal} {Nature communications}\ }\textbf {\bibinfo {volume} {10}},\ \bibinfo {pages} {4745} (\bibinfo {year} {2019})}\BibitemShut {NoStop}%
\bibitem [{\citenamefont {Druckmann}\ and\ \citenamefont {Chklovskii}(2012)}]{druckmann2012neuronal}%
  \BibitemOpen
  \bibfield  {author} {\bibinfo {author} {\bibfnamefont {S.}~\bibnamefont {Druckmann}}\ and\ \bibinfo {author} {\bibfnamefont {D.~B.}\ \bibnamefont {Chklovskii}},\ }\bibfield  {title} {\bibinfo {title} {Neuronal circuits underlying persistent representations despite time varying activity},\ }\href@noop {} {\bibfield  {journal} {\bibinfo  {journal} {Current Biology}\ }\textbf {\bibinfo {volume} {22}},\ \bibinfo {pages} {2095} (\bibinfo {year} {2012})}\BibitemShut {NoStop}%
\bibitem [{\citenamefont {Kaufman}\ \emph {et~al.}(2014)\citenamefont {Kaufman}, \citenamefont {Churchland}, \citenamefont {Ryu},\ and\ \citenamefont {Shenoy}}]{kaufman2014cortical}%
  \BibitemOpen
  \bibfield  {author} {\bibinfo {author} {\bibfnamefont {M.~T.}\ \bibnamefont {Kaufman}}, \bibinfo {author} {\bibfnamefont {M.~M.}\ \bibnamefont {Churchland}}, \bibinfo {author} {\bibfnamefont {S.~I.}\ \bibnamefont {Ryu}},\ and\ \bibinfo {author} {\bibfnamefont {K.~V.}\ \bibnamefont {Shenoy}},\ }\bibfield  {title} {\bibinfo {title} {Cortical activity in the null space: permitting preparation without movement},\ }\href@noop {} {\bibfield  {journal} {\bibinfo  {journal} {Nature neuroscience}\ }\textbf {\bibinfo {volume} {17}},\ \bibinfo {pages} {440} (\bibinfo {year} {2014})}\BibitemShut {NoStop}%
\bibitem [{\citenamefont {Wu}\ \emph {et~al.}(2020)\citenamefont {Wu}, \citenamefont {Hu}, \citenamefont {Xiong}, \citenamefont {Huan}, \citenamefont {Braverman},\ and\ \citenamefont {Zhu}}]{wu2020noisy}%
  \BibitemOpen
  \bibfield  {author} {\bibinfo {author} {\bibfnamefont {J.}~\bibnamefont {Wu}}, \bibinfo {author} {\bibfnamefont {W.}~\bibnamefont {Hu}}, \bibinfo {author} {\bibfnamefont {H.}~\bibnamefont {Xiong}}, \bibinfo {author} {\bibfnamefont {J.}~\bibnamefont {Huan}}, \bibinfo {author} {\bibfnamefont {V.}~\bibnamefont {Braverman}},\ and\ \bibinfo {author} {\bibfnamefont {Z.}~\bibnamefont {Zhu}},\ }\bibfield  {title} {\bibinfo {title} {On the noisy gradient descent that generalizes as sgd},\ }in\ \href@noop {} {\emph {\bibinfo {booktitle} {International Conference on Machine Learning}}}\ (\bibinfo {organization} {PMLR},\ \bibinfo {year} {2020})\ pp.\ \bibinfo {pages} {10367--10376}\BibitemShut {NoStop}%
\bibitem [{\citenamefont {Noh}\ \emph {et~al.}(2017)\citenamefont {Noh}, \citenamefont {You}, \citenamefont {Mun},\ and\ \citenamefont {Han}}]{noh2017regularizing}%
  \BibitemOpen
  \bibfield  {author} {\bibinfo {author} {\bibfnamefont {H.}~\bibnamefont {Noh}}, \bibinfo {author} {\bibfnamefont {T.}~\bibnamefont {You}}, \bibinfo {author} {\bibfnamefont {J.}~\bibnamefont {Mun}},\ and\ \bibinfo {author} {\bibfnamefont {B.}~\bibnamefont {Han}},\ }\bibfield  {title} {\bibinfo {title} {Regularizing deep neural networks by noise: Its interpretation and optimization},\ }\href@noop {} {\bibfield  {journal} {\bibinfo  {journal} {Advances in Neural Information Processing Systems}\ }\textbf {\bibinfo {volume} {30}} (\bibinfo {year} {2017})}\BibitemShut {NoStop}%
\bibitem [{\citenamefont {Mignacco}\ and\ \citenamefont {Urbani}(2022)}]{mignacco2022effective}%
  \BibitemOpen
  \bibfield  {author} {\bibinfo {author} {\bibfnamefont {F.}~\bibnamefont {Mignacco}}\ and\ \bibinfo {author} {\bibfnamefont {P.}~\bibnamefont {Urbani}},\ }\bibfield  {title} {\bibinfo {title} {The effective noise of stochastic gradient descent},\ }\href@noop {} {\bibfield  {journal} {\bibinfo  {journal} {Journal of Statistical Mechanics: Theory and Experiment}\ }\textbf {\bibinfo {volume} {2022}},\ \bibinfo {pages} {083405} (\bibinfo {year} {2022})}\BibitemShut {NoStop}%
\bibitem [{\citenamefont {Dalalyan}(2017)}]{dalalyan2017further}%
  \BibitemOpen
  \bibfield  {author} {\bibinfo {author} {\bibfnamefont {A.}~\bibnamefont {Dalalyan}},\ }\bibfield  {title} {\bibinfo {title} {Further and stronger analogy between sampling and optimization: Langevin monte carlo and gradient descent},\ }in\ \href@noop {} {\emph {\bibinfo {booktitle} {Conference on Learning Theory}}}\ (\bibinfo {organization} {PMLR},\ \bibinfo {year} {2017})\ pp.\ \bibinfo {pages} {678--689}\BibitemShut {NoStop}%
\bibitem [{\citenamefont {Shan}\ and\ \citenamefont {Bordelon}(2021)}]{shan2021theory}%
  \BibitemOpen
  \bibfield  {author} {\bibinfo {author} {\bibfnamefont {H.}~\bibnamefont {Shan}}\ and\ \bibinfo {author} {\bibfnamefont {B.}~\bibnamefont {Bordelon}},\ }\bibfield  {title} {\bibinfo {title} {A theory of neural tangent kernel alignment and its influence on training},\ }\href@noop {} {\bibfield  {journal} {\bibinfo  {journal} {arXiv preprint arXiv:2105.14301}\ } (\bibinfo {year} {2021})}\BibitemShut {NoStop}%
\bibitem [{\citenamefont {Vyas}\ \emph {et~al.}(2022)\citenamefont {Vyas}, \citenamefont {Bansal},\ and\ \citenamefont {Nakkiran}}]{vyas2022limitations}%
  \BibitemOpen
  \bibfield  {author} {\bibinfo {author} {\bibfnamefont {N.}~\bibnamefont {Vyas}}, \bibinfo {author} {\bibfnamefont {Y.}~\bibnamefont {Bansal}},\ and\ \bibinfo {author} {\bibfnamefont {P.}~\bibnamefont {Nakkiran}},\ }\bibfield  {title} {\bibinfo {title} {Limitations of the ntk for understanding generalization in deep learning},\ }\href@noop {} {\bibfield  {journal} {\bibinfo  {journal} {arXiv preprint arXiv:2206.10012}\ } (\bibinfo {year} {2022})}\BibitemShut {NoStop}%
\bibitem [{\citenamefont {Canatar}\ and\ \citenamefont {Pehlevan}(2022)}]{canatar2022kernel}%
  \BibitemOpen
  \bibfield  {author} {\bibinfo {author} {\bibfnamefont {A.}~\bibnamefont {Canatar}}\ and\ \bibinfo {author} {\bibfnamefont {C.}~\bibnamefont {Pehlevan}},\ }\bibfield  {title} {\bibinfo {title} {A kernel analysis of feature learning in deep neural networks},\ }in\ \href@noop {} {\emph {\bibinfo {booktitle} {2022 58th Annual Allerton Conference on Communication, Control, and Computing (Allerton)}}}\ (\bibinfo {organization} {IEEE},\ \bibinfo {year} {2022})\ pp.\ \bibinfo {pages} {1--8}\BibitemShut {NoStop}%
\bibitem [{\citenamefont {Bordelon}\ and\ \citenamefont {Pehlevan}(2022)}]{bordelon2022self}%
  \BibitemOpen
  \bibfield  {author} {\bibinfo {author} {\bibfnamefont {B.}~\bibnamefont {Bordelon}}\ and\ \bibinfo {author} {\bibfnamefont {C.}~\bibnamefont {Pehlevan}},\ }\bibfield  {title} {\bibinfo {title} {Self-consistent dynamical field theory of kernel evolution in wide neural networks},\ }\href@noop {} {\bibfield  {journal} {\bibinfo  {journal} {arXiv preprint arXiv:2205.09653}\ } (\bibinfo {year} {2022})}\BibitemShut {NoStop}%
\bibitem [{\citenamefont {Flesch}\ \emph {et~al.}(2022)\citenamefont {Flesch}, \citenamefont {Juechems}, \citenamefont {Dumbalska}, \citenamefont {Saxe},\ and\ \citenamefont {Summerfield}}]{flesch2022orthogonal}%
  \BibitemOpen
  \bibfield  {author} {\bibinfo {author} {\bibfnamefont {T.}~\bibnamefont {Flesch}}, \bibinfo {author} {\bibfnamefont {K.}~\bibnamefont {Juechems}}, \bibinfo {author} {\bibfnamefont {T.}~\bibnamefont {Dumbalska}}, \bibinfo {author} {\bibfnamefont {A.}~\bibnamefont {Saxe}},\ and\ \bibinfo {author} {\bibfnamefont {C.}~\bibnamefont {Summerfield}},\ }\bibfield  {title} {\bibinfo {title} {Orthogonal representations for robust context-dependent task performance in brains and neural networks},\ }\href@noop {} {\bibfield  {journal} {\bibinfo  {journal} {Neuron}\ }\textbf {\bibinfo {volume} {110}},\ \bibinfo {pages} {1258} (\bibinfo {year} {2022})}\BibitemShut {NoStop}%
\bibitem [{\citenamefont {van Meegen}\ and\ \citenamefont {Sompolinsky}(2024)}]{van2024coding}%
  \BibitemOpen
  \bibfield  {author} {\bibinfo {author} {\bibfnamefont {A.}~\bibnamefont {van Meegen}}\ and\ \bibinfo {author} {\bibfnamefont {H.}~\bibnamefont {Sompolinsky}},\ }\bibfield  {title} {\bibinfo {title} {Coding schemes in neural networks learning classification tasks},\ }\href@noop {} {\bibfield  {journal} {\bibinfo  {journal} {arXiv preprint arXiv:2406.16689}\ } (\bibinfo {year} {2024})}\BibitemShut {NoStop}%
\bibitem [{\citenamefont {Woodworth}\ \emph {et~al.}(2020)\citenamefont {Woodworth}, \citenamefont {Gunasekar}, \citenamefont {Lee}, \citenamefont {Moroshko}, \citenamefont {Savarese}, \citenamefont {Golan}, \citenamefont {Soudry},\ and\ \citenamefont {Srebro}}]{woodworth2020kernel}%
  \BibitemOpen
  \bibfield  {author} {\bibinfo {author} {\bibfnamefont {B.}~\bibnamefont {Woodworth}}, \bibinfo {author} {\bibfnamefont {S.}~\bibnamefont {Gunasekar}}, \bibinfo {author} {\bibfnamefont {J.~D.}\ \bibnamefont {Lee}}, \bibinfo {author} {\bibfnamefont {E.}~\bibnamefont {Moroshko}}, \bibinfo {author} {\bibfnamefont {P.}~\bibnamefont {Savarese}}, \bibinfo {author} {\bibfnamefont {I.}~\bibnamefont {Golan}}, \bibinfo {author} {\bibfnamefont {D.}~\bibnamefont {Soudry}},\ and\ \bibinfo {author} {\bibfnamefont {N.}~\bibnamefont {Srebro}},\ }\bibfield  {title} {\bibinfo {title} {Kernel and rich regimes in overparametrized models},\ }in\ \href@noop {} {\emph {\bibinfo {booktitle} {Conference on Learning Theory}}}\ (\bibinfo {organization} {PMLR},\ \bibinfo {year} {2020})\ pp.\ \bibinfo {pages} {3635--3673}\BibitemShut {NoStop}%
\bibitem [{\citenamefont {Azulay}\ \emph {et~al.}(2021)\citenamefont {Azulay}, \citenamefont {Moroshko}, \citenamefont {Nacson}, \citenamefont {Woodworth}, \citenamefont {Srebro}, \citenamefont {Globerson},\ and\ \citenamefont {Soudry}}]{azulay2021implicit}%
  \BibitemOpen
  \bibfield  {author} {\bibinfo {author} {\bibfnamefont {S.}~\bibnamefont {Azulay}}, \bibinfo {author} {\bibfnamefont {E.}~\bibnamefont {Moroshko}}, \bibinfo {author} {\bibfnamefont {M.~S.}\ \bibnamefont {Nacson}}, \bibinfo {author} {\bibfnamefont {B.~E.}\ \bibnamefont {Woodworth}}, \bibinfo {author} {\bibfnamefont {N.}~\bibnamefont {Srebro}}, \bibinfo {author} {\bibfnamefont {A.}~\bibnamefont {Globerson}},\ and\ \bibinfo {author} {\bibfnamefont {D.}~\bibnamefont {Soudry}},\ }\bibfield  {title} {\bibinfo {title} {On the implicit bias of initialization shape: Beyond infinitesimal mirror descent},\ }in\ \href@noop {} {\emph {\bibinfo {booktitle} {International Conference on Machine Learning}}}\ (\bibinfo {organization} {PMLR},\ \bibinfo {year} {2021})\ pp.\ \bibinfo {pages} {468--477}\BibitemShut {NoStop}%
\bibitem [{\citenamefont {Pacelli}\ \emph {et~al.}(2023)\citenamefont {Pacelli}, \citenamefont {Ariosto}, \citenamefont {Pastore}, \citenamefont {Ginelli}, \citenamefont {Gherardi},\ and\ \citenamefont {Rotondo}}]{pacelli2023statistical}%
  \BibitemOpen
  \bibfield  {author} {\bibinfo {author} {\bibfnamefont {R.}~\bibnamefont {Pacelli}}, \bibinfo {author} {\bibfnamefont {S.}~\bibnamefont {Ariosto}}, \bibinfo {author} {\bibfnamefont {M.}~\bibnamefont {Pastore}}, \bibinfo {author} {\bibfnamefont {F.}~\bibnamefont {Ginelli}}, \bibinfo {author} {\bibfnamefont {M.}~\bibnamefont {Gherardi}},\ and\ \bibinfo {author} {\bibfnamefont {P.}~\bibnamefont {Rotondo}},\ }\bibfield  {title} {\bibinfo {title} {A statistical mechanics framework for bayesian deep neural networks beyond the infinite-width limit},\ }\href@noop {} {\bibfield  {journal} {\bibinfo  {journal} {Nature Machine Intelligence}\ }\textbf {\bibinfo {volume} {5}},\ \bibinfo {pages} {1497} (\bibinfo {year} {2023})}\BibitemShut {NoStop}%
\bibitem [{\citenamefont {Cui}\ \emph {et~al.}(2023)\citenamefont {Cui}, \citenamefont {Krzakala},\ and\ \citenamefont {Zdeborov{\'a}}}]{cui2023bayes}%
  \BibitemOpen
  \bibfield  {author} {\bibinfo {author} {\bibfnamefont {H.}~\bibnamefont {Cui}}, \bibinfo {author} {\bibfnamefont {F.}~\bibnamefont {Krzakala}},\ and\ \bibinfo {author} {\bibfnamefont {L.}~\bibnamefont {Zdeborov{\'a}}},\ }\bibfield  {title} {\bibinfo {title} {Bayes-optimal learning of deep random networks of extensive-width},\ }in\ \href@noop {} {\emph {\bibinfo {booktitle} {International Conference on Machine Learning}}}\ (\bibinfo {organization} {PMLR},\ \bibinfo {year} {2023})\ pp.\ \bibinfo {pages} {6468--6521}\BibitemShut {NoStop}%
\bibitem [{\citenamefont {Fischer}\ \emph {et~al.}(2024)\citenamefont {Fischer}, \citenamefont {Lindner}, \citenamefont {Dahmen}, \citenamefont {Ringel}, \citenamefont {Kr{\"a}mer},\ and\ \citenamefont {Helias}}]{fischer2024critical}%
  \BibitemOpen
  \bibfield  {author} {\bibinfo {author} {\bibfnamefont {K.}~\bibnamefont {Fischer}}, \bibinfo {author} {\bibfnamefont {J.}~\bibnamefont {Lindner}}, \bibinfo {author} {\bibfnamefont {D.}~\bibnamefont {Dahmen}}, \bibinfo {author} {\bibfnamefont {Z.}~\bibnamefont {Ringel}}, \bibinfo {author} {\bibfnamefont {M.}~\bibnamefont {Kr{\"a}mer}},\ and\ \bibinfo {author} {\bibfnamefont {M.}~\bibnamefont {Helias}},\ }\bibfield  {title} {\bibinfo {title} {Critical feature learning in deep neural networks},\ }\href@noop {} {\bibfield  {journal} {\bibinfo  {journal} {arXiv preprint arXiv:2405.10761}\ } (\bibinfo {year} {2024})}\BibitemShut {NoStop}%
\bibitem [{\citenamefont {Rubin}\ \emph {et~al.}(2024)\citenamefont {Rubin}, \citenamefont {Ringel}, \citenamefont {Seroussi},\ and\ \citenamefont {Helias}}]{rubin2024unified}%
  \BibitemOpen
  \bibfield  {author} {\bibinfo {author} {\bibfnamefont {N.}~\bibnamefont {Rubin}}, \bibinfo {author} {\bibfnamefont {Z.}~\bibnamefont {Ringel}}, \bibinfo {author} {\bibfnamefont {I.}~\bibnamefont {Seroussi}},\ and\ \bibinfo {author} {\bibfnamefont {M.}~\bibnamefont {Helias}},\ }\bibfield  {title} {\bibinfo {title} {A unified approach to feature learning in bayesian neural networks},\ }in\ \href@noop {} {\emph {\bibinfo {booktitle} {High-dimensional Learning Dynamics 2024: The Emergence of Structure and Reasoning}}}\ (\bibinfo {year} {2024})\BibitemShut {NoStop}%
\bibitem [{\citenamefont {M{\'e}zard}\ \emph {et~al.}(1987)\citenamefont {M{\'e}zard}, \citenamefont {Parisi},\ and\ \citenamefont {Virasoro}}]{mezard1987spin}%
  \BibitemOpen
  \bibfield  {author} {\bibinfo {author} {\bibfnamefont {M.}~\bibnamefont {M{\'e}zard}}, \bibinfo {author} {\bibfnamefont {G.}~\bibnamefont {Parisi}},\ and\ \bibinfo {author} {\bibfnamefont {M.~A.}\ \bibnamefont {Virasoro}},\ }\href@noop {} {\emph {\bibinfo {title} {Spin glass theory and beyond: An Introduction to the Replica Method and Its Applications}}},\ Vol.~\bibinfo {volume} {9}\ (\bibinfo  {publisher} {World Scientific Publishing Company},\ \bibinfo {year} {1987})\BibitemShut {NoStop}%
\bibitem [{\citenamefont {Franz}\ \emph {et~al.}(1992)\citenamefont {Franz}, \citenamefont {Parisi},\ and\ \citenamefont {Virasoro}}]{franz1992replica}%
  \BibitemOpen
  \bibfield  {author} {\bibinfo {author} {\bibfnamefont {S.}~\bibnamefont {Franz}}, \bibinfo {author} {\bibfnamefont {G.}~\bibnamefont {Parisi}},\ and\ \bibinfo {author} {\bibfnamefont {M.~A.}\ \bibnamefont {Virasoro}},\ }\bibfield  {title} {\bibinfo {title} {The replica method on and off equilibrium},\ }\href@noop {} {\bibfield  {journal} {\bibinfo  {journal} {Journal de Physique I}\ }\textbf {\bibinfo {volume} {2}},\ \bibinfo {pages} {1869} (\bibinfo {year} {1992})}\BibitemShut {NoStop}%
\bibitem [{\citenamefont {Gardner}(1988)}]{gardner1988space}%
  \BibitemOpen
  \bibfield  {author} {\bibinfo {author} {\bibfnamefont {E.}~\bibnamefont {Gardner}},\ }\bibfield  {title} {\bibinfo {title} {The space of interactions in neural network models},\ }\href@noop {} {\bibfield  {journal} {\bibinfo  {journal} {Journal of physics A: Mathematical and general}\ }\textbf {\bibinfo {volume} {21}},\ \bibinfo {pages} {257} (\bibinfo {year} {1988})}\BibitemShut {NoStop}%
\bibitem [{\citenamefont {Gabri{\'e}}\ \emph {et~al.}(2018)\citenamefont {Gabri{\'e}}, \citenamefont {Manoel}, \citenamefont {Luneau}, \citenamefont {Macris}, \citenamefont {Krzakala}, \citenamefont {Zdeborov{\'a}} \emph {et~al.}}]{gabrie2018entropy}%
  \BibitemOpen
  \bibfield  {author} {\bibinfo {author} {\bibfnamefont {M.}~\bibnamefont {Gabri{\'e}}}, \bibinfo {author} {\bibfnamefont {A.}~\bibnamefont {Manoel}}, \bibinfo {author} {\bibfnamefont {C.}~\bibnamefont {Luneau}}, \bibinfo {author} {\bibfnamefont {N.}~\bibnamefont {Macris}}, \bibinfo {author} {\bibfnamefont {F.}~\bibnamefont {Krzakala}}, \bibinfo {author} {\bibfnamefont {L.}~\bibnamefont {Zdeborov{\'a}}}, \emph {et~al.},\ }\bibfield  {title} {\bibinfo {title} {Entropy and mutual information in models of deep neural networks},\ }\href@noop {} {\bibfield  {journal} {\bibinfo  {journal} {Advances in Neural Information Processing Systems}\ }\textbf {\bibinfo {volume} {31}} (\bibinfo {year} {2018})}\BibitemShut {NoStop}%
\bibitem [{\citenamefont {Carleo}\ \emph {et~al.}(2019)\citenamefont {Carleo}, \citenamefont {Cirac}, \citenamefont {Cranmer}, \citenamefont {Daudet}, \citenamefont {Schuld}, \citenamefont {Tishby}, \citenamefont {Vogt-Maranto},\ and\ \citenamefont {Zdeborov{\'a}}}]{carleo2019machine}%
  \BibitemOpen
  \bibfield  {author} {\bibinfo {author} {\bibfnamefont {G.}~\bibnamefont {Carleo}}, \bibinfo {author} {\bibfnamefont {I.}~\bibnamefont {Cirac}}, \bibinfo {author} {\bibfnamefont {K.}~\bibnamefont {Cranmer}}, \bibinfo {author} {\bibfnamefont {L.}~\bibnamefont {Daudet}}, \bibinfo {author} {\bibfnamefont {M.}~\bibnamefont {Schuld}}, \bibinfo {author} {\bibfnamefont {N.}~\bibnamefont {Tishby}}, \bibinfo {author} {\bibfnamefont {L.}~\bibnamefont {Vogt-Maranto}},\ and\ \bibinfo {author} {\bibfnamefont {L.}~\bibnamefont {Zdeborov{\'a}}},\ }\bibfield  {title} {\bibinfo {title} {Machine learning and the physical sciences},\ }\href@noop {} {\bibfield  {journal} {\bibinfo  {journal} {Reviews of Modern Physics}\ }\textbf {\bibinfo {volume} {91}},\ \bibinfo {pages} {045002} (\bibinfo {year} {2019})}\BibitemShut {NoStop}%
\bibitem [{\citenamefont {Bahri}\ \emph {et~al.}(2020)\citenamefont {Bahri}, \citenamefont {Kadmon}, \citenamefont {Pennington}, \citenamefont {Schoenholz}, \citenamefont {Sohl-Dickstein},\ and\ \citenamefont {Ganguli}}]{bahri2020statistical}%
  \BibitemOpen
  \bibfield  {author} {\bibinfo {author} {\bibfnamefont {Y.}~\bibnamefont {Bahri}}, \bibinfo {author} {\bibfnamefont {J.}~\bibnamefont {Kadmon}}, \bibinfo {author} {\bibfnamefont {J.}~\bibnamefont {Pennington}}, \bibinfo {author} {\bibfnamefont {S.~S.}\ \bibnamefont {Schoenholz}}, \bibinfo {author} {\bibfnamefont {J.}~\bibnamefont {Sohl-Dickstein}},\ and\ \bibinfo {author} {\bibfnamefont {S.}~\bibnamefont {Ganguli}},\ }\bibfield  {title} {\bibinfo {title} {Statistical mechanics of deep learning},\ }\href@noop {} {\bibfield  {journal} {\bibinfo  {journal} {Annual Review of Condensed Matter Physics}\ }\textbf {\bibinfo {volume} {11}},\ \bibinfo {pages} {501} (\bibinfo {year} {2020})}\BibitemShut {NoStop}%
\bibitem [{\citenamefont {Saglietti}\ and\ \citenamefont {Zdeborov{\'a}}(2022)}]{saglietti2022solvable}%
  \BibitemOpen
  \bibfield  {author} {\bibinfo {author} {\bibfnamefont {L.}~\bibnamefont {Saglietti}}\ and\ \bibinfo {author} {\bibfnamefont {L.}~\bibnamefont {Zdeborov{\'a}}},\ }\bibfield  {title} {\bibinfo {title} {Solvable model for inheriting the regularization through knowledge distillation},\ }in\ \href@noop {} {\emph {\bibinfo {booktitle} {Mathematical and Scientific Machine Learning}}}\ (\bibinfo {organization} {PMLR},\ \bibinfo {year} {2022})\ pp.\ \bibinfo {pages} {809--846}\BibitemShut {NoStop}%
\bibitem [{\citenamefont {Parikh}\ \emph {et~al.}(2014)\citenamefont {Parikh}, \citenamefont {Boyd} \emph {et~al.}}]{parikh2014proximal}%
  \BibitemOpen
  \bibfield  {author} {\bibinfo {author} {\bibfnamefont {N.}~\bibnamefont {Parikh}}, \bibinfo {author} {\bibfnamefont {S.}~\bibnamefont {Boyd}}, \emph {et~al.},\ }\bibfield  {title} {\bibinfo {title} {Proximal algorithms},\ }\href@noop {} {\bibfield  {journal} {\bibinfo  {journal} {Foundations and trends{\textregistered} in Optimization}\ }\textbf {\bibinfo {volume} {1}},\ \bibinfo {pages} {127} (\bibinfo {year} {2014})}\BibitemShut {NoStop}%
\bibitem [{\citenamefont {Polson}\ \emph {et~al.}(2015)\citenamefont {Polson}, \citenamefont {Scott},\ and\ \citenamefont {Willard}}]{polson2015proximal}%
  \BibitemOpen
  \bibfield  {author} {\bibinfo {author} {\bibfnamefont {N.~G.}\ \bibnamefont {Polson}}, \bibinfo {author} {\bibfnamefont {J.~G.}\ \bibnamefont {Scott}},\ and\ \bibinfo {author} {\bibfnamefont {B.~T.}\ \bibnamefont {Willard}},\ }\bibfield  {title} {\bibinfo {title} {Proximal algorithms in statistics and machine learning},\ }\href@noop {} {\bibfield  {journal} {\bibinfo  {journal} {arXiv preprint arXiv:1502.07944}\ } (\bibinfo {year} {2015})}\BibitemShut {NoStop}%
\bibitem [{\citenamefont {Teboulle}(1997)}]{teboulle1997convergence}%
  \BibitemOpen
  \bibfield  {author} {\bibinfo {author} {\bibfnamefont {M.}~\bibnamefont {Teboulle}},\ }\bibfield  {title} {\bibinfo {title} {Convergence of proximal-like algorithms},\ }\href@noop {} {\bibfield  {journal} {\bibinfo  {journal} {SIAM Journal on Optimization}\ }\textbf {\bibinfo {volume} {7}},\ \bibinfo {pages} {1069} (\bibinfo {year} {1997})}\BibitemShut {NoStop}%
\bibitem [{\citenamefont {Drusvyatskiy}\ and\ \citenamefont {Lewis}(2018)}]{drusvyatskiy2018error}%
  \BibitemOpen
  \bibfield  {author} {\bibinfo {author} {\bibfnamefont {D.}~\bibnamefont {Drusvyatskiy}}\ and\ \bibinfo {author} {\bibfnamefont {A.~S.}\ \bibnamefont {Lewis}},\ }\bibfield  {title} {\bibinfo {title} {Error bounds, quadratic growth, and linear convergence of proximal methods},\ }\href@noop {} {\bibfield  {journal} {\bibinfo  {journal} {Mathematics of Operations Research}\ }\textbf {\bibinfo {volume} {43}},\ \bibinfo {pages} {919} (\bibinfo {year} {2018})}\BibitemShut {NoStop}%
\bibitem [{\citenamefont {Robbins}\ and\ \citenamefont {Monro}(1951)}]{robbins1951stochastic}%
  \BibitemOpen
  \bibfield  {author} {\bibinfo {author} {\bibfnamefont {H.}~\bibnamefont {Robbins}}\ and\ \bibinfo {author} {\bibfnamefont {S.}~\bibnamefont {Monro}},\ }\bibfield  {title} {\bibinfo {title} {A stochastic approximation method},\ }\href@noop {} {\bibfield  {journal} {\bibinfo  {journal} {The annals of mathematical statistics}\ ,\ \bibinfo {pages} {400}} (\bibinfo {year} {1951})}\BibitemShut {NoStop}%
\bibitem [{\citenamefont {Amari}(1998)}]{amari1998natural}%
  \BibitemOpen
  \bibfield  {author} {\bibinfo {author} {\bibfnamefont {S.-I.}\ \bibnamefont {Amari}},\ }\bibfield  {title} {\bibinfo {title} {Natural gradient works efficiently in learning},\ }\href@noop {} {\bibfield  {journal} {\bibinfo  {journal} {Neural computation}\ }\textbf {\bibinfo {volume} {10}},\ \bibinfo {pages} {251} (\bibinfo {year} {1998})}\BibitemShut {NoStop}%
\bibitem [{\citenamefont {Beck}\ and\ \citenamefont {Teboulle}(2003)}]{beck2003mirror}%
  \BibitemOpen
  \bibfield  {author} {\bibinfo {author} {\bibfnamefont {A.}~\bibnamefont {Beck}}\ and\ \bibinfo {author} {\bibfnamefont {M.}~\bibnamefont {Teboulle}},\ }\bibfield  {title} {\bibinfo {title} {Mirror descent and nonlinear projected subgradient methods for convex optimization},\ }\href@noop {} {\bibfield  {journal} {\bibinfo  {journal} {Operations Research Letters}\ }\textbf {\bibinfo {volume} {31}},\ \bibinfo {pages} {167} (\bibinfo {year} {2003})}\BibitemShut {NoStop}%
\bibitem [{\citenamefont {Bae}\ \emph {et~al.}(2022)\citenamefont {Bae}, \citenamefont {Vicol}, \citenamefont {HaoChen},\ and\ \citenamefont {Grosse}}]{bae2022amortized}%
  \BibitemOpen
  \bibfield  {author} {\bibinfo {author} {\bibfnamefont {J.}~\bibnamefont {Bae}}, \bibinfo {author} {\bibfnamefont {P.}~\bibnamefont {Vicol}}, \bibinfo {author} {\bibfnamefont {J.~Z.}\ \bibnamefont {HaoChen}},\ and\ \bibinfo {author} {\bibfnamefont {R.~B.}\ \bibnamefont {Grosse}},\ }\bibfield  {title} {\bibinfo {title} {Amortized proximal optimization},\ }\href@noop {} {\bibfield  {journal} {\bibinfo  {journal} {Advances in Neural Information Processing Systems}\ }\textbf {\bibinfo {volume} {35}},\ \bibinfo {pages} {8982} (\bibinfo {year} {2022})}\BibitemShut {NoStop}%
\bibitem [{\citenamefont {Shan}\ \emph {et~al.}(2024)\citenamefont {Shan}, \citenamefont {Li},\ and\ \citenamefont {Sompolinsky}}]{shan2024order}%
  \BibitemOpen
  \bibfield  {author} {\bibinfo {author} {\bibfnamefont {H.}~\bibnamefont {Shan}}, \bibinfo {author} {\bibfnamefont {Q.}~\bibnamefont {Li}},\ and\ \bibinfo {author} {\bibfnamefont {H.}~\bibnamefont {Sompolinsky}},\ }\bibfield  {title} {\bibinfo {title} {Order parameters and phase transitions of continual learning in deep neural networks},\ }\href@noop {} {\bibfield  {journal} {\bibinfo  {journal} {arXiv preprint arXiv:2407.10315}\ } (\bibinfo {year} {2024})}\BibitemShut {NoStop}%
\bibitem [{\citenamefont {Franz}\ and\ \citenamefont {Parisi}(1998)}]{franz1998effective}%
  \BibitemOpen
  \bibfield  {author} {\bibinfo {author} {\bibfnamefont {S.}~\bibnamefont {Franz}}\ and\ \bibinfo {author} {\bibfnamefont {G.}~\bibnamefont {Parisi}},\ }\bibfield  {title} {\bibinfo {title} {Effective potential in glassy systems: theory and simulations},\ }\href@noop {} {\bibfield  {journal} {\bibinfo  {journal} {Physica A: Statistical Mechanics and its Applications}\ }\textbf {\bibinfo {volume} {261}},\ \bibinfo {pages} {317} (\bibinfo {year} {1998})}\BibitemShut {NoStop}%
\end{thebibliography}%

\appendix

\clearpage
\onecolumngrid
\setcounter{page}{1} 
\setcounter{section}{0} 
\setcounter{equation}{0} 
\begin{center} 	
    {\Large\textbf{Supplemental Information}}
\end{center}

\section{Markov Proximal Learning \label{sec:SI Markov-proximal-learning}}
We introduce a Markov Proximal Learning (MPL) framework
for learning dynamics in fully connected Deep Neural Networks (DNNs). This method allows us to construct a dynamical mean field theory for Langevin dynamics in the infinite width limit, and is a novel way to discritize Langevin dynamics and formulate out-of-equilibrium statistical mechanics.
We formally write down the moment-generating function (MGF) of the
predictor. We then use the well-known replica method in statistical
physics \cite{mezard1987spin,franz1992replica}, which has also been
shown to be a powerful tool for deriving analytical results for learning
in NNs \cite{gardner1988space,gabrie2018entropy,carleo2019machine,bahri2020statistical, saglietti2022solvable}.
We analytically calculate the MGF after averaging over the posterior
distribution of the network weights in the infinite width limit, which
enables us to compute statistics of the predictor.

\subsection{Definition of Markov Proximal Learning \label{subsec:Definition of Markov Proximal Learning}}

We consider the network learning dynamics as a Markov proximal process,
which is a generalized version of the \emph{deterministic} proximal
algorithm (\cite{parikh2014proximal,polson2015proximal}). Deterministic
proximal algorithm with $L_{2}$ regularization is a sequential update
rule defined as 
\begin{equation}
    \Theta_{t}\left(\Theta_{t-1},\mathcal{D}\right)=\arg\min_{\Theta}\left(E\left(\Theta|\mathcal{D}\right)+\frac{\lambda}{2}\left|\Theta-\Theta_{t-1}\right|^{2}\right)
\end{equation}
where $\lambda$ is a parameter determining the strength of the proximity
constraint. This algorithm has been proven to converge to the global
minimum for convex cost functions \cite{teboulle1997convergence,drusvyatskiy2018error},
and many optimization algorithms widely used in machine learning can
be seen as its approximations \cite{robbins1951stochastic, amari1998natural,beck2003mirror,bae2022amortized}. 
We define a stochastic extension of proximal learning, the Markov
proximal learning. This method was also inspired by continual learning methods \cite{shan2024order} and Franz-Parisi potential \cite{franz1998effective}. The process is characterized by the following transition matrix 
\begin{equation}
\mathcal{T}\left(\Theta_{t}|\Theta_{t-1}\right)=\frac{1}{Z\left(\Theta_{t-1}\right)}\exp\left(-\frac{1}{2}\beta\left(E\left(\Theta_{t}\right)+\frac{\lambda}{2}\left|\Theta_{t}-\Theta_{t-1}\right|^{2}\right)\right)\label{eq:transitiondensity}
\end{equation}
where $Z\left(\Theta_{t-1}\right)$ is the single-time partition function, which imposes normalization throughout the Markov process,
$Z\left(\Theta_{t-1}\right)=\intop d\Theta^{\prime}\exp\left(-\frac{1}{2}\beta\left(E\left(\Theta^{\prime}\right)+\frac{\lambda}{2}\left|\Theta^{\prime}-\Theta_{t-1}\right|^{2}\right)\right)$ $\beta$ is an inverse temperature parameter characterizing
the level of 'uncertainty' and $\beta\rightarrow\infty$ limit recovers
the deterministic proximal algorithm. We note that in the large $\lambda$ limit,
the difference between $\Theta_{t}$ and $\Theta_{t-1}$ is infinitesimal,
and $\Theta_{t}$ becomes a smooth function of continuous time, where the time variable is the discrete time divided by $\lambda$. 

The joint probability of the parameters is given by $(\Theta_{0},\Theta_{1},...,\Theta_{t}).$

\begin{equation}
P\left(\Theta_{0},\Theta_{1},...,\Theta_{t}\right)=\left[\prod_{\tau=1}^{t}\mathcal{T}\left(\Theta_{\tau}|\Theta_{\tau-1}\right)\right]P\left(\Theta_{0}\right)
\end{equation}

where $P\left(\Theta_{0}\right)$ is the distribution of the
initial condition of the parameters.

\subsection{Large $\lambda$ Limit and Langevin dynamics:}

We prove that in the limit of large $\lambda$ and differentiable cost
function this algorithm is equivalent to Langevin dynamics. We define $\delta\Theta_{t}=\Theta_{t}-\Theta_{t-1}$
. In the limit of large $\lambda$, we can expand the transition matrix
around $\delta\Theta_{t}=0$: 
\begin{align}
\mathcal{T}\left(\delta\Theta_{t}|\Theta_{t-1}\right) & \approx\left(\frac{\lambda\beta}{4\pi}\right)^{\frac{d}{2}}\exp\left[-\frac{\lambda\beta}{4}\left|\delta\Theta_{t}+\frac{1}{\lambda}\nabla E\left(\Theta_{t-1}\right)\right|^{2}\right]
\end{align}

$\delta\Theta_{t}|\Theta_{t-1}$ is a Gaussian random variable with the statistics: 
\begin{equation}
\left\langle \delta\Theta_{t}|\Theta_{t-1}\right\rangle =-\frac{1}{\lambda}\nabla_\Theta E\left(\Theta_{t-1}\right)
\end{equation}
\begin{equation}
\text{var}\left(\delta\Theta_{t}\delta\Theta_{t^{\prime}}^{\top}|\Theta_{t-1}\right)=\frac{2}{\lambda\beta}\delta_{t,t^{\prime}} I
\end{equation}

which is equivalent to Langevin dynamics in Itô discretization: 
\begin{equation}
\delta\Theta_{t}=\left(-\nabla_\Theta E\left(\Theta_{t-1}\right)+\eta_{t-1}\right)dt\label{eq:langevin}
\end{equation}

with 
\begin{equation}
\left\langle \eta_{t}\eta_{t^{\prime}}^{\top}\right\rangle =\frac{2T}{dt}\delta_{t,t^{\prime}}I,\left\langle \eta_{t}\right\rangle =0
\end{equation}

where $\frac{1}{\lambda}=dt$,$\beta=\frac{1}{T}$.

\section{The Statistics of the Predictor\label{sec:Replica-calculation-of}}

\subsection{Replica Calculation of the Moment-Generating Function of the
Predictor}

The transition density can be written using the replica method, where
$Z^{-1}\left(\Theta_{t-1}\right)=\lim_{n\rightarrow0}Z^{n-1}\left(\Theta_{t-1}\right)$,:
\begin{align}
\mathcal{T}\left(\Theta_{t}|\Theta_{t-1}\right) & =\lim_{n\rightarrow0}Z^{n-1}\left(\Theta_{t-1}\right)\exp\left(-\frac{1}{2}\beta\left(E\left(\Theta_{t}\right)+\frac{\lambda}{2}\left|\Theta_{t}-\Theta_{t-1}\right|^{2}\right)\right) \\
 & =\lim_{n\rightarrow0}\prod_{\alpha=1}^{n-1}\intop d\Theta_{t}^{\alpha}\exp\left(-\frac{\beta}{2}\left(\sum_{\alpha=1}^{n}E\left(\Theta_{t}^{\alpha}\right)+\frac{\lambda}{2}\sum_{\alpha=1}^{n}\left|\Theta_{t}^{\alpha}-\Theta_{t-1}^{n}\right|^{2}\right)\right)\nonumber
\end{align}

Here $\alpha=1,\cdots,n-1$ are the 'replicated copies' of the physical
variable $\left\{ \Theta_{\tau}^{n}\right\} _{\tau=1,\cdots,t}$.
To calculate the statistics of the dynamical process, we consider
the MGF for arbitrary functions of the trajectory $g(\left\{ \Theta_{\tau}^{n}\right\} _{\tau=0,\cdots t})$
\begin{align}
& \mathcal{M}\left[\ell\right] = \left\langle \exp\left(\sum_{t=1}^{\infty}\ell_{t}g\left(\left\{ \Theta_{\tau}^{n}\right\} _{\tau=0,\dots,t}\right)\right)\right\rangle _{\Theta}\label{eq:moment-1}\\ & =\prod_{\tau=0}^{\infty}\intop d\Theta_{\tau}\left[\prod_{\tau=1}^{\infty}\mathcal{T}\left(\Theta_{\tau}|\Theta_{\tau-1}\right)\right]P\left(\Theta_{0}\right)\exp\left(\sum_{t=1}^{\infty}\ell_{t}g\left(\left\{ \Theta_{\tau}^{n}\right\} _{\tau=0,...t}\right)\right) \nonumber\\
 & =\lim_{n\rightarrow0}\prod_{\alpha=1}^{n}\prod_{\tau=1}^{\infty}\intop d\Theta_{t}^{\alpha}\intop d\Theta_{0}^{n}P\left(\Theta_{0}^{n}\right)\nonumber \\
 & \exp\left(-\frac{\beta}{2}\sum_{\tau=1}^{\infty}\left(\sum_{\alpha=1}^{n}E\left(\Theta_{\tau}^{\alpha}\right)+\frac{\lambda}{2}\sum_{\alpha=1}^{n}\left|\Theta_{\tau}^{\alpha}-\Theta_{\tau-1}^{n}\right|^{2}\right)+\sum_{t=1}^{\infty}\ell_{t}g\left(\left\{ \Theta_{\tau}^{n}\right\} _{\tau=0,\cdots t}\right)\right)\nonumber
\end{align}

We apply this formalism to the supervised learning cost function introduced in Sec.$\text{\ref{subsec:notations}}$ in the main text.
\begin{equation}
E\left(\Theta_{t}|\mathcal{D}\right)=\frac{1}{2}\sum_{\mu=1}^{P}\left(f\left({\bf x}^{\mu},\Theta_{t}\right)-y^{\mu}\right)^{2}+\frac{T}{2\sigma^{2}}\left|\Theta_{t}\right|^{2}
\end{equation}

and the predictor statistics at time $t$, $g(\left\{ \Theta_{\tau}^{n}\right\} _{\tau=0,\cdots t})=f\left({\bf x},\Theta_{t}^{n}\right),$yielding
\begin{align}
\mathcal{M}\left[\ell\right] & =\lim_{n\rightarrow0}\prod_{\alpha=1}^{n}\prod_{\tau=1}^{\infty}\intop d\Theta_{\tau}^{\alpha}\int d\Theta_{0}\exp\left(-\frac{\beta}{4}\sum_{\tau=1}^{\infty}\sum_{\alpha=1}^{n}\left(f_{\text{train}}\left(\Theta_{\tau}^{\alpha}\right)-Y\right)^{2}+\sum_{t=1}^\infty\sum_{\bf x}\ell_{t,{\bf x}}f\left({\bf x},\Theta_{t}^{n}\right)-S_{0}\left[\Theta\right]\right)
\end{align}
\begin{equation}
S_{0}\left[\Theta\right]=\frac{1}{4}\sum_{\tau=1}^{\infty}\sum_{\alpha=1}^{n}\left(\sigma^{-2}\left|\Theta{}_{\tau}^{\alpha}\right|^{2}+\lambda\beta\left|\Theta_{\tau}^{\alpha}-\Theta_{\tau-1}^{n}\right|^{2}\right)+\frac{1}{2}\sigma_{0}^{-2}\left|\Theta_{0}^{n}\right|^{2}\label{eq:gaussianprior}
\end{equation}

Where we define $f_{\text{train}}\left(\Theta_{\tau}^{\alpha}\right)\equiv\left[f\left({\bf x}^{1},\Theta_{\tau}^{\alpha}\right),\cdots,f\left({\bf x}^{P},\Theta_{\tau}^{\alpha}\right)\right]^{T}\in\mathbb{R}^{P}$
a vector contains the predictor on the training dataset, and $Y\in\mathbb{R}^{P}$
such that $Y^{\mu}=y^{\mu}$, similar to Sec.\ref{subsec:notations}. $S_{0}\left[\Theta\right]$ denote the
Gaussian prior on the parameters including the hidden layer weights
and the readout weights.

To perform the integration over ${\bf a}_{\tau}^\alpha $,
we use Hubbard-Stratonovich (H.S.) transformation and introduce a
new vector field $v_{\tau}^{\alpha}\in\mathbb{R}^{P}$ 
\begin{align}
\mathcal{M}\left[\ell\right] & =\lim_{n\rightarrow0}\prod_{\alpha=1}^{n}\prod_{\tau=1}^{\infty}\intop d\Theta_{\tau}^{\alpha}\intop dv_{\tau}^{\alpha}\int d\Theta_{0}\label{eq:hstransform}\\
 & \exp\left(-\frac{i\beta}{2}\sum_{\tau=1}^{\infty}\sum_{\alpha=1}^{n}\left(\frac{1}{\sqrt{N_{L}}}f_{\text{train}}\left(\Theta^\alpha_\tau\right)-Y\right)^{\top}v_{\tau}^{\alpha}\right.\left.-\frac{\beta}{4}\sum_{\tau=1}^{\infty}\sum_{\alpha=1}^{n}\left|v_{\tau}^{\alpha}\right|^{2}+\sum_{t=1}^{\infty}\sum_{{\bf x}}\ell_{t,{\bf x}}f\left({\bf x},\Theta_{t}^{n}\right)-S_{0}\left[\Theta\right]\right)\nonumber 
\end{align}

\textbf{Averaging over the readout weights:}

We integrate over ${\bf a}_{\tau}^{\alpha}$. For convenience, we denote the set of all hidden layer weights collectively as ${\bf W}_t=\left\{ {\bf W}_{t}^{\ell=1},\dots,{\bf W}_{t}^{L}\right\} $, similar to the main text.

\begin{align}
\mathcal{M}\left[\ell\right]= & \lim_{n\rightarrow0}\prod_{\tau=1}^{\infty}\prod_{\alpha=1}^{n}\intop dv_{\tau}^{\alpha}\intop d{\bf W}_{\tau}^{\alpha}\exp\left(-S\left[v,\bf{W}\right]-Q\left[\ell,v,\bf{W}\right]-S_{0}\left[\bf{W}\right]\right)\label{eq:finite-lambda-MGF}
\end{align}
\begin{align}
S\left[v,\bf{W}\right] & =\frac{\beta}{4}\left(\sum_{\alpha,\beta=1}^{n}\sum_{\tau=1}^{\infty}\frac{\beta}{2}v_{\tau}^{\alpha\top}m_{\tau,\tau^{\prime}}^{\alpha\beta}K_{\tau,\tau^{\prime}}^{L,\alpha\beta}\left(\bf{W}_{\tau}^{\alpha}\right)v_{\tau^{\prime}}^{\beta}+\sum_{\alpha=1}^{n}\sum_{\tau=1}^{\infty}\left(v_{\tau}^{\alpha}-2iY\right)^{\top}v_{\tau}^{\alpha}\right)\label{eq:action}
\end{align}

and the source term action 
\begin{align}
Q\left[\ell,v,\bf{W}\right]= & i\frac{\beta}{2}\sum_{\alpha=1}^{n}\sum_{t,\tau=1}^{\infty}\sum_{{\bf x}}v_{\tau}^{\alpha\top}m_{t,\tau}^{\alpha n}k_{t,\tau}^{L,\alpha n}\left(\bf{W}_{\tau}^{\alpha},{\bf x}\right)\ell_{t,{\bf x}}-\frac{1}{2}\sum_{t,t^{\prime}=1}^{\infty}\sum_{{\bf x},{\bf x}^{\prime}}m_{t,t^{\prime}}^{nn}K_{t,t^{\prime}}^{L,nn}\left({\bf{W}}_{\tau}^{n},{\bf x},{\bf x}\right)\ell_{t,{\bf x}}\ell_{t^{\prime},{\bf x'}}
\end{align}

Where $m_{\tau,\tau^{\prime}}^{\alpha\beta}$ is a scalar function
independent of the data, and represents the averaging w.r.t. to the
replica dependent prior $S_{0}\left[\Theta\right]$,
such that  $\left\langle \left(\Theta_{\tau}^{\alpha}\right)_{i}\left(\Theta_{\tau^{\prime}}^{\beta}\right)_{j}\right\rangle _{S_{0}}=\delta_{ij}m_{\tau,\tau^{\prime}}^{\alpha\beta}$
\begin{equation}
m_{\tau,\tau^{\prime}}^{\alpha\beta}=\begin{cases}
m_{\tau,\tau^{\prime}}^{1}=\tilde{\sigma}^{2}\left(\tilde{\lambda}^{\left|\tau-\tau^{\prime}\right|}+\gamma\tilde{\lambda}^{\tau+\tau^{\prime}}\right) & \left\{ \alpha=\beta,\tau=\tau^{\prime}\right\} \cup\left\{ \alpha=n,\tau<\tau^{\prime}\right\} \cup\left\{ \beta=n,\tau>\tau^{\prime}\right\} \\
m_{\tau,\tau^{\prime}}^{0}=\tilde{\sigma}^{2}\left(\tilde{\lambda}^{2}\tilde{\lambda}^{\left|\tau-\tau^{\prime}\right|}+\gamma\tilde{\lambda}^{\tau+\tau^{\prime}}\right) & otherwise
\end{cases}\label{eq:mtt}
\end{equation}
Where we have defined new functions of the parameters for convenience,
\begin{equation}
\tilde{\lambda}=\frac{\lambda}{\lambda+T\sigma^{-2}},\tilde{\sigma}^{2}=\sigma^{2}\frac{\lambda+T\sigma^{-2}}{\lambda+\frac{1}{2}T\sigma^{-2}},\gamma=\frac{\sigma_{0}^{2}}{\tilde{\sigma}^{2}}-1\label{eq:params}
\end{equation}

The time-dependent and replica-dependent kernels $K_{\tau,\tau^{\prime}}^{L,\alpha\beta}\in\mathbb{R}^{P\times P},k_{\tau,\tau^{\prime}}^{L,\alpha\beta}\left({\bf x}\right)\in\mathbb{R}^{P},K_{\tau,\tau^{\prime}}^{L,\alpha\beta}\left({\bf x},{\bf x}\right)$ defined as: 
\begin{equation}
\mathcal{\mathcal{K}}_{\tau,\tau^{\prime}}^{L,\alpha\beta}\left({\bf x},{\bf x}^{\prime}\right)=\frac{1}{N_{L}}\left({\bf x}_{\tau}^{L}\left({\bf x},\bf{W}_{\tau}^{\alpha}\right)\cdot{\bf x}_{\tau^{\prime}}^{L}\left({\bf x}^{\prime},\bf{W}_{\tau^{\prime}}^{\beta}\right)\right)\label{eq:replica kernel}
\end{equation}

And $K_{\tau,\tau^{\prime}}^{L,\alpha\beta}\in\mathbb{R}^{P\times P},k_{\tau,\tau^{\prime}}^{L,\alpha\beta}\left({\bf x}\right)\in\mathbb{R}^{P}$
are given by applying the kernel function on the training data and
test data, respectively. 

\textbf{Averaging over the hidden layer weights:}

In the infinite width limit, the statistics of $\bf{W}^\alpha_\tau$ is dominated
by its Gaussian prior (Eq.$\text{\ref{eq:gaussianprior}}$) with zero
mean and covariance $\langle{\bf{W}}_{\tau}^{\alpha}{\bf W}_{\tau'}^{\beta\top}\rangle=m_{\tau,\tau^{\prime}}^{\alpha\beta}I$
.Thus the averaged kernel function $K_{\tau,\tau^{\prime}}^{\alpha\beta}$
(Eq.$\text{\ref{eq:replica kernel}}$) over the prior yields two kinds
of statistics for a given pair of times $\left\{ \tau,\tau^{\prime}\right\} $, which we denote as $\mathcal{K}_{\tau,\tau^{\prime}}^{1,L}\left({\bf x},{\bf x}^{\prime}\right)$,
and $\mathcal{K}_{\tau,\tau^{\prime}}^{0,L}\left({\bf x},{\bf x}^{\prime}\right)$
: 
\begin{equation}
\mathcal{K}_{\tau,\tau^{\prime}}^{\alpha\beta}=\begin{cases}
\mathcal{K}_{\tau,\tau^{\prime}}^{1} & \left\{ \alpha=\beta,\tau=\tau^{\prime}\right\} \cup\left\{ \alpha=n,\tau<\tau^{\prime}\right\} \cup\left\{ \beta=n,\tau>\tau^{\prime}\right\} \\
\mathcal{K}_{\tau,\tau^{\prime}}^{0} & otherwise
\end{cases}
\end{equation}

And they obey the iterative relations: 
\begin{align}
\mathcal{K}_{\tau,\tau^{\prime}}^{1,L}\left({\bf x},{\bf x}^{\prime}\right) & =F\left(m_{\tau,\tau}^{1}\mathcal{K}_{\tau,\tau}^{1,L-1}\left({\bf x},{\bf x}\right),m_{\tau^{\prime},\tau^{\prime}}^{1}\mathcal{K}_{\tau^{\prime},\tau^{\prime}}^{1,L-1}\left({\bf x}^{\prime},{\bf x}^{\prime}\right),m_{\tau,\tau^{\prime}}^{1}\mathcal{K}_{\tau,\tau^{\prime}}^{1,L-1}\left({\bf x},{\bf x}^{\prime}\right)\right)\label{eq:K1}
\end{align}
\begin{align}
\mathcal{K}_{\tau,\tau^{\prime}}^{0,L}\left({\bf x},{\bf x}^{\prime}\right) & =F\left(m_{\tau,\tau}^{1}\mathcal{K}_{\tau,\tau}^{1,L-1}\left({\bf x},{\bf x}\right),m_{\tau^{\prime},\tau^{\prime}}^{1}\mathcal{K}_{\tau^{\prime},\tau^{\prime}}^{1,L-1}\left({\bf x}^{\prime},{\bf x}^{\prime}\right),m_{\tau,\tau^{\prime}}^{0}\mathcal{K}_{\tau,\tau^{\prime}}^{0,L-1}\left({\bf x},{\bf x}^{\prime}\right)\right)\label{eq:K0}
\end{align}
\begin{equation}
\mathcal{K}^{1,L=0}\left({\bf x},{\bf x}^{\prime}\right)=\mathcal{K}^{0,L=0}\left({\bf x},{\bf x}^{\prime}\right)=\mathcal{K}^{in}\left({\bf x},{\bf x}^{\prime}\right)
\end{equation}

\begin{equation}
\mathcal{K}_{in}\left({\bf x},{\bf x}^{\prime}\right)=\frac{1}{N_{0}}\sum_{i=1}^{N_0}{\bf x}_{i}{\bf x}_{i}^{\prime}\label{eq:initialconditionKin}
\end{equation}

where $F\left(\left\langle z^{2}\right\rangle ,\left\langle z^{\prime2}\right\rangle ,\left\langle zz^{\prime}\right\rangle \right)$
is a nonlinear function of the variances of two Gaussian variables
$z$ and $z^{\prime}$ and their covariance, whose form depends on
the nonlinearity of the network \cite{cho2009kernel}. As we see in
Eqs.$\text{\ref{eq:K1},\ref{eq:K0}}$ these variances and covariances
depend on the kernel functions of the previous layer and on the replica-dependent prior statistics represented by $m_{\tau,\tau^{\prime}}^{1,0}$.

The MGF can be written as a function of the statistics of one of these
kernels, and their difference, which we will denote as $\Delta_{\tau,\tau^{\prime}}^{L}\left({\bf x},{\bf x^{\prime}}\right)=\frac{\lambda\beta}{2}\left(\mathcal{K}_{\tau,\tau^{\prime}}^{1,L}\left({\bf x},{\bf x^{\prime}}\right)-\mathcal{K}_{\tau,\tau^{\prime}}^{0,L}\left({\bf x},{\bf x^{\prime}}\right)\right)$.
It is useful to define a new kernel, the discrete neural dynamical
kernel $K_{\tau,\tau^{\prime}}^{d,L}=\lim_{n\rightarrow0}\frac{\lambda\beta}{2}\sum_{\alpha=1}^{n}m_{\tau,\tau^{\prime}}^{n\beta}K_{\tau,\tau^{\prime}}^{n\beta,L}$,
which controls the dynamics of the mean predictor. It has a simple
expression in terms of the kernel $\mathcal{K}_{\tau,\tau'}^{0,L}({\bf x},{\bf x}')$
and the kernel difference $\Delta_{\tau,\tau^{\prime}}^{L}$. 
\begin{equation}
\mathcal{K}_{\tau,\tau^{\prime}}^{d,L}\left({\bf x},{\bf x^{\prime}}\right)=\begin{cases}
0 & \tau\leq\tau^{\prime}\\
m_{\tau,\tau^{\prime}}^{1}\Delta_{\tau,\tau^{\prime}}^{L}\left({\bf x},{\bf x^{\prime}}\right)+\tilde{\lambda}^{\left|\tau-\tau^{\prime}\right|+1}\mathcal{K}_{\tau,\tau^{\prime}}^{0,L}\left({\bf x},{\bf x^{\prime}}\right) & \tau>\tau^{\prime}
\end{cases}\label{eq:discrete Kd}
\end{equation}

We integrate over the replicated hidden layers variables $\bf{W}_{\tau}^{\alpha}$,
which replaces the $\bf{W}_{\tau}^{\alpha}$ dependent kernels with the averaged
kernels. We thus get an MGF that depends only of the $v_{\tau}^{\alpha}$
variables

\begin{align}
\mathcal{M}\left[\ell\right] & =\lim_{n\rightarrow0}\prod_{\alpha=1}^{n}\prod_{\tau=1}^{\infty}\intop dv_{\tau}^{\alpha}\exp\left(-S\left[v\right]-Q\left[\ell,v\right]\right)
\end{align}
\begin{align}
S\left[v\right] & =\frac{\beta}{4}\sum_{\tau=1}^{\infty}\left(\frac{\beta}{2}\sum_{\alpha,\beta=1}^{n}\sum_{\tau^{\prime}=1}^{\infty}v_{\tau}^{\alpha\top}m_{\tau,\tau^{\prime}}^{0}K_{\tau,\tau^{\prime}}^{0}v_{\tau^{\prime}}^{\beta}+\frac{2}{\lambda}\sum_{\alpha=1}^{n}\sum_{\tau^{\prime}=1}^{t-1}v_{\tau}^{\alpha\top}K_{\tau,\tau^{\prime}}^{d}v_{\tau^{\prime}}^{n}\right.\\
 & \left.+\frac{1}{\lambda}\sum_{\alpha=1}^{n}v_{\tau}^{\alpha\top}K_{\tau,\tau}^{d}v_{\tau}^{\alpha}+\sum_{\alpha=1}^{n}v_{\tau}^{\alpha\top}\left(v_{\tau}^{\alpha}-2iY\right)\right)\nonumber 
\end{align}
\begin{align}
Q\left[\ell,v\right]= & \frac{i\beta}{2}\sum_{\beta=1}^{n}
\sum_{t,\tau^{\prime}=1}^{\infty}
\sum_{{\bf x}}\ell_{t,{\bf x}}m_{t,\tau^{\prime}}^{0}k_{t,\tau^{\prime}}^{0\top}\left(\mathbf{x}\right)v_{\tau^{\prime}}^{\beta}+\frac{i}{\lambda}\sum_{t,\tau^{\prime}=1}^{t}\sum_{{\bf x}}\ell_{t,{\bf x}}k_{t,\tau^{\prime}}^{d\top}\left(\mathbf{x}\right)v_{\tau^{\prime}}^{n}\label{eq:this_equation} \\
 & +\frac{i}{\lambda}\sum_{\beta=1}^{n}\sum_{t=1}^{\infty}\sum_{\tau^{\prime}=t+1}^{\infty}\sum_{{\bf x}}\ell_{t,{\bf x}}k_{t,\tau^{\prime}}^{d\top}\left(\mathbf{x}\right)v_{\tau^{\prime}}^{\beta}-\sum_{t=1}^{\infty}\sum_{{\bf x},{\bf x'}}\frac{1}{2}m_{t,t'}^{1}\ell_{t,{\bf x}}\ell_{t',{\bf x'}}\mathcal{K}_{t,t'}^{1}\left(\mathbf{x},\mathbf{x'}\right)\nonumber
\end{align}

\subsection{Integrate Out Replicated Variables $v_{\tau}^{\alpha}$\label{subsec:rid of replica}}

We define a new variable $u_{\tau}=\frac{\lambda\beta}{2}\sum_{\alpha=1}^{n}v_{\tau}^{\alpha}$,
and integrate out $v_{\tau}^{\alpha\neq n}$. We obtain a simpler
expression of the MGF which is no longer replica dependent (after taking the limit $n\rightarrow0$).
\begin{align}
\mathcal{M}\left[\ell\right] & =\prod_{\tau=1}^{\infty}\intop dv_{\tau}\intop du_{\tau}\exp\left(-S\left[v,u\right]-Q\left[\ell,v,u\right]\right)\label{eq:discrete MGF with no replica}
\end{align}
\begin{align}
S\left[v,u\right] & =\frac{1}{2\lambda^{2}}\sum_{\tau,\tau^{\prime}=1}^{\infty}u_{\tau}^{\top}\left(m_{\tau,\tau^{\prime}}^{0}K_{\tau,\tau^{\prime}}^{0}-\frac{2}{\beta}\delta_{\tau,\tau^{\prime}}\left(I+\frac{1}{\lambda}K_{\tau,\tau}^{d}\right)\right)u_{\tau^{\prime}}\\
 & +\frac{1}{\lambda}\sum_{\tau=1}^{\infty}\left(\frac{1}{\lambda}\sum_{\tau^{\prime}=1}^{\tau-1}K_{\tau,\tau^{\prime}}^{d}v_{\tau^{\prime}}+\left(I+\frac{1}{\lambda}K_{\tau,\tau}^{d}\right)v_{\tau}-iY\right)^{\top}u_{\tau}\nonumber 
\end{align}
\begin{align}
\hspace{-1.5cm}Q\left[\ell,v,u\right]= & \frac{i}{\lambda}\sum_{t=1}^{\infty}\sum_{{\bf x}}\ell_{t,{\bf x}}\left(\sum_{\tau^{\prime}=1}^{\infty}m_{t,\tau^{\prime}}^{0}k_{t,\tau^{\prime}}^{0\top}u_{\tau^{\prime}}+\sum_{\tau^{\prime}=1}^{t}k_{t,\tau^{\prime}}^{d\top}v_{\tau^{\prime}}+\frac{2}{\lambda\beta}\sum_{\tau^{\prime}=t+1}^{\infty}k_{t,\tau^{\prime}}^{d\top}u_{\tau^{\prime}}\right)\\
 & -\sum_{t,t'=1}^{\infty}\sum_{{\bf x},{\bf x}^{\prime}}\frac{1}{2}\ell_{t,{\bf x}}\ell_{t',{\bf x^{\prime}}}m_{t,t'}^{1}k_{t,t'}^{1}\left({\bf x},{\bf x}\right)\nonumber 
\end{align}

\subsection{Detailed Calculation of the Mean Predictor\label{subsec:Large--limit}}

To derive the mean predictor we take the derivative of the MGF w.r.t.
$\ell_{t,{\bf x}}$:

\begin{equation}
\left\langle f\left(t,{\bf x}\right)\right\rangle =\left.\frac{\partial\mathcal{M}\left[\ell\right]}{\partial\ell_{t,{\bf x}}}\right|_{\ell_{t,{\bf x}}=0}
\end{equation}
which yields 
\begin{equation}
\left\langle f\left(t,{\bf x}\right)\right\rangle =\frac{1}{\lambda}\sum_{t^{\prime}=1}^{t}k_{t,t^{\prime}}^{d,L\top}\left({\bf x}\right)\left\langle -iv_{t^{\prime}}\right\rangle 
\end{equation}
Furthermore, from the H.S. transformation in Eq.$\text{\ref{eq:hstransform}}$,
we can relate $\left\langle v_{\tau}\right\rangle $ to the mean predictor
on the training data $f_{\text{train}}\left(t\right)$

\begin{equation}
iv_{t} = f_{\text{train}}\left(t\right) -Y\label{eq:vmean}
\end{equation}
For all moments of $f_{\text{train}}\left(t\right)$.
On the other hand we can get the statistics of $iv_{t}$ from the
MGF in Eq.\ref{eq:discrete MGF with no replica}. 
\begin{equation}
\left\langle f_{\text{train}}({t})\right\rangle =\left(I\lambda+K_{t,t}^{d,L}\right)^{-1}\sum_{t^{\prime}=1}^{t-1}K_{t,t^{\prime}}^{d,L}\left(Y-\left\langle f_{\text{train}}({t^{\prime})}\right\rangle \right)\label{eq:ftrmean}
\end{equation}
\begin{equation}
\left\langle f\left(t,{\bf x}\right)\right\rangle =\frac{1}{\lambda}\sum_{t^{\prime}=1}^{t}k_{t,t^{\prime}}^{d,L\top}\left({\bf x}\right)\left(Y-\left\langle \left(f_{\text{train}}\right)_{t^{\prime}}\right\rangle \right)\label{eq:fmean}
\end{equation}

where $K^{d,L}_{t,t'}$ is a $P\times P$ dimensional
kernel matrix defined as $\mathcal{K}_{\mu\nu,t,t^{\prime}}^{d,L}=K_{t,t^{\prime}}^{d,L}\left({\bf x}^{\mu},{\bf x}^{\nu}\right)$.
Now we can compute $\left\langle f\left({\bf x},\Theta_{t}\right)\right\rangle $
iteratively by combining Eqs.$\text{\ref{eq:ftrmean},\ref{eq:fmean}}$.

\subsection{Large $\lambda$ Limit\label{subsec:large lambda}}

All the results so far hold for any $T$ and $\lambda$. Now, we consider
the limit where the Markov proximal learning algorithm is equivalent
to Langevin dynamics in order to get expressions that are relevant
to a continuous time gradient descent. We consider $\lambda\rightarrow\infty$ and $t_{discrete}\sim O\left(\lambda\right)$,
and thus define a new continues time $t=t_{discrete}/\lambda\sim O\left(1\right).$
In this limit, the parameters defined in Eq.$\text{\ref{eq:params}}$
becomes 
\begin{equation}
\tilde{\lambda}^{t_{discrete}}=e^{-T\sigma^{-2}t},\tilde{\sigma}^{2}=\sigma^{2},\gamma=\frac{\sigma_0^2}{\sigma^2}-1
\end{equation}

Taking the limit of large $\lambda$ limit of Eq.\ref{eq:discrete MGF with no replica}
is straightforward, and yields
\begin{equation}
\mathcal{M}\left[\ell\right]=\intop Dv\intop Du\exp\left(-S\left[v,u\right]-Q\left[\ell,v,u\right]\right)\label{eq:SI MGF}
\end{equation}

Where
\begin{align}
S\left[v,u\right] & =\frac{1}{2}\intop_{0}^{\infty}dt\intop_{0}^{\infty}dt^{\prime}m\left(t,t^{\prime}\right)u^{\top}\left(t\right)K^{L}\left(t,t^{\prime}\right)u\left(t^{\prime}\right)\label{eq:SI S}\\
 & +\intop_{0}^{\infty}dt\left(\intop_{0}^{t}dt^{\prime}K^{L}_{d}\left(t,t^{\prime}\right)v\left(t^{\prime}\right)+v\left(t\right)-iY\right)^{\top}u\left(t\right)\nonumber 
\end{align}

and the source term action is 
\begin{align}
Q\left[\ell,v,u\right]= & i\intop_{0}^{\infty}dt\intop_{0}^{t}dt^{\prime}\left(K^{L}_{d}\left(t,t^{\prime}\right)\right)^{\top}v\left(t^{\prime}\right)\ell\left(t\right)\label{eq:SI Q}\\
 & +i\intop_{0}^{\infty}dt\intop_{0}^{\infty}dt^{\prime}m\left(t,t^{\prime}\right)\left(k^{L}\left(t,t^{\prime}\right)\right)^{\top}u\left(t^{\prime}\right)\ell\left(t\right)\nonumber \\
 & -\frac{1}{2}\intop_{0}^{\infty}dt\intop_{0}^{\infty}dt^{\prime}m\left(t,t^{\prime}\right)k^{L}\left(t,t^{\prime},{\bf x},{\bf x}\right)\ell\left(t\right)\ell\left(t^{\prime}\right)\nonumber 
\end{align}

Where in the infinite width limit, we can identify $v(t)$ with $f_\text{traim}(t)$ by $ iv_{t} = f_{\text{train}}\left(t\right) -Y$, which holds for all moments of $f_{\text{train}}(t)$, and thus to write the MGF in terms of $f_{\text{train}}(t)$,  as was done in the main text Eqs.\ref{eq:MGF}, \ref{eq:S}, \ref{eq:Q}.

For convenience, in the continuous time limit, we denote the NDK with a lower index $d$. The NDK in Eq.$\text{\ref{eq:discrete Kd}}$
can be rewritten as 
\begin{align}
\mathcal{K}^{L}_{d}\left(t,t^{\prime},{\bf x},{\bf x}^{\prime}\right)=m\left(t,t^{\prime}\right)\Delta^{L}\left(t,t^{\prime},{\bf x},{\bf x}^{\prime}\right)+e^{-T\sigma^{-2}\left|t-t^{\prime}\right|}\mathcal{K}^{L}\left(t,t^{\prime},{\bf x},{\bf x}^{\prime}\right)
\end{align}
with 
\begin{align}
\Delta^{L}\left(t,t^{\prime},{\bf x},{\bf x}^{\prime}\right) & =\frac{\lambda}{2T}\left(\mathcal{K}^{L,1}\left(t,t^{\prime},{\bf x},{\bf x}^{\prime}\right)-\mathcal{K}^{L,0}\left(t,t^{\prime},{\bf x},{\bf x}^{\prime}\right)\right)\\
 & =\mathcal{K}^{d,L-1}\left(t,t^{\prime},{\bf x},{\bf x}^{\prime}\right)\dot{\mathcal{K}}^{L}\left(t,t^{\prime},{\bf x},{\bf x}^{\prime}\right)\nonumber 
\end{align}
\begin{equation}
m\left(t,t^{\prime}\right)=\sigma^{2}e^{-T\sigma^{-2}\left|t-t^{\prime}\right|}+\left(\sigma_{0}^{2}-\sigma^{2}\right)e^{-T\sigma^{-2}\left(t+t^{\prime}\right)}
\end{equation}

With the kernels defined in Sec.\ref{subsec:The-neural-dynamical}
in the main text. Here the quantity $m\left(t,t^{\prime}\right)$
is the continuous time limit of $m_{t,t^{\prime}}^{1}$. As defined
in Eq.$\text{\ref{eq:mtt}}$, it represents the covariance of the
prior
\begin{equation}
\left\langle \Theta_{t}^{i}\Theta_{t^{\prime}}^{j}\right\rangle _{S_{0}}=\delta_{ij}m\left(t,t^{\prime}\right),\left\langle \Theta_{t}^{i}\right\rangle _{S_{0}}=0
\end{equation}
.

The above calculation leads to the recursion relation of $\mathcal{K}^{L}_{d}\left(t,t^{\prime},{\bf x},{\bf x}^{\prime}\right)$
given in Eq.$\text{\ref{eq:recursive kd}}$ in the main text: 
\begin{align}
\mathcal{K}^{L}_{d}\left(t,t^{\prime},{\bf x},{\bf x}^{\prime}\right)= & m\left(t,t^{\prime}\right)\mathcal{K}_d^{L-1}\left(t,t^{\prime},{\bf x},{\bf x}^{\prime}\right)\dot{\mathcal{K}}^{L}\left(t,t^{\prime},{\bf x},{\bf x}^{\prime}\right)\\
 & +e^{-T\sigma^{-2}\left|t-t^{\prime}\right|}\mathcal{K}^{L}\left(t,t^{\prime},{\bf x},{\bf x}^{\prime}\right)\nonumber 
\end{align}

with initial condition

\begin{equation}
\mathcal{K}_d^{L=0}\left(t,t^{\prime},{\bf x},{\bf x}^{\prime}\right)=e^{-T\sigma^{-2}\left|t-t^{\prime}\right|}\mathcal{K}_{in}\left({\bf x},{\bf x}^{\prime}\right)\label{eq:initialcond1}
\end{equation}

Where $\mathcal{K}_{in}\left({\bf x},{\bf x}^{\prime}\right)$ was defined in
Eq.\ref{eq:initialconditionKin}. We refer to this continuous time $K^{L}_{d}\left(t,t^{\prime},{\bf x},{\bf x}^{\prime}\right)$
as the Neural Dynamical Kernel (NDK). Note that it follows directly
from Eq.$\text{\ref{eq:recursive kd}}$ that 
\begin{equation}
\mathcal{K}^{L}_{d}\left(0,0,{\bf x},{\bf x}^{\prime}\right)=\mathcal{K}_{NTK}^{L}\left({\bf x},{\bf x}^{\prime}\right).
\end{equation}

For the mean predictor we use the results from the previous section
Eqs.$\text{\ref{eq:vmean},\ref{eq:ftrmean},\ref{eq:fmean}}$, take
the large $\lambda$ limit and turn the sums into integrals, we obtain
\begin{equation}
\left\langle f_{\text{train}}\left(t\right)\right\rangle =\intop_{0}^{t}dt^{\prime}K^{L}_{d}\left(t,t^{\prime}\right)\left(Y-\left\langle f_{\text{train}}\left(t^{\prime}\right)\right\rangle \right)\label{eq:meanftrain app}
\end{equation}
\begin{equation}
\left\langle f\left(t,{\bf x}\right)\right\rangle =\intop_{0}^{t}dt^{\prime}\left(k^{L}_{d}\left(t,t^{\prime},{\bf x}\right)\right)^{\top}\left(Y-\left\langle f_{\text{train}}\left(t^{\prime}\right)\right\rangle \right)\label{eq:meanf app}
\end{equation}

as given in Eqs.$\text{\ref{eq:meanftrain}},\ref{eq:meanf}$ in the
main text.

\subsection{Low $T$ limit\label{sec:lowT}}

We aim to formally take the limit $T\rightarrow0$ of Eqs. \ref{eq:meanftrain app}, \ref{eq:meanf app}. In this limit, it is natural to rescale the times $t_{scaled}=(T\sigma^{-2})t$, and consider $t_{sclaed}\sim\mathcal{O}(1)$, which accounts for the diffusive learning phase, where $t\sim\mathcal{O}(1/T)$. For convenience, we will drop the "scaled" and consider this subsection purely in scaled time. We first look at the leading contribution of the gradient-driven phase described by the NTK.
\begin{equation}
\lim_{T\rightarrow0}\left\langle f_{train}\left(t\right)\right\rangle =\lim_{T\rightarrow0}\left(\left(I-\exp\left(-\frac{\sigma^{2}}{T}K_{NTK}^{L}t\right)\right)Y\right)=Y-T\sigma^{-2}\delta\left(t\right)\left(K_{NTK}^{L}\right)^{-1}Y
\end{equation}
 We expand $\left\langle f_{train}\left(t\right)\right\rangle $ around Y to leading correction in $T$
 \begin{equation}
\left\langle f_{train}\left(t\right)\right\rangle \approx Y-T\sigma^{-2}\left(\delta\left(t\right)K_{NTK}^{-1}Y+f_{1}\left(t\right)\right)
\end{equation}
Using  Eqs.\ref{eq:meanftrain app}, \ref{eq:meanf app}, we find the integral equation for $f_1(t)$
\begin{equation}
\intop_{0}^{t}dt^{\prime}K_{d}^{L}\left(t,t^{\prime}\right)f_{1}\left(t^\prime\right)=\left(I-K_{d}^{L}\left(t,0\right)K_{NTK}^{-1}\right)Y\label{eq:f1}
\end{equation}
And the equation for $\left\langle f\left(t,{\bf x}\right)\right\rangle$ in terms of $f_1(t)$
\begin{equation}
\left\langle f\left(t,{\bf x}\right)\right\rangle =k_{d}^{L}\left(t,0\right)^{\top}\left(K_{NTK}^{L}\right)^{-1}Y+\intop_{0}^{t}dt^{\prime}k_{d}^{L}\left(t,t^{\prime},{\bf x}\right)^{\top}f_{1}\left(t^{\prime}\right)\label{eq:scaled time meanf}
\end{equation}

At $t=0$, $\left\langle f\left({\bf x},0\right)\right\rangle =\left(k_{NTK}^{L}\right)^{\top}\left(K_{NTK}^{L}\right)^{-1}Y$, which is the NTK equilibrium, marks the transition to the diffusive learning phase. At long time, looking for a constant solution to Eq.\ref{eq:scaled time meanf}, and using the identity in Eq.\ref{eq:kd and nngp} we find the equilibrium
\begin{equation}
    \lim_{t\rightarrow\infty}\left\langle f\left(t,{\bf x}\right)\right\rangle =k_{GP}^{\top}\left(K_{GP}^{L}\right)^{-1}Y
\end{equation}
Which is the NNGP equilibrium when taking $T\rightarrow0$.

\section{Second Moment\label{sec: variance}}
Our formalism allows for the derivation of higher moments of the predictor. In particular, we are interested in the covariance $\left\langle \delta f\left(t,{\bf x}\right)\delta f\left(t^{\prime},{\bf x}^{\prime}\right)\right\rangle \equiv\left\langle f\left(t,{\bf x}\right)f\left(t^{\prime},{\bf x}^{\prime}\right)\right\rangle -\left\langle f\left(t,{\bf x}\right)\right\rangle \left\langle f\left(t^{\prime},{\bf x}^{\prime}\right)\right\rangle $. We focus on the continuous time $\lambda\rightarrow\infty$ limit described in Sec.\ref{subsec:large lambda}, which is equivalent to Langevin dynamics. In order to calculate the second moment, we need to invert one time-dependent operator, which we denote as $B(t,t')\in\mathbb{R}^{P\times P}$:
\begin{equation}
B\left(t,t^{\prime}\right)=I\delta\left(t-t^{\prime}\right)+K^{L}_{d}\left(t,t^{\prime}\right), 
\end{equation}
\begin{equation}
\intop_{0}^{t}d\tau B\left(t,\tau\right)B^{-1}\left(\tau,t^{\prime}\right)=I\delta\left(t-t^{\prime}\right)
\end{equation}

The full statistics of the Gaussian field $v(t), u(t)$ can be written in terms of $B^{-1}(t,t')$
\begin{equation}
\left\langle v\left(t\right)\right\rangle =i\int_{0}^{t}dt'B^{-1}\left(t,t^{\prime}\right)Y
\end{equation}
\begin{equation}
\left\langle \delta v\left(t\right)\delta v^{\top}\left(t^{\prime}\right)\right\rangle =-\int_{0}^{\infty}d\tau^{\prime}\int_{0}^{\infty}d\tau B^{-1}\left(t,\tau\right)m\left(\tau,\tau^{\prime}\right)K^{L}\left(\tau,\tau^{\prime}\right)B^{-1}\left(t^{\prime},\tau^{\prime}\right)
\end{equation}
\begin{equation}
\left\langle v\left(t\right)u^{\top}\left(t^{\prime}\right)\right\rangle =B^{-1}\left(t,t^{\prime}\right)
\end{equation}

It is useful to separate the smooth part from the delta function in the inverse operator $B^{-1}(t,t')$. We denote the smooth function as $J(t,t')\in\mathbb{R}^{P\times P}$, which satisfies the following integral equation:
\begin{equation}
J\left(t,t^{\prime}\right)=\begin{cases}
K_{d}^{L}\left(t,t^{\prime}\right)-\intop_{t^{\prime}}^{t}d\tau K_{d}^{L}\left(t,\tau\right)J\left(\tau,t^{\prime}\right) & t\geq t^{\prime}\\
0 & t<t^{\prime}
\end{cases}
\label{eq:Jeq}
\end{equation}
\begin{equation}
B^{-1}\left(t,t^{\prime}\right)=I\delta\left(t-t^{\prime}\right)-J\left(t,t^{\prime}\right)
\end{equation}

We take the second derivative of the MGF (Eq.\ref{eq:SI MGF}):
\begin{align}
    \left\langle \delta f\left({\bf x},t\right)\delta f\left({\bf x}^{\prime},t^{\prime}\right)\right\rangle =\left.\frac{\partial^{2}\mathcal{M}\left[\ell\right]}{\partial\ell\left(t,{\bf x}\right)\partial\ell\left(t^{\prime},{\bf x}^{\prime}\right)}\right|_{\ell\left(t,{\bf x}\right)=\ell\left(t^{\prime},{\bf x}^{\prime}\right)=0}-\left\langle f\left({\bf x},t\right)\right\rangle \left\langle f\left({\bf x}^{\prime},t^{\prime}\right)\right\rangle 
\end{align}
Which we can express in terms of $J(t,t')$ using the derived statistics of $v(t)$,$u(t)$

\begin{align}
& 
\left\langle \delta f_{\text{train}}\left(t\right)\delta f_{\text{train}}^{\top}\left(t^{\prime}\right)\right\rangle =m\left(t,t^{\prime}\right)K^{L}\left(t,t^{\prime}\right)-\intop_{0}^{t}d\tau\left[J\left(t,\tau\right)m\left(t^{\prime},\tau\right)K^{L}\left(t^{\prime},\tau\right)\right] \\ & -\intop_{0}^{t^{\prime}}d\tau\left[J\left(t^{\prime},\tau\right)m\left(t,\tau\right)K^{L}\left(t,\tau\right)\right]+\intop_{0}^{t}d\tau\intop_{0}^{t^{\prime}}d\tau^{\prime}\left[J\left(t,\tau\right)m\left(\tau,\tau^{\prime}\right)K^{L}\left(\tau,\tau^{\prime}\right)J\left(t^{\prime},\tau^{\prime}\right)\right]\nonumber
\end{align}
\begin{align}
& 
\left\langle \delta f\left(t,{\bf x}\right)\delta f\left(t^{\prime},{\bf x}^{\prime}\right)\right\rangle =\intop_{0}^{t}d\tau\intop_{0}^{t^{\prime}}d\tau^{\prime}[k_{d}^{L}\left(t,\tau,{\bf x}\right)^{\top}\left\langle \delta f_{train}\left(\tau\right)\delta f_{train}^{\top}\left(\tau^{\prime}\right)\right\rangle k_{d}^{L}\left(t^{\prime},\tau^{\prime},{\bf x}^{\prime}\right)] \label{eq:var f}
\\ & 
+\intop_{0}^{t}d\tau\intop_{0}^{\tau}d\tau^{\prime}[k_{d}^{L}\left(t,\tau,{\bf x}\right)^{\top}J\left(\tau,\tau^{\prime}\right)m\left(t^{\prime},\tau^{\prime}\right)k^{L}\left(t^{\prime},\tau^{\prime},{\bf x}^{\prime}\right)]+\intop_{0}^{t^{\prime}}d\tau\intop_{0}^{\tau}d\tau^{\prime}[m\left(t,\tau\right)k^{L}\left(t,\tau,{\bf x}\right)^{\top}J\left(\tau,\tau^{\prime}\right)k^{L}_{d}\left(t^{\prime},\tau^{\prime},{\bf x}^{\prime}\right)]\nonumber \\ & -\intop_{0}^{t}d\tau [k^{L}_{d}\left(t,\tau,{\bf x}\right)^{\top}m\left(t^{\prime},\tau\right)k^{L}\left(t^{\prime},\tau,{\bf x}^{\prime}\right)]-\intop_{0}^{t^{\prime}}d\tau [m\left(t,\tau\right)k^{L}\left(t,\tau,{\bf x}\right)^{\top}k^{L}_{d}\left(t^{\prime},\tau,{\bf x}^{\prime}\right)]+m\left(t,t^{\prime}\right)\mathcal{K}^{L}\left(t,t^{\prime},{\bf x},{\bf x'}\right)\nonumber
\end{align}

The equation becomes simpler for the correlation with initial condition, achieved by plugging $t'=0$ in Eq.\ref{eq:var f}
\begin{align}
& 
\left\langle \delta f\left(t,{\bf x}\right)\delta f\left(t^{\prime}=0,{\bf x}^{\prime}\right)\right\rangle =m\left(t,0\right)\mathcal{K}^{L}\left(t,0,{\bf x},{\bf x'}\right)-\intop_{0}^{t}d\tau [k^{L}_{d}\left(t,\tau,{\bf x}\right)^{\top}m\left(\tau,0\right)k^{L}\left(\tau,0,{\bf x}^{\prime}\right)]
\\ & +\intop_{0}^{t}d\tau\intop_{0}^{\tau}d\tau^{\prime}[k_{d}^{L}\left(t,\tau,{\bf x}\right)^{\top}J\left(\tau,\tau^{\prime}\right)m\left(\tau^{\prime},0\right)k^{L}\left(\tau^{\prime},0,{\bf x}^{\prime}\right)]\nonumber
\end{align}
We note that the mean predictor can also be written using the $J(t,t')$ operator:
\begin{equation}
\left\langle f_{\text{train}}\left(t\right)\right\rangle=\intop_{0}^{t}dt^{\prime}J\left(t,t^{\prime}\right)Y\label{eq:ftrain J}
\end{equation}

\begin{equation}
\left\langle f\left(t,{\bf x}\right)\right\rangle =\intop_{0}^{t}dt^{\prime}\left[k_{d}^{L}\left(t,t^{\prime},{\bf x}\right)^{\top}\left(I-\intop_{0}^{t^{\prime}}dt^{\prime\prime}J\left(t^{\prime},t^{\prime\prime}\right)\right)\right]Y\label{eq:f J}
\end{equation}

Solving the integral equation for $J(t,t')$ for a general nonlinearity is complex. 
However, the equations are tractable in two cases: Linear networks and the NTK limit ($T\rightarrow0,t\sim\mathcal{O}(1)$), which are presented below.

\subsection{The NTK Limit}
The time dependence of all kernels arises from $m(t,t')$, and thus at the NTK limit, defined by $T\rightarrow0,t\sim\mathcal{O}(1)$, we can substitute all the kernels and temporal correlations with their values at initialization, specifically $K_d^L(t,t')\approx K_{NTK}^L,K(t,t')=K_{GP_0},m(t,t')=\sigma_0^2$. Solving $J(t,t')$ with a constant NDK yields
\begin{equation}
    J\left(t,t^{\prime}\right)=\begin{cases}
K_{NTK}\exp\left(-K_{NTK}^{L}\left(t-t^{\prime}\right)\right) & t\geq t^{\prime}\\
0 & t<t^{\prime}
\end{cases}
\end{equation}
The only time dependence in the covariance equation (Eq.\ref{eq:var f}) comes from $J(t,t')$, as the kernels and $m(t,t')$ are constant. Performing the integral over the exponential $J(t,t')$ results in

\begin{align}
& \lim_{T\rightarrow0}\space\sigma_{0}^{-2}\left\langle \delta f\left(t,{\bf x}\right)\delta f\left(t^{\prime},{\bf x}^{\prime}\right)\right\rangle = \mathcal{K}_{GP_{0}}^{L}\left({\bf x},{\bf x^{\prime}}\right)-k_{GP_{0}}^{L}\left({\bf x}\right)(K_{GP_{0}}^{L})^{-1}k_{GP_{0}}^{L}\left({\bf x^{\prime}}\right)\\ & 
+\left[\left(I-\exp\left(-K_{NTK}^{L}t\right)\right)(K_{NTK}^{L})^{-1}k_{NTK}^{L}\left({\bf x}\right)-(K_{GP_{0}}^{L})^{-1}k_{GP_{0}}^{L}\left({\bf x}\right)\right]^{\top}K_{GP_{0}}^{L}\nonumber
\\ & \cdot\left[\left(I-\exp\left(-K_{NTK}^{L}t^{\prime}\right)\right)(K_{NTK}^{L})^{-1}k_{NTK}^{L}\left({\bf x}^{\prime}\right)-(K_{GP_{0}}^{L})^{-1}k_{GP_{0}}^{L}\left({\bf x^{\prime}}\right)\right]\nonumber\\
 & \nonumber
 \end{align}
Which is the result from Sec.\ref{subsec:Gradient-driven-phase-correspond}.

\subsection{Linear Network \label{sec:linear network}}
For a linear network, the NDK can be written in terms of the sum of exponents (see Sec.\ref{subsec:SI The-neural-dynamical}), and the integral equations for the first and second moments are tractable. We can represent both of them in terms of the function $J(t,t')$ (Eq.\ref{eq:Jeq})
\begin{align}
& J\left(t,t^{\prime}\right)=K_{d}^{L}\left(t^{\prime},t^{\prime}\right) & \\ &\exp\left(-\left(L+1\right)\left(\left(K_{GP}^{L}+IT\sigma^{-2}\right)\left(t-t^{\prime}\right)+\frac{1}{2T\sigma^{-2}}K_{GP}^{L}\sum_{n=1}^{L}\frac{L!}{n!\left(L-n\right)!}\frac{\gamma^{n}}{n}\left(e^{-2T\sigma^{-2}nt^{\prime}}-e^{-2T\sigma^{-2}nt}\right)\right)\right)\nonumber
\end{align}

Where $K_{GP}^L=\sigma^{2L}K_{in}$, $K_{in}$ is defined in Eq.\ref{eq:initialconditionKin} and $K_d^L(t,t')$ is given in linear network in Eq.\ref{eq:linear NDK}.

The mean predictor and the covariance can be calculated by substituting the expression for $J(t,t')$ into Eqs.\ref{eq:var f}, \ref{eq:f J}, leading to integrals that can be evaluated numerically, rather than integral equations like in the nonlinear case.

\textbf{Low $T$ Limit: }

We can further simplify the expressions by taking the limit of $T\rightarrow0$. In this limit, $J(t,t')$ is singular around $t=t'$ and is given by
\begin{equation}
J\left(t,t^{\prime}\right)=T\sigma^{-2}\left(I\delta\left(t-t^{\prime}\right)+T\sigma^{-2}\left(K_{d}^{L}\left(t,t\right)\right)^{-1}\left(\delta^{\prime}\left(t-t^{\prime}\right)+\left(L+1\right)\delta\left(t-t^{\prime}\right)\right)\right)
\end{equation}

Where $\delta(t-t')$ and $\delta^\prime(t-t')$ are the Dirac delta function and its derivative, respectively. The leading order in $T$ of the mean predictor is
\begin{equation}
f\left(t,{\bf x}\right)=k_{in}\left({\bf x}\right)^{\top}K_{in}^{-1}\left(I-\exp\left(-\left(L+1\right)\sigma_0^{2L}K_{in}t\right)\right)Y
\end{equation}
It is important to note that in a linear network, the NTK equilibrium identifies with the NNGP equilibrium, and thus, the mean predictor dynamics are identical to the NTK dynamics, and reaches equilibrium at $t\sim\mathcal{O}(1)$. 

The covariance equation in the low $T$ limit take the following simple form

\begin{equation}
    \left\langle \delta f\left(t,{\bf x}\right)\delta f\left(t^{\prime},{\bf x}^{\prime}\right)\right\rangle = m^{L+1}\left(t,t^{\prime}\right)\left[\mathcal{K}_{in}\left({\bf x},{\bf x}^{\prime}\right) - k_{in}\left({\bf x}\right)^{\top}\left(K_{in}\right)^{-1}k_{in}\left({\bf x}^{\prime}\right)\right] \nonumber
\end{equation}
The covariance can exhibit non-trivial dynamics at the diffusive phase, depending on the values of $\sigma,\sigma_0$, as shown in Fig.\ref{fig:linear var}.

\section{Multiple Outputs}\label{sec:multiple outputs}
The derivation from SI Sec.\ref{sec:Replica-calculation-of} can be repeated in the case of multiple outputs. We denote the number of outputs as $m$, and the predictor and target labels are m-dimensional vectors $f({\bf x},t)\in\mathbb{R}^m,y^\mu\in\mathbb{R}^m$, and the readout is a matrix ${\bf a}(t)\in\mathbb{R}^{N_{L}\times m}$. As long as the prior is an elementwise norm of the parameters $\Theta$, the MGF breaks down to $m$ uncoupled components, leading to the following mean field equations for the mean predictor:
\begin{equation}
\left\langle f_{\text{train}}\left(t\right)\right\rangle =\intop_{0}^{t}dt^{\prime}K^{L}_{d}\left(t,t^{\prime}\right)\left(Y-\left\langle f_{\text{train}}\left(t^{\prime}\right)\right\rangle \right)
\end{equation}

Where $ f_{\text{train}}\left(t\right),Y\in\mathbb{R}^{P\times m}$ are matrices, and $K_d^L(t,t')\in\mathbb{R}^{P\times P}$ is the NDK as defined in previous parts (Sec. \ref{subsec:The-neural-dynamical}). The equation for the predictor on a test point is given by
\begin{equation}
\left\langle f\left(t,{\bf x}\right)\right\rangle =\intop_{0}^{t}dt^{\prime}\left(Y-\left\langle f_{\text{train}}\left(t^{\prime}\right)\right\rangle \right)^{\top}k_{d}^{L}\left(t,t^{\prime},{\bf x}\right)
\end{equation}
Where $k_d^L(t,t')\in\mathbb{R}^{P}$ is the NDK test vector as defined in previous parts (Sec. \ref{subsec:The-neural-dynamical}). The result is an m-dimensional vector $f(t,{\bf x})\in\mathbb{R}^m$ as required, and $f_{\text{train}}^\mu\left(t\right)=f^\top(t,{\bf x}_\mu)$.

Due to the uncoupling of the MGF, the covariance is diagonal in the outputs

\begin{equation}
    \left\langle \delta f_{m}\left(t,{\bf x}\right)\delta f_{m^{\prime}}\left(t^{\prime},{\bf x}^{\prime}\right)\right\rangle \propto\delta_{m,m^{\prime}}
\end{equation}
Where for each output the covariance satisfies the same equations as described in SI Sec.\ref{sec: variance}, and the covariance between different outputs is zero.

\section{The Neural Dynamical Kernel\label{subsec:SI The-neural-dynamical}}

We focus on the continuous time limit derived above, and present
several examples where the NDK has explicit expressions, and provide
proofs of properties of the NDK presented in the main text. We have derived
\begin{align}
\mathcal{K}^{L}_{d}\left(t,t^{\prime},{\bf x},{\bf x}^{\prime}\right)= & m\left(t,t^{\prime}\right)\mathcal{K}_d^{L-1}\left(t,t^{\prime},{\bf x},{\bf x}^{\prime}\right)\dot{\mathcal{K}}^{L}\left(t,t^{\prime},{\bf x},{\bf x}^{\prime}\right)\\
 & +e^{-T\sigma^{-2}\left|t-t^{\prime}\right|}\mathcal{K}^{L}\left(t,t^{\prime},{\bf x},{\bf x}^{\prime}\right)\nonumber 
\end{align}
In order to complete the calculation of the NDK, we would provide explicit analytical expressions for $\mathcal{K}(t,t',{\bf x}, {\bf x'})$ and $\dot{\mathcal{K}}(t,t',{\bf x}, {\bf x'})$ in cases where they are available, namely linear activation, and ReLU and error function nonlinearities.

\subsection{Linear Activation:}

For linear activation: 
\begin{equation}
\mathcal{K}^{L}\left(t,t^{\prime},{\bf x},{\bf x}^{\prime}\right)=\left(m\left(t,t^{\prime}\right)\right)^{L}\mathcal{K}_{in}\left({\bf x},{\bf x}^{\prime}\right)
\end{equation}
\begin{equation}
\dot{\mathcal{K}}^{L}\left(t,t^{\prime},{\bf x},{\bf x}^{\prime}\right)=I
\end{equation}

The recursion relation for the NDK can be solved explicitly, yielding
\begin{equation}
\mathcal{K}^{L}_{d}\left(t,t^{\prime},{\bf x},{\bf x}^{\prime}\right)=\left(m\left(t,t^{\prime}\right)\right)^{L}\left(L+1\right)e^{-T\sigma^{-2}\left|t-t^{\prime}\right|}\mathcal{K}_{in}\left({\bf x},{\bf x}^{\prime}\right)\label{eq:linear NDK}
\end{equation}

The NDK of linear activation is proportional to the input kernel $\mathcal{K}_{in}\left({\bf x},{\bf x}^{\prime}\right)$
regardless of the data. The effect of network depth only changes the
magnitude but not the shape of the NDK. As a result, the NNGP and
NTK kernels also only differ by their magnitude, and thus the mean
predictor at the NNGP and NTK equilibria only differ by $\mathcal{O}\left(T\right)$.
This suggests that the diffusive phase has very little effect on the
mean predictor in the low $T$ regime, in linear network, as discussed in Sec.\ref{sec:linear network}.

\subsection{ReLU Activation:}

For ReLU activation, we define the function $J\left(\theta\right)$
\cite{cho2009kernel}:

\begin{equation}
J\left(\theta^{L}\left(t,t^{\prime},{\bf x},{\bf x}^{\prime}\right)\right)=\left(\pi-\theta^{L}\left(t,t^{\prime},{\bf x},{\bf x}^{\prime}\right)\right)\cos\left(\theta^{L}\left(t,t^{\prime},{\bf x},{\bf x}^{\prime}\right)\right)+\sin\left(\theta^{L}\left(t,t^{\prime},{\bf x},{\bf x}^{\prime}\right)\right)
\end{equation}

where the angle between ${\bf x}$ and ${\bf x}^{\prime}$ is given
by :

\begin{equation}
\theta^{L}\left(t,t^{\prime},{\bf x},{\bf x}^{\prime}\right)=\cos^{-1}\left(\frac{m\left(t,t^{\prime}\right)}{\sqrt{m\left(t,t\right)m\left(t^{\prime},t^{\prime}\right)}}\frac{1}{\pi}J\left(\theta^{L-1}\left(t,t^{\prime},{\bf x},{\bf x}^{\prime}\right)\right)\right)
\end{equation}

$\theta^{L}\left(t,t^{\prime},{\bf x},{\bf x}^{\prime}\right)$ is
defined through a recursion equation, and 
\begin{equation}
\theta^{L=0}\left(t,t^{\prime},{\bf x},{\bf x}^{\prime}\right)=\cos^{-1}\left(\frac{m\left(t,t^{\prime}\right)}{\sqrt{m\left(t,t\right)m\left(t^{\prime},t^{\prime}\right)}}\frac{\mathcal{K}_{in}\left({\bf x},{\bf x}^{\prime}\right)}{\sqrt{\mathcal{K}_{in}({\bf x},{\bf x})\mathcal{K}_{in}({\bf x}^{\prime},{\bf x}^{\prime})}}\right)
\end{equation}
the kernel functions are then given by

\begin{equation}
\dot{\mathcal{K}}^{L}\left(t,t^{\prime},{\bf x},{\bf x}^{\prime}\right)=\frac{1}{2\pi}\left(\pi-\theta^{L}\left(t,t^{\prime},{\bf x},{\bf x}^{\prime}\right)\right)
\end{equation}
\begin{equation}
\mathcal{K}^{L}\left(t,t^{\prime},{\bf x},{\bf x}^{\prime}\right)=\frac{\sqrt{\mathcal{K}_{in}\left({\bf x},{\bf x}\right)\mathcal{K}_{in}\left({\bf x}^{\prime},{\bf x}^{\prime}\right)}}{\pi2^{L}}\left(m\left(t,t\right)m\left(t^{\prime},t^{\prime}\right)\right)^{L/2}J\left(\theta^{L-1}\left(t,t^{\prime},{\bf x},{\bf x}^{\prime}\right)\right)
\end{equation}
We obtain an explicit expression for the NDK by plugging these kernels
into Eqs.$\text{\ref{eq:recursive kd},\ref{eq:initialcond1}}.$

\subsection{Error Function Activation}

For error function activation \cite{williams1996computing}: 
\begin{equation}
\mathcal{K}^{L}\left(t,t^{\prime},{\bf x},{\bf x}^{\prime}\right)=\frac{2}{\pi}\sin^{-1}\left(\frac{2m\left(t,t^{\prime}\right)\mathcal{K}^{L-1}\left(t,t^{\prime},{\bf x},{\bf x}^{\prime}\right)}{\sqrt{\left(1+2m\left(t,t\right)\mathcal{K}^{L-1}\left(t,t,{\bf x},{\bf x}\right)\right)\left(1+2m\left(t^{\prime},t^{\prime}\right)\mathcal{K}^{L-1}\left(t^{\prime},t^{\prime},{\bf x}^{\prime},{\bf x}^{\prime}\right)\right)}}\right)
\end{equation}
\begin{align}
\hspace{-1.5cm}\dot{\mathcal{K}}_{\mu\nu}^{L}\left(t,t^{\prime},{\bf x},{\bf x}^{\prime}\right)= & \frac{4}{\pi}\left(\left(1+2m\left(t,t\right)\mathcal{K}^{L-1}\left(t,t,{\bf x},{\bf x}\right)\right)\left(1+2m\left(t^{\prime},t^{\prime}\right)\mathcal{K}^{L-1}\left(t^{\prime},t^{\prime},{\bf x}^{\prime},{\bf x}^{\prime}\right)\right)\right.\nonumber \\
 & \left.-4\left(m\left(t,t^{\prime}\right)\mathcal{K}^{L-1}\left(t,t^{\prime},{\bf x},{\bf x}^{\prime}\right)\right)^{2}\right)^{-1/2}
\end{align}

Again we can obtain an explicit expression for the NDK by plugging
these kernels into Eqs.$\text{\ref{eq:recursive kd},\ref{eq:initialcond1}}.$

\subsection{Long Time Behavior of the NDK\label{subsec:long-time-ndk}}

We define the long time limit as $t,t^{\prime}\rightarrow\infty,t-t^{\prime}\sim\mathcal{O}\left(T^{-1}\right)$.
At a long time the statistics of $\bf{W}$ w.r.t. the prior becomes
only a function of the time difference: 
\begin{equation}
\left\langle {\bf{W}}_{t}{\bf{W}}_{t^{\prime}}^{\top}\right\rangle =\sigma^{2}e^{-T\sigma^{-2}\left|t-t^{\prime}\right|}=m\left(\left|t-t^{\prime}\right|\right)
\end{equation}

And thus, the kernels defined above also will be only functions of
the time difference. We look at the time derivative of the kernel
(w.l.o.g. we assume $t>t^{\prime}$), which can be obtained with a
chain rule: 
\begin{equation}
\frac{d}{dt^{\prime}}\mathcal{K}^{L}\left(t-t^{\prime},{\bf x},{\bf x}^{\prime}\right)=\dot{\mathcal{K}}^{L}\left(t-t^{\prime},{\bf x},{\bf x}^{\prime}\right)\frac{d}{dt^{\prime}}\left(\mathcal{K}^{L-1}\left(t-t^{\prime},{\bf x},{\bf x}^{\prime}\right)m\left(t-t^{\prime}\right)\right)
\end{equation}

We prove by induction: 
\begin{equation}
\frac{1}{T}\frac{d}{dt^{\prime}}\left(m\left(t-t^{\prime}\right)\mathcal{K}^{L}\left(t-t^{\prime},{\bf x},{\bf x}^{\prime}\right)\right)=\mathcal{K}^{L}_{d}\left(t-t^{\prime},{\bf x},{\bf x}^{\prime}\right)
\end{equation}

The induction basis for $L=0$ is trivial. For arbitrary $L+1$: 
\begin{align}
\hspace{-1.5cm}\frac{1}{T}\frac{d}{dt^{\prime}}\left(m\left(t-t^{\prime}\right)\mathcal{K}^{L+1}\left(t-t^{\prime},{\bf x},{\bf x}^{\prime}\right)\right) & =m\left(t-t^{\prime}\right)\dot{\mathcal{K}}^{L+1}\left(t-t^{\prime},{\bf x},{\bf x}^{\prime}\right)\frac{1}{T}\frac{d}{dt^{\prime}}\left(\mathcal{K}^{L}\left(t-t^{\prime},{\bf x},{\bf x}^{\prime}\right)m\left(t-t^{\prime}\right)\right)\nonumber \\
 & +e^{-T\sigma^{-2}\left(t-t^{\prime}\right)}\mathcal{K}^{L+1}\left(t-t^{\prime},{\bf x},{\bf x}^{\prime}\right)
\end{align}

And using the induction assumption we get: 
\begin{align}
\hspace{-1.5cm}\frac{1}{T}\frac{d}{dt^{\prime}}\left(m\left(t-t^{\prime}\right)\mathcal{K}^{L+1}\left(t-t^{\prime},{\bf x},{\bf x}^{\prime}\right)\right) & =m\left(t-t^{\prime}\right)\dot{\mathcal{K}}^{L+1}\left(t-t^{\prime},{\bf x},{\bf x}^{\prime}\right)\mathcal{K}^{L}_{d}\left(t-t^{\prime},{\bf x},{\bf x}^{\prime}\right)\nonumber \\
 & +e^{-T\sigma^{-2}\left(t-t^{\prime}\right)}\mathcal{K}^{L+1}\left(t-t^{\prime},{\bf x},{\bf x}^{\prime}\right)
\end{align}
Which is the expression for $\mathcal{K}^{d,L+1}\left(t-t^{\prime}\right)$.
Using this identity, we can get a simple expression for the integral
over $\mathcal{K}^{L}_{d}\left(t-t^{\prime}\right)$ at long times: 
\begin{equation}
\lim_{t\rightarrow\infty}\left(\frac{T}{\sigma^{2}}\intop_{0}^{t}dt^{\prime}\mathcal{K}_{d}^{L}\left(t-t^{\prime},{\bf x},{\bf x}^{\prime}\right)\right)=\mathcal{K}_{GP}^{L}\left({\bf x},{\bf x}^{\prime}\right)
\end{equation}

\subsection{NDK as a Generalized Time-Dependent NTK\label{subsec:ndk-interp}}

In Eq.$\text{\ref{eq:interp}}$ in the main text, we claimed that
the NDK has the following interpretation as a generalized two-time
NTK

\begin{equation}
\mathcal{K}^{L}_{d}\left(t,t^{\prime},{\bf x},{\bf x}^{\prime}\right)=e^{-T\sigma^{-2}\left|t-t^{\prime}\right|}\left\langle \nabla_{\Theta_{t}}f\left({\bf x},\Theta_{t}\right)\cdot\nabla_{\Theta_{t^{\prime}}}f\left({\bf x^{\prime}},\Theta_{t^{\prime}}\right)\right\rangle _{0}t\geq t^{\prime}\label{eq:interp-1-1}
\end{equation}

where $\langle\cdot\rangle_{0}$ denotes averaging w.r.t. the
prior distribution of the parameters $\Theta$, with the statistics
defined in Eq.\ref{eq:mtt'}.

Now we provide a formal proof.

We separate $\nabla_{\Theta_{t}}f\left({\bf x},\Theta_{t}\right)$
into two parts including the derivative w.r.t. the readout weights
$a_{t}$ and the hidden layer weights $\bf{W}_{t}$

\textbf{Derivative w.r.t. the readout weights:} 
\begin{equation}
\left\langle \partial_{{\bf a}_{t}}f\left({\bf x},\Theta_{t}\right)\cdot\partial_{{\bf a}_{t^{\prime}}}f\left({\bf x},\Theta_{t^{\prime}}\right)\right\rangle _{0}=\mathcal{K}^{L}\left(t,t^{\prime},{\bf x},{\bf x}^{\prime}\right)\label{eq:derivativewa}
\end{equation}

\textbf{Derivative w.r.t. the hidden layer weights:}

We have 
\begin{equation}
\partial_{{\bf W}_{t}^{l}}{\bf x}_{t}^{L}\left({\bf x},{\bf{W}}_{t}\right)=\frac{1}{\sqrt{N_{L-1}\cdots N_{l-1}}}\Pi_{k=l+1}^{L}\left[\phi^{\prime}\left(z_{t}^{k}\right){\bf W}_{t}^{k}\right]\phi^{\prime}\left(z_{t}^{l}\right){\bf x}_{t}^{l-1}
\end{equation}
and 
\begin{align}
 & \left\langle \partial_{{\bf {\bf W}}_{t}^{l}}f\left({\bf x},\Theta_{t}\right)\cdot\partial_{{\bf W}_{t^{\prime}}^{l}}f\left({\bf x},\Theta_{t^{\prime}}\right)\right\rangle _{0}\nonumber \\
 & =\left\langle N_{L}^{-1}{\bf a}_{t}\cdot{\bf a}_{t^{\prime}}\right\rangle \left(\Pi_{k=l+1}^{L}\left\langle N_{k}^{-1}N_{k-1}^{-1}{\bf W}_{t}^{k}\cdot{\bf W}_{t^{\prime}}^{k}\right\rangle \right)\left(\Pi_{k=l}^{L}\dot{\mathcal{K}}^{k}\left(t,t^{\prime},{\bf x},{\bf x}^{\prime}\right)\right)\mathcal{K}^{l-1}\left(t,t^{\prime},{\bf x},{\bf x}^{\prime}\right)\nonumber \\
 & =m\left(t,t^{\prime}\right)^{L-l+1}\left(\Pi_{k=l}^{L}\dot{\mathcal{K}}^{k}\left(t,t^{\prime},{\bf x},{\bf x}^{\prime}\right)\right)\mathcal{K}^{l-1}\left(t,t^{\prime},{\bf x},{\bf x}^{\prime}\right)
\end{align}
To leading order in $N_{l}$ the averages over ${\bf a}$ and $\bf{W}$
can be performed separately for each layer, and are dominated by their
prior, where each element of the weights is an independent Gaussian
given by Eq.$\text{\ref{eq:gaussianprior}}$. The term $m\left(t,t^{\prime}\right)$
comes from the covariance of the priors in $\bf{W}$ and ${\bf a}$,
since there are a total of $L-l$ layers of $\bf{W}$ and one
layer of ${\bf a}$, we have $m\left(t,t^{\prime}\right)^{L-l+1}$.
The kernel $\dot{\mathcal{K}}^{k}\left(t,t^{\prime},{\bf x},{\bf x}^{\prime}\right)$
comes from the inner product between $\phi^{\prime}\left(z_{t}^{k}\right)$
and $\phi^{\prime}\left(z_{t^{\prime}}^{k}\right)$, and the kernel
$\mathcal{K}^{l-1}\left(t,t^{\prime},{\bf x},{\bf x}^{\prime}\right)$ comes
from the inner product between ${\bf x}_{t}^{l-1}$ and ${\bf x}_{t^{\prime}}^{l-1}$.

Using proof by induction as for the NTK \cite{jacot2018neural}, we
obtain 
\begin{equation}
\left\langle \partial_{{\bf{W}}_{t}}f\left({\bf x},\Theta_{t}\right)\cdot\partial_{{\bf{W}}_{t^{\prime}}}f\left({\bf x},\Theta_{t^{\prime}}\right)\right\rangle _{0}=e^{T\sigma^{-2}\left|t-t^{\prime}\right|}m\left(t,t^{\prime}\right)\dot{\mathcal{K}}^{L}\left(t,t^{\prime},{\bf x},{\bf x}^{\prime}\right)\mathcal{K}^{d,L-1}\left(t,t^{\prime},{\bf x},{\bf x}^{\prime}\right)\label{eq:derivativeww}
\end{equation}

Combine Eq.$\text{\ref{eq:derivativeww}}$ with Eq.$\text{\ref{eq:derivativewa}}$
and with the definition of $\mathcal{K}^{L}_{d}\left(t,t^{\prime},{\bf x},{\bf x}^{\prime}\right)$
in Eq.$\text{\ref{eq:recursive kd}}$, we have 
\begin{equation}
e^{-T\sigma^{-2}\left|t-t^{\prime}\right|}\left\langle \nabla_{\Theta_{t}}f\left({\bf x},\Theta_{t}\right)\cdot\nabla_{\Theta_{t^{\prime}}}f\left({\bf x^{\prime}},\Theta_{t^{\prime}}\right)\right\rangle _{0}=\mathcal{K}^{L}_{d}\left(t,t^{\prime},{\bf x},{\bf x}^{\prime}\right)
\end{equation}

\section{Representational drift\label{subsec:details-rep-drift}}

To capture the phenomenon of representational drift, we consider the
case where the learning signal stops at some time $t_{0}$, while
the hidden layers continue to drift according to the dynamics of the
prior. If all the weights of the system are allowed to drift, the
performance of the mean predictor will deteriorate to chance, thus
we consider stable readout weights fixed at the end time of learning
$t_{0}$. This scenario can be theoretically evaluated using similar
techniques to Sec.\ref{sec:Replica-calculation-of} , leading to
the following equation for the network output: 
\begin{equation}
\left\langle f_{\text{drift}}\left({\bf x},t,t_{0}\right)\right\rangle =\intop_{0}^{t_{0}}\left(k^{L}_{d}\left({\bf x},t,t^{\prime}\right)\right)^{\top}\left(Y-\left\langle f_{\text{train}}\left(t^{\prime}\right)\right\rangle \right)
\end{equation}

We see here that if $t_{0}=t$ it naturally recovers the full mean
predictor. It is interesting to look at the limit where the freeze
time $t_{0}$ is at NNGP equilibrium. In this case, the expression can be simplified
due to the long time identity of the NDK (Eq.\ref{eq:kd and nngp} in the main text).
\begin{equation}
\left\langle f_{\text{drift}}\left({\bf x},t,t_{0}\right)\right\rangle =\left(k^{L}\left({\bf x},t,t_{0}\right)\right)^{\top}\left(IT\sigma^{-2}+K_{GP}^{L}\right)^{-1}Y
\end{equation}

which has a simple meaning of two samples of hidden layer weights
from different times at equilibrium. Even at long time differences,
the network performance does not decrease to chance, but reaches a
new static state.

\begin{equation}
\lim_{t-t_{0}\rightarrow\infty}\left\langle f_{\text{drift}}\left({\bf x},t,t_{0}\right)\right\rangle =\left(k_{mean}^{L}\left({\bf x}\right)\right)^{\top}\left(IT\sigma^{-2}+K_{GP}^{L}\right)^{-1}Y
\end{equation}

Where the mean kernel is defined in Eq.\ref{eq:K mean}. We can evaluate it with the usual kernel functions described in Sec.\ref{subsec:SI The-neural-dynamical}, with $m(t,t_0)=0,m(t,t)=m(t_0,t_0)=\sigma^2$.

We can assess the network's ability to separate classes in a binary
classification task by using a threshold between the two distributions
of outputs, as described in Sec.\ref{sec:Details-of-the}.

\subsection{Limited Receptive Field}\label{subsec:limited receptive field}
We consider the case where each neuron receives inputs only from a limited patch of the image. The kernel function in this scenario can be calculated by summing up the contribution from each patch:
\begin{equation}
    \tilde{\mathcal{K}}\left({\bf x},{\bf x}^{\prime}\right)=\sum_{b=1}^{B}\mathcal{K}\left({\bf x}_{b},{\bf x}_{b}^{\prime}\right)
\end{equation}
Where the input is split into $B$ patches, and $\tilde{\mathcal{K}}$ is the appropriate kernel in this scenario. The predictor can be calculated using the same methods described above with the modified kernel function. 
\section{NTK and NNGP equilibria}\label{Sec:NTK and NNGP equilibria}
\begin{figure}[h]
    \includegraphics[width=0.8\textwidth]{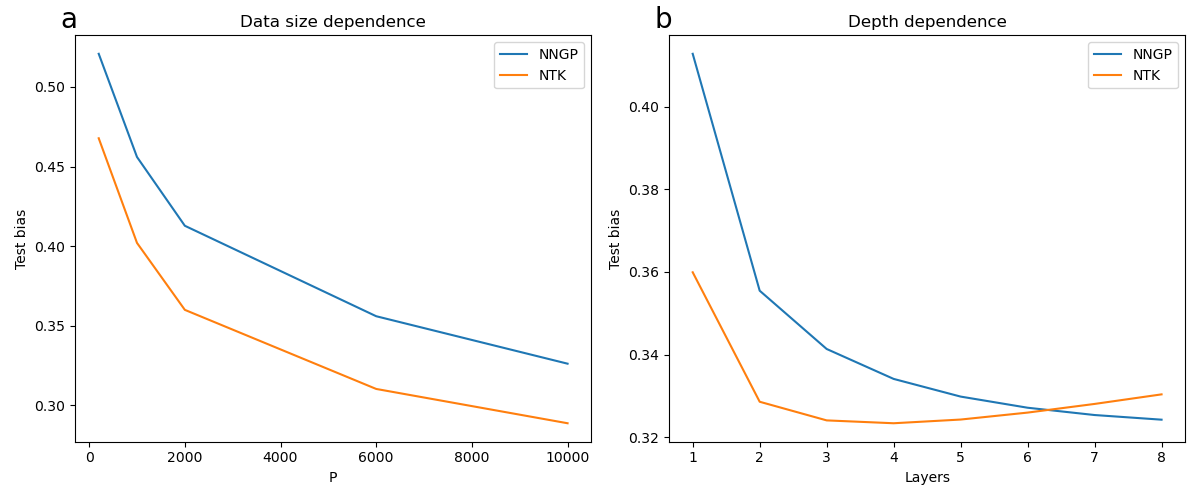}
    \caption{NTK vs. NNGP Comparison: The test bias, $\left(\langle f \rangle - Y\right)^{2}$, averaged over the test dataset, as a function of the number of training data points $P$ and network depth. (a) ReLU network with a single hidden layer, shown for different values of $P$. The relations between NTK and NNGP are approximately preserved as the training dataset size varies. (b) ReLU network with $P = 2000$ training data points, shown across different depths. Deeper networks tend to favor the NNGP equilibrium over the NTK.
}\label{Fig:NTK-NNGP P-L Depndence}
\end{figure}

The NTK equilibrium marks the starting state of the diffusion learning phase, whereas the NNGP equilibrium represents the final state of the learning process. Characterizing the diffusion learning phase requires analyzing how the differences between these equilibria depend on various parameters in real-world datasets. We use CIFAR-10 dataset and focus on their dependence on dataset size (\(P\)) and network depth. In a ReLU network in the infinite-width regime, these two parameters are the only ones affecting these equilibria, making it a convenient setup for studying this dependence. 
As shown in Fig.~\ref{Fig:NTK-NNGP P-L Depndence}(a), increasing \(P\) improves both equilibria in a comparable manner, preserving their relative relationship. Fig.~\ref{Fig:NTK-NNGP P-L Depndence}(b) demonstrates that increasing network depth improves the NNGP equilibrium to a greater extent than the NTK equilibrium. 

\begin{figure}
    \includegraphics[width=0.6\textwidth]{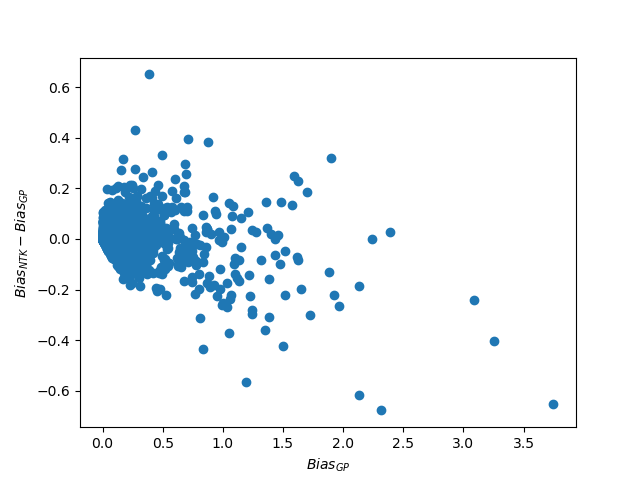}
    \caption{NTK vs. NNGP of Individual Points: The test bias, $\left(\langle f \rangle - Y\right)^{2}$, in a ReLU network with six hidden layers and $P=2000$ training data points. The difference between the NTK and NNGP bias is shown as a function of the NNGP bias. While some points exhibit large differences, the distribution is roughly symmetric, leading to a smaller overall difference when averaged.}\label{Fig:NTK-NNGP scatter}
\end{figure}

Although the average difference is small in some cases, individual points with large differences may still exist. Figure \ref{Fig:NTK-NNGP scatter} illustrates that a small difference may arise because instances where the NTK outperforms the NNGP are balanced by cases where the NNGP outperforms the NTK.

\newpage
\section{Theory and Simulation}
\begin{figure*}[h!]
\centering{}\includegraphics[width=0.5\textwidth]{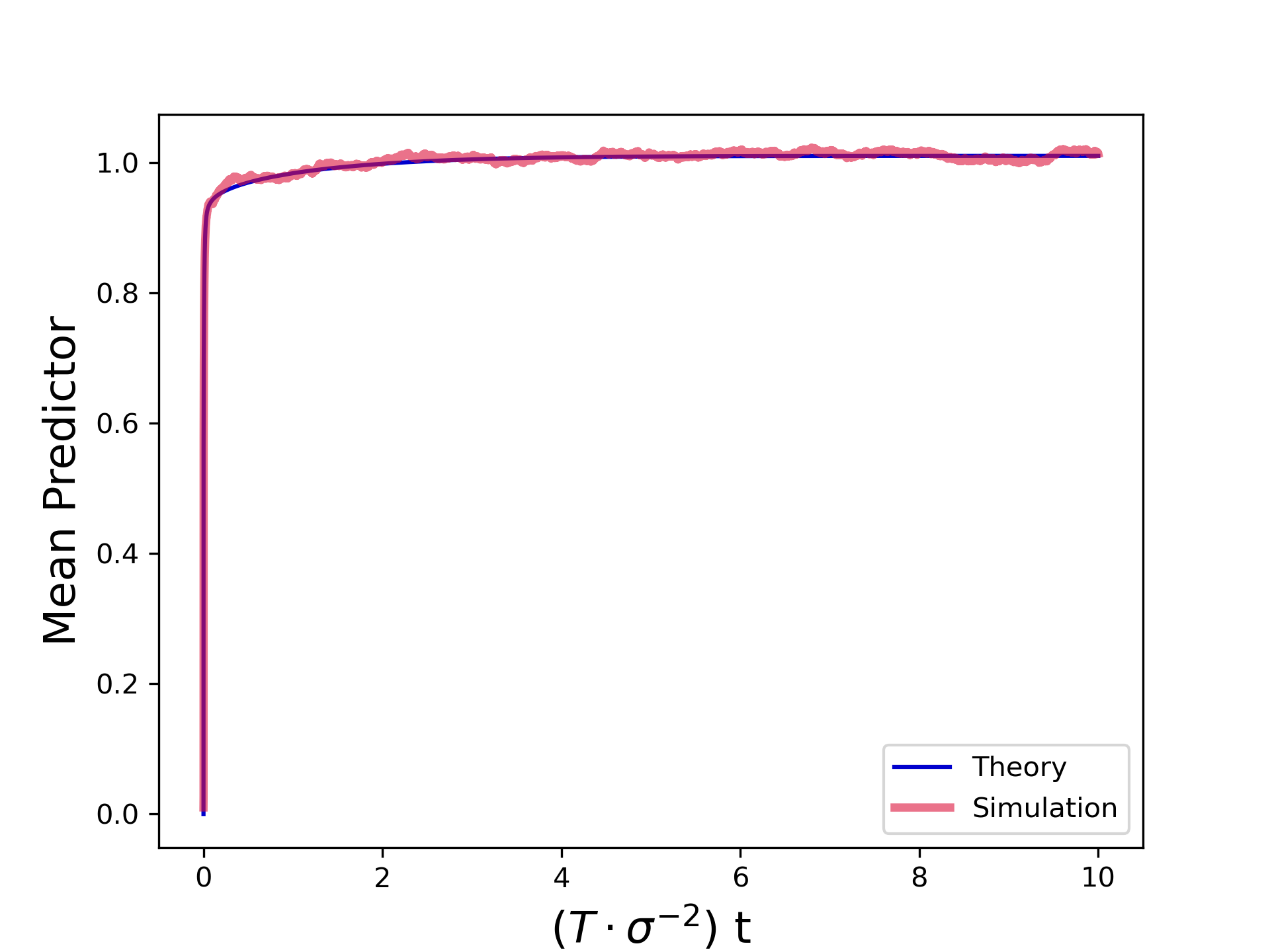}\caption{We compare the predictor calculated using our theory to an ensemble of finite width neural network trained with Langevin learning algorithm (Eq.\ref{eq:Langevin}) on MNIST binary classification, with the digits 0, 1. We average over 5000 networks to get the mean predictor of the ensemble, and compared it to the theory. The parameters are $T=0.01, \sigma=1,\sigma_0=1.44,dt=lr=0.1$. The network was trained on $P=10$ data points, and its hidden size was $N=1000$. The example presented is the test point 2591 from MNIST test dataset. Theory and simulation show remarkable agreement.}
\end{figure*}



\section{Details of the numerical simulations\label{sec:Details-of-the}}
\textbf{Figure 1: } We trained 100 deep networks with one hidden layer and an error function nonlinearity using Langevin dynamics (Eq. \ref{eq:Langevin}) for binary classification between the 'airplane' and 'frog' categories in the CIFAR-10 dataset \cite{krizhevsky2014cifar}, with $P = 400$ data points. The hidden layer size was $N = 10^4$ neurons, the Langevin dynamics temperature was $T = 10^{-4}$, the learning rate $lr=0.1$, the initialization variance $\sigma_0=0.2$, and the prior variance $\sigma=1.0$. The networks were trained for $3 \cdot 10^5$ epochs. The test loss in (a) was computed using MSE over 2000 test examples, and the average across all networks is also presented. The training loss in (b) was calculated using MSE over the training set, with the average across all networks also shown. In (c), the dynamics of 100 randomly selected weights from the input to the hidden layer are shown for one network, along with the standard deviation of all the hidden layer weights.

\textbf{Figure 2: } The NDK was calculated for MNIST binary classification with the digits 0, 1, for $P=100$ data points, and one hidden layer. The NDK was calculated for equal time $K_d^L(t,t)$ (Eq.\ref{eq:equal time}, a-c,g-i), and time difference from initialization $K_d^L(0,t)$ (Eq.\ref{eq:interp},d-f,j-l), for ReLU nonlinearity (a-f) and error function nonlinearity (g-l), according to SI Sec.\ref{subsec:SI The-neural-dynamical}. For ReLU the parameters are $\sigma_0=0.2,\sigma=1$, and for error function $\sigma_0=0.2,\sigma=10$.

\textbf{Figure 3: } The NTK theory (Sec.\ref{subsec:Gradient-driven-phase-correspond}) was calculated for the ReLU network with one hidden layer, and initialization variance $\sigma_0=1.0$. The task is binary classification between "airplane" and "frog" in CIFAR-10 dataset with $P=200$ data points. The mean predictor, variance, and correlation with the initial condition were calculated according to theory.

\textbf{Figure 4: } The covariance in a linear network was calculated according to theory (Eq.\ref{eq:main text linear var}) in a network with one hidden layer and parameters $\sigma_0=1$. The task is binary classification between "airplane" and "frog" with $P=200$ data points. The theory was calculated. In (a), the values of $\sigma$ are noted in the legend, and in (b), $\sigma=1$.

\textbf{Figure 5: } The mean predictor was calculated at the limit $T\rightarrow0$, according to Sec.\ref{sec:lowT}. The starting point is the NTK equilibrium, after the initial gradient driven phase. The network is with one hidden layer and activation function according to figure titles. The mean field equations at low $T$ (Eqs.\ref{eq:f1}, \ref{eq:scaled time meanf}) was solved numerically by inverting the $\left\{ t\times P\right\} \times\left\{ t\times P\right\} $ kernel matrix. The theory was calculated with $dt=0.0005$. The task is the binary classification between "airplane" and "frog" in the CIFAR-10 dataset, with $P=1000$ data points and $P_{test}=2000$ test points. The results presented are the bias averaged over the test points $\frac{1}{P_{test}}\sum_{{\bf x}}\left(Y\left({\bf x}\right)-\left\langle f\left(t,{\bf x}\right)\right\rangle \right)^{2}$. The accuracy was calculated with a threshold at $\left\langle f\left(t,{\bf x}\right)\right\rangle =0$, and the categories label are $\pm1$, such that a correct answer is when the predictor and the label have the same sign.

\textbf{Figure 6: } We simulated a deep network with one hidden layer and ReLU neurons. The weights are drawn from a Gaussian distribution with zero mean and variance $\sigma^2=1$. The hidden layer size is $N=10^4$. For each data point, the hidden layer neuron with maximum activation was chosen, and the activation was normalized by its maximum value such that the diagonal is always 1. (a) The task is binary classification between the digits 0, 1 in MNIST, with $P=200$ data points. (b) Binary classification between the digits 4, 9 in MNIST, with $P=200$ data points. (c) A low dimensional data was constructed, as a sum of harmonics with decaying amplitude governed by a single scalar $\theta$. Each data point ${\bf x}$ obeys ${\bf x}=\sum_{n=1}^{\infty}\frac{1}{n}{\bf v}_{n}\cos\left(n\theta\right)$, where ${\bf v}_n\in\mathbb{R}^N$ is a $N=10^4$ dimensional Gaussian vector, which induces both random direction and a random phase. $P=200$ angles $\theta$ were drawn from the range $[0,\pi]$ such that $\theta_\mu=
\pi\mu/P$. In the figure, the series was cut at $n_{max}=30$.  

\textbf{Figure 7: } We simulated a deep network with one hidden layer and ReLU neurons. The weights dynamics was Langevin dynamics of the Gaussian prior (without learning), such that they obey the time-dependent statistics described in Eq.\ref{eq:prior statistics}, with $\sigma=1,T=10^{-3}$. The data is the sum of harmonics described in Figure 6 detailed above, and for each data point the neurons with maximum activation was chosen and the activation was normalized. In (a-d) we let each neuron drift according to the prior and track its activation on the different data points, and the times are marked in the title. In (e-h) for each time frame we reorder the same neurons according to their maximum activation, similar to the starting time.  

\textbf{Figure 8: } We simulated a deep network with one hidden layer and ReLU neurons. The weights dynamics was Langevin dynamics (Eq.\ref{eq:Langevin}). The network is trained on binary classification in CIFAR-10 dataset between the categories "airplane" and "frog" with $P=400$ training data points. The parameters are $lr=0.1$, $\sigma=0.3,\sigma_0=0.7,T=10^{-3}$, and the size of the hidden layer is $N=10^4$. The activations of the neurons in the hidden layer were tracked, and for each epoch after $t_0=10^4$, we computed the SVD of the activation matrix $\phi({\bf z}^{l=1}_t({\bf x}_\mu))\in\mathbb{R}^{N\times P}$ for all the training inputs ${\bf x}_\mu$. For each SVD we took the normalized top right singular vector $h(\tau)\in\mathbb{R}^P$  and the top left normalized singular vector ${\bf g}(\tau)\in\mathbb{R}^N$ , and computed $\rho_h(\tau)=h^\top(t_0+\tau)h(t_0),\rho_{\bf g}(\tau)={\bf g}^\top(t_0+\tau){\bf g}(t_0)$. We plotted the two cosine similarities as a function of time difference from $t_0$. 

\textbf{Figure 9: } We consider the mean predictor with frozen readout weights at time $t_0$ after learning ${\bf a}(t_0)$, and the hidden layers weights ${\bf W}(t)$ drift withtout learning. We calculate the mean predictor in this scenario using Eq.\ref{eq:drift predictor}, in MNIST dataset with $P=10^4$ training data points. For each time $t$ (in the titles of the subfigures), we trained a threshold $C$ using perceptron algorithm such that $f>C$ is classified as $+1$ and $f<C$ is $-1$, and the target labels are the original labels of the data points.  The accuracy was evaluated by comparing the classification of the threshold and the original labels. 

\textbf{Figure 10: } We consider the mean predictor with frozen readout weights at time $t_0$ after learning ${\bf a}(t_0)$, and the hidden layers weights ${\bf W}(t)$ drift withtout learning, in MNIST dataset with $P=2\cdot10^4$ training data points. Each neuron receives inputs from a limited receptive field. To simulate that, we split each MNIST image to $B^2$ patches, where each patch is with size $m\times m$, and $B=28/m$.  We calculate the mean predictor in this scenario using Eq.\ref{eq:drift predictor}, with a modified kernel as described in SI Sec.\ref{subsec:limited receptive field}. Similar methods to Figure 9 were used to asses the accuracy in this scenario.

\end{document}